\documentclass[graybox]{svmult}

\usepackage{type1cm}        
\usepackage{makeidx}         
\usepackage{graphicx}        
\usepackage{multicol}        
\usepackage[bottom]{footmisc}

\usepackage{newtxtext}       %
\usepackage[varvw]{newtxmath}       
\usepackage{mathtools}
\usepackage{subcaption}
\usepackage{booktabs}
\usepackage{todonotes}
\usepackage{algorithm}
\usepackage{algorithmic}
\usepackage{calc}
\usepackage{amsmath}

\definecolor{c1}{HTML}{8dd3c7}
\definecolor{c2}{HTML}{ffffb3}
\definecolor{c3}{HTML}{bebada}
\definecolor{c4}{HTML}{fb8072}
\definecolor{c5}{HTML}{80b1d3}
\definecolor{c6}{HTML}{fdb462}
\definecolor{c7}{HTML}{b3de69}

\definecolor{c1p}{HTML}{e41a1c}
\definecolor{c2p}{HTML}{377eb8}
\definecolor{c3p}{HTML}{4daf4a}
\definecolor{c4p}{HTML}{984ea3}
\definecolor{c5p}{HTML}{ff7f00}
\definecolor{c6p}{HTML}{ffff33}
\definecolor{c7p}{HTML}{a65628}

\definecolor{mpl1}{HTML}{1f77b4}
\definecolor{mpl2}{HTML}{ff7f0e}
\definecolor{mpl3}{HTML}{2ca02c}
\definecolor{mpl4}{HTML}{d62728}
\definecolor{mpl5}{HTML}{9467bd}
\definecolor{mpl6}{HTML}{8c564b}
\definecolor{mpl7}{HTML}{e377c2}

\makeindex             

\begin{document}

\title*{Advances in Scientific Machine Learning for Coupled Fluid Flow and Transport}
\author{Gabriel F. Barros\orcidID{0009-0000-1259-510X}, \\ 
Rômulo M. Silva \orcidID{0000-0003-1755-2605} and \\
Alvaro L. G. A. Coutinho\orcidID{0000-0002-4764-1142}}
\institute{G. F. Barros \at COPPE $-$ Federal University of Rio de Janeiro - UFRJ, Rio de Janeiro, Brazil  \\ \email{gabriel.barros@coc.ufrj.br}
\and R. M. Silva \at COPPE $-$ Federal University of Rio de Janeiro - UFRJ, Rio de Janeiro, Brazil  \\ \email{romulo.silva@coc.ufrj.br}
\and A. L. G. A. Coutinho \at COPPE $-$ Federal University of Rio de Janeiro - UFRJ, Rio de Janeiro, Brazil \\ \email{alvaro@nacad.ufrj.br}}

%
%
\maketitle

\abstract*{Each chapter should be preceded by an abstract (no more than 200 words) that summarizes the content. The abstract will appear \textit{online} at \url{www.SpringerLink.com} and be available with unrestricted access. This allows unregistered users to read the abstract as a teaser for the complete chapter.
Please use the 'starred' version of the \texttt{abstract} command for typesetting the text of the online abstracts (cf. source file of this chapter template \texttt{abstract}) and include them with the source files of your manuscript. Use the plain \texttt{abstract} command if the abstract is also to appear in the printed version of the book.}

\abstract{
This chapter provides an overview of recent advances in Scientific Machine Learning (SciML) for modeling coupled fluid flow and transport phenomena governed by the incompressible Navier–Stokes and scalar transport equations. These systems, common in applications such as turbidity currents and thermal convection, exhibit strong nonlinear coupling and multiscale behavior, making high-fidelity simulations computationally expensive. To address this, the chapter reviews state-of-the-art SciML approaches for constructing efficient surrogate models, including linear reduced-order methods based on Singular Value Decomposition (e.g., Dynamic Mode Decomposition) and nonlinear neural network-based techniques such as Physics-Informed Neural Networks (PINNs) and $\beta$-Variational Autoencoders ($\beta-$VAEs). First, we cover some of the authors' work regarding the use of these models with High Performance Computing (HPC) strategies such as Adaptive Mesh Refinement/Coarsening (AMR/C) and scientific floating-point data compression. Then, we bring two new contributions regarding these applications: surrogate modeling of turbidity currents through PINNs and the extraction of disentangled nonlinear modes from thermal flows using $\beta-$VAEs. The governing equations and representative benchmark problems, including lock-exchange flows and Rayleigh–Bénard convection, are presented to illustrate these methodologies. This chapter is purposefully long to cover the mathematical and physical foundations of coupled fluid flow as well as the computational aspects for the state-of-the-art modeling in this field. Overall, the chapter demonstrates how SciML can enable fast and accurate approximations of complex coupled systems within the specific data regimes and modeling assumptions considered here, while significantly reducing computational cost relative to full-order simulations. Broader capabilities such as real-time prediction and uncertainty quantification remain active research directions whose feasibility depends strongly on the problem at hand.. 
}

\section{A Computational Science Perspective on Coupled Fluid Flows}
Coupled fluid flows represent a class of physical phenomena where the fluid dynamics interact inherently with secondary transport mechanisms \cite{Jiji2009, deteix2014}. The fundamental description of the fluid motion is governed by the Navier-Stokes equations, which dictate the conservation of mass and momentum. When coupled with a transport equation, the system describes how a scalar quantity, such as temperature or concentration, is advected by the flow field and diffuses through the medium. 

These coupled systems are ubiquitous in natural and engineering environments. Contaminant dispersion \cite{allen1982numerical, patil2014contaminant} in water bodies is a critical environmental application in which pollutants are carried by river or ocean currents. Turbidity currents represent another complex example, involving the flow of sediment-laden water down submerged slopes driven by density differences \cite{Meiburg, NECKER2002279, NECKER_HARTEL_KLEISER_MEIBURG_2005}. Thermal fluid flow \cite{bae2005direct, deen2012direct}, prevalent in heat exchangers and atmospheric convection, demonstrates the tight interplay between temperature gradients and fluid motion. Other instances include chemical reactions in fluid flows \cite{leonard1988direct} and salinity transport in estuaries \cite{XING201353}.

Resolving these physical processes presents profound mathematical and computational challenges. The tight coupling between the governing equations requires simultaneous or highly coordinated iterative solution strategies, significantly increasing the computational burden \cite{valli_1}. These systems typically exhibit multiscale behavior and complex nonlinear phenomena, such as turbulence, which require extremely fine spatial and temporal discretization to be accurately captured. Consequently, a numerical model must resolve an exceptionally large number of degrees of freedom, often pushing the limits of current computational architectures and rendering high-fidelity simulations prohibitively expensive for iterative design or real-time forecasting \cite{altman2018curse, Peherstorfer2018SurveyOptimization}. The current state of the art in tackling these coupled systems relies heavily on advanced computational fluid dynamics methodologies. High-fidelity approaches, encompassing Direct Numerical Simulation and Large Eddy Simulation, provide accurate resolutions of the governing physics at the cost of immense computational resources \cite{axtmann2017scalability}. To mitigate these costs, industry and research often employ Reynolds-Averaged Navier-Stokes models combined with specialized scalar transport solvers. Furthermore, advanced discretization techniques, including high-order finite element, finite volume, and spectral methods, are continuously optimized alongside highly parallelized solvers running on massive supercomputers to handle the enormous scale of these discrete coupled problems \cite{rudi2015extreme}.

Scientific Machine Learning (SciML) has recently emerged as a transformative paradigm to address the computational bottlenecks of traditional numerical methods. SciML models can be purely data-driven methods with algebraic foundations such as Operator Inference (OpInf) \cite{PEHERSTORFER2016196}, Reduced Basis (RB) \cite{quarteroni2015reduced}, Proper Orthogonal Decomposition (POD) \cite{Lumley1967} and Dynamic Mode Decomposition (DMD) \cite{Schmid2010DynamicData, Schmid2021DynamicVariants, COLBROOK2024127}. Most of these models are based on some kind of factorization of the dataset, such as the Singular Value Decomposition (SVD). Another subset of SciML models rely on neural networks (NNs) for their architecture. We highlight methods based on operator learning, namely neural operators \cite{Kovachki2023}, such as Deep Operator Networks (DeepONets) \cite{Lu2021} and Fourier Neural Operators (FNOs) \cite{Li2021}, besides a wide family of autoencoders, variational autoencoders, convolutional neural networks (CNNs) \cite{ruthotto2020deep}, and more. For NN-based SciML models, the physics can be embedded into the loss functions, improving the physical sense of the model's solutions \cite{karniadakis2021physics, raissi2019physics}. These models can approximate the solution manifold of the coupled Navier-Stokes and transport equations, enabling rapid surrogate evaluations, each model with its pros and cons. Under suitable assumptions regarding data availability, training regimes, and the complexity of the target dynamics, the integration of SciML can accelerate the prediction of complex scalar transport phenomena, and open new avenues for inverse problem solving, uncertainty quantification, and real-time control in coupled fluid environments \cite{chen2026aihw2035shapingdecade}. However, the accuracy and generalization of these models are inherently dependent on the quality and coverage of the training data, the choice of hyperparameters, and the specific flow regime under consideration. Extrapolation outside the training distribution remains a significant challenge, and the computational cost of training neural network-based models can be substantial.

The goal of this chapter is to cover, to the best of our knowledge, the state-of-the-art of SciML modeling on fluid flow applications where fluid motion and transport phenomena are strongly coupled. In specific sections, we present some of the authors' contributions to the field for illustrative purposes. For more theoretical sections, we add a summary at the end to cover the most important topics discussed. The structure of this chapter is: In Section \ref{sec:eqs}, we introduce the governing equations for coupled fluid flow problems. We also show how the coupling occurs between the transport equation and the incompressible Navier-Stokes equations for turbidity currents and thermal fluid flows. We finish the section by presenting two representative problems: the lock-exchange configuration on turbidity currents, a form of gravity current driven by the excess density due to suspended sediment within the flow, and the Rayleigh-Bénard instability in thermal fluid flows. Next, in Section \ref{sec:sciml}, we cover how SciML methods can be used to improve scientific discovery or improve many-query processes in coupled fluid flow applications. This section is split into two: SVD-based models are covered in Section \ref{sec:svd}, and NN-based models are discussed in Section \ref{sec:nn}. Additionally, we highlight the authors' contributions and add two more: the reconstruction and parameter estimation of gravity currents using Physics-Informed Neural Networks (PINNs) and the extraction of disentangled nonlinear modes from complex Rayleigh-Bénard simulations using adaptive weights $\beta-$VAEs. These contributions are exposed in sections \ref{sec:pinns} and \ref{sec:betavaes}, respectively. The Chapter ends with a summary of our main findings and future perspectives.


\section{Governing Equations}
\label{sec:eqs}
\par 
The mathematical formulation of coupled fluid flows relies on defining a robust system of partial differential equations, supplemented by precise initial and boundary conditions. These conditions are paramount for ensuring a well-posed problem and represent the physical constraints of the specific domain being modeled. That said, coupled fluid flows are modeled through the Navier-Stokes equations and the transport equation. In this chapter, we focus only on incompressible flows, meaning that the fluid motion is dictated by the Incompressible Navier-Stokes equations. With that, for a given spatial domain $\Omega \subset \mathbb{R}^d$ (where the spatial dimension $d \in \{2, 3\}$) bounded by a sufficiently smooth surface $\partial\Omega$, and over a temporal interval $t \in [0, T]$, the evolution of the flow is determined by the conservation of mass and momentum. The fluid state is completely characterized by its velocity vector field $\mathbf{u}(\mathbf{x}, t)$ and its kinematic pressure field $p(\mathbf{x}, t)$, where $\mathbf{x} \in \Omega$ represents the spatial coordinates and $t$ represents time. The incompressible Navier-Stokes equations are expressed mathematically as:

\begin{align}
    \nabla \cdot \mathbf{u} = 0, \\
    \frac{\partial \mathbf{u}}{\partial t} +(\mathbf{u} \cdot \nabla ) \mathbf{u} = -\nabla p +\nu \nabla^2 \mathbf{u} + \mathbf{f}
\end{align}

\noindent
where $\mathbf{u}$ is the fluid velocity, $p$ is the pressure, $\nu$ denotes the kinematic viscosity of the fluid, and $\mathbf{f}$ represents the sum of all external body forces acting on the fluid volume, such as gravity or electromagnetic fields. In the problems treated in this chapter, the body force $\mathbf{f}$ depends on the transported scalar field $\phi$, i.e., $\mathbf{f}=\mathbf{f}{(\phi})$. This dependency arises from the Boussinesq approximation (detailed in Section \ref{sec:coupled_form}), in which density variations, driven by temperature or concentration differences, are neglected everywhere except in the gravitational body force term. Simultaneously, the transport of a scalar quantity $\phi(\mathbf{x}, t)$, which might represent sediments, temperature or chemical concentration, is governed by the advection-diffusion-reaction equation:


\begin{equation}
    \frac{\partial \phi}{\partial t} + (\mathbf{u} \cdot \nabla )\phi = \alpha \nabla^2 \phi +S \phi
\end{equation}

\noindent where $\alpha$ is the diffusion coefficient of the scalar within the fluid medium, and $S$ accounts for any internal sources or sinks generating or depleting the scalar quantity (such as chemical reactions). The transport equation depends on the velocity field, while the body force depends on the scalar. This coupling is what makes the system two-way coupled: the scalar field affects the fluid motion through buoyancy, while the fluid motion advects the scalar field.
\par 
To mathematically close this system of partial differential equations and guarantee a unique solution, physically consistent initial and boundary conditions must be established. Initial conditions are responsible for prescribing the precise spatial distribution of the state variables throughout the entire computational domain at the onset of the simulation. This establishes the fundamental baseline physical state from which the transient evolution of the fluid and scalar fields is computed, evaluated at time $t=0$:

\begin{align}
    \mathbf{u}(\mathbf{x}, 0) = \mathbf{u}_0(\mathbf{x}) \quad \forall \mathbf{x} \in \Omega, \\
    \phi(\mathbf{x}, 0) = \phi_0(\mathbf{x}) \quad \forall \mathbf{x} \in \Omega
\end{align}

\noindent
Boundary conditions dictate how the fluid and the scalar field interact with the physical boundaries $\partial\Omega$ for all subsequent times $t > 0$. Dirichlet boundary conditions impose explicit, fixed values directly on the boundary surface, such as the no-slip condition at a solid wall ($\mathbf{u} = \mathbf{u}_{\text{wall}}$), a constant wall temperature, or a fixed chemical concentration at an inlet. Alternatively, Neumann boundary conditions specify the spatial gradient or flux of a quantity at the boundary interface. Common applications include adiabatic solid walls, modeled by setting the normal scalar gradient to zero, and computational outflow boundaries, which assume zero normal gradients for the velocity components to simulate fully developed flow exiting the domain. Furthermore, because the incompressible Navier-Stokes equations solve for the pressure gradient rather than the absolute pressure, it is strictly necessary to prescribe the pressure value at a minimum of one reference point within the domain or boundary to ensure a well-posed and solvable problem.

\subsection{Distinguishing Mass and Energy Transport}
The general scalar transport equation introduced previously serves as a fundamental template. However, the physical interpretation of the transported variable $\phi$ fundamentally alters the nature of the physical problem. In most engineering and environmental fluid mechanics applications, this scalar represents either mass concentration or thermal energy. Distinguishing between these two transport mechanisms is crucial prior to examining their respective coupling effects with the fluid flow. In mass transport modeling, the scalar quantity $ c$ represents the concentration of a distinct chemical species or particulate matter. The governing equation describes how this species is advected by the bulk fluid motion and diffuses due to spatial concentration gradients. In this context, the diffusion coefficient is the mass diffusivity, $D$. This process is driven by random molecular motion, resulting in a net flux of the species from regions of high concentration to regions of low concentration, as described by Fick's law of diffusion. Mass transport governs physical phenomena such as the dispersion of chemical pollutants in a river, the mixing of fresh and saltwater in estuaries, and gravity currents in the atmosphere and underwater. Conversely, energy transport focuses on the spatial and temporal evolution of the fluid temperature field, $T$. The transport equation in this scenario is derived directly from the fundamental principle of conservation of energy. The relevant diffusion parameter is the thermal diffusivity, $\alpha = k_h / (\rho c_p)$, where $k_h$ is the thermal conductivity, $\rho$ is the fluid density, and $c_p$ is the specific heat capacity at constant pressure. Unlike mass transport, which inherently involves the physical translation of distinct molecules relative to the mixture, thermal diffusion involves the transfer of kinetic energy between adjacent molecules through collisions. Mathematically, this mechanism is described by Fourier's Law, and it dictates temperature distributions in applications ranging from industrial heat exchangers to atmospheric thermals. While both mechanisms are mathematically described by the advection-diffusion equation, their physical driving forces and typical timescales diverge significantly. The rate at which mass diffuses is often orders of magnitude slower than the rate at which momentum or heat diffuses in a given fluid. This fundamental physical disparity requires careful consideration when establishing the computational mesh~\cite{carey-grids}, time step sizes~\cite{valli2005control}, and the numerical algorithm for handling coupling in numerical simulations, namely, monolithic, staggered, or decoupled schemes~\cite{valli_1}. Because mass diffuses slowly, the concentration boundary layers will be much thinner than the hydrodynamic or thermal boundary layers, requiring highly refined spatial discretization to resolve them accurately.

\subsection{Derivation of Coupled Formulations}\label{sec:coupled_form}
To transition from the decoupled Navier-Stokes and transport equations to a fully two-way coupled system, we assume the validity of the Boussinesq approximation. This widely adopted simplification assumes that variations in fluid density are negligible in the continuity and inertial terms, becoming significant solely within the gravitational body force term of the momentum equation, denoted previously as $\mathbf{f}$. By defining a reference state with density $\rho_0$, temperature $T_0$, and concentration $C_0$, we can express the local density $\rho$ through a linearized equation of state. This approach mathematically links the scalar transport directly to the fluid's momentum, enabling the modeling of buoyancy-driven flows.

\subsubsection{Energy Transport: Rayleigh-Bénard Convection}In Rayleigh-Bénard convection, a fluid layer is destabilized by a vertical temperature gradient. The density varies with temperature according to $\rho = \rho_0 [1 - \beta_T(T - T_0)]$, where $\beta_T$ is the coefficient of thermal expansion. Substituting this relationship into the gravitational body force yields a buoyancy term proportional to the local temperature difference. In dimensional form, the Boussinesq-approximated momentum equation reads:

\begin{equation}
\dfrac{\partial{\mathbf{u}}}{\partial t} + (\mathbf{u} \cdot \nabla ) \mathbf{u} = -\frac{1}{\rho_0} \nabla p +\nu \nabla^2 \mathbf{u} + g \beta_T (T - T_0) \mathbf{e}^g
\end{equation}

To generalize the system and identify the dominant physical mechanisms, we non-dimensionalize the variables. For thermally driven flows, the standard practice uses the domain height $L$ as the reference length, the thermal diffusion time scale $L^2/\alpha$ as the reference time, and a reference temperature difference $\Delta T$ to scale the temperature field. Defining the dimensionless variables as $\mathbf{x}^{*} = \mathbf{x}/L$, $t^{*} = t \alpha / L^2$, $\mathbf{u}^{*} = \mathbf{u} L/ \alpha$, $p^{*} = p L^2 / (\rho_0 \alpha^2)$, and $T^{*} = (T - T_0) / \Delta T$, and substituting into the dimensional equations, we obtain the following. The momentum equation becomes (dropping the asterisks for brevity):

\begin{align}
		\nabla \cdot \mathbf{u} = 0, \\
		\frac{\partial \mathbf{u}}{\partial t} + (\mathbf{u} \cdot \nabla )\mathbf{u} + \nabla p - \sqrt{\frac{Pr}{Ra}} \nabla^2 \mathbf{u} - T \mathbf{e}^g = 0, \\
		\frac{\partial T}{\partial t} + (\mathbf{u} \cdot \nabla )T - \nabla^2 T = 0, 
\end{align}



Here, $T$ now represents the dimensionless temperature field, and $\mathbf{e}^g$ is the unit vector aligned with gravity. The system dynamics are dictated by two critical dimensionless parameters. The Prandtl number ($Pr = \nu/\alpha$) describes the fluid's intrinsic properties. The Rayleigh number ($Ra = \frac{g \beta_T \Delta T L^3}{\nu \alpha}$) dictates the onset of convective instability, balancing the driving buoyancy forces against the stabilizing effects of thermal and momentum diffusion.

\subsubsection{Mass Transport: Turbidity Currents}
Turbidity currents are underwater flows where buoyancy variations are induced by suspended particulate mass rather than thermal energy \cite{NECKER2002279, Meiburg, NECKER_HARTEL_KLEISER_MEIBURG_2005}. The local mixture density is modeled as $\rho = \rho_0 [1 + \beta_c(c - c_0)]$, where $\beta_c$ represents the volumetric expansion coefficient due to the mass concentration $c$. Here,  $c$ is normalized to map the interaction between fluids such that $c = c_B/c_A$ - being $c_A$ and $c_B$ are the characteristic concentrations of the ambient and sediment-laden fluids, respectively - and equals $c = 1.0$ for the part of the domain where the fluid is mixed with sediments and $c = 0.0$ for the remainder of the domain where the fluid does not contain sediments. This concentration directly modifies the body force term, driving the heavier fluid mixture along submerged inclines. For mass-driven flows, we scale the system using the viscous diffusion time scale $L^2/\nu$, rather than the thermal scale used in convection. Non-dimensionalizing the spatial coordinates with $L$ and the concentration with a characteristic difference $\Delta c$, the dimensionless governing equations take the following form:


\begin{align}
		\nabla \cdot \mathbf{u} = 0,  \label{eq:mass_pdgf}\\
		\frac{\partial \mathbf{u}}{\partial t} + (\mathbf{u} \cdot \nabla ){\mathbf{u}} + \nabla p - \frac{1}{\sqrt{Gr}} \nabla^2 \mathbf{u} - c \mathbf{e}^g = 0, \label{eq:momentum_pdgf}\\
		\frac{\partial c}{\partial t} + (\mathbf{u} + u_s \mathbf{e}^g) \cdot \nabla c - \frac{1}{Sc \sqrt{Gr}} \nabla^2 c = 0,  \label{eq:transport_pdgf}
\end{align}

In the above, Eq. \eqref{eq:mass_pdgf} is the incompressibility constraint, Eq. \eqref{eq:momentum_pdgf} is the momentum equation where the term $\frac{1}{\sqrt{Gr}} \nabla^2 \mathbf{u}$ represents dimensionless viscous diffusion and $c \mathbf{e}^g$ is the buoyancy forcing proportional to the local sediment concentration, and Eq. \eqref{eq:transport_pdgf} governs the evolution of the sediment concentration field. Here, $u_s$ is the dimensionless sediment fall velocity, which accounts for the gravitational settling of particles relative to the surrounding fluid. After non-dimensionalizing the equations, two dimensionless numbers are obtained: $Gr = g \beta_c \Delta c L^3 / \nu^2$ is the Grashof number, which expresses the ratio of buoyancy to viscous effects, and $Sc = \nu/D$ is the Schmidt number, which represents the ratio of momentum diffusivity to mass diffusivity. In the transport equation, the advection term $(\mathbf{u} + u_s \mathbf{e} ^g) \cdot \nabla c$ combines the fluid-driven advection with the settling effect, while $\frac{1}{Sc\sqrt{Gr}} \nabla^2 c$ represents molecular diffusion of the sediment, scaled by both the Schmidt and Grashof numbers. The $1/\sqrt{Gr}$ factor in both the momentum and transport equations arises from the choice of scaling: using the buoyancy velocity $u_b = \sqrt{g'L}$ (where $g' = g \beta_c \Delta c$ is the reduced gravity) as the velocity scale and $L/u_b$ as the time scale, the viscous term scales as $\nu/(u_b L) = 1/Re_b$, where $Re_b = \sqrt{Gr}$ is the buoyancy Reynolds number.


The dimensionless numbers governing coupled fluid flows provide a fundamental framework for comparing disparate physical systems. The Prandtl number ($Pr$) and the Schmidt number ($Sc$) serve analogous roles in their respective domains. $Pr$ quantifies the ratio of momentum diffusivity to thermal diffusivity, whereas $Sc$ represents the ratio of momentum diffusivity to mass diffusivity. Both numbers are intrinsic fluid properties that determine the relative thicknesses of the hydrodynamic, thermal, and concentration boundary layers. The Grashof number ($Gr$) and the Rayleigh number ($Ra$) quantify the driving forces in buoyancy-induced flows. $Gr$ isolates the competition between buoyancy and viscous forces. $Ra$ incorporates the thermal diffusion rate by multiplying the thermal Grashof number by the Prandtl number ($Ra = Gr \times Pr$). This formulation makes $Ra$ the definitive parameter for establishing the onset of thermal convection. These buoyancy parameters are fundamentally related to the Reynolds number ($Re$), which defines the ratio of inertial to viscous forces in forced flows. In mixed convection regimes, the ratio $Gr / Re^2$ determines the dominant flow mechanism. When $Gr / Re^2 \gg 1$, natural buoyancy forces govern the system. When $Gr / Re^2 \ll 1$, forced advection dictates the flow dynamics, and the coupling is effectively one-way. In purely gravity-driven flows, the Reynolds number ceases to be an independent input and emerges as a consequence of the Grashof or Rayleigh numbers, scaling as $Re \sim \sqrt{Gr}$.

\subsection{Lock-exchange and Rayleigh-Bénard Problems}
\label{sec:le_rb}
Having established the dimensionless governing equations for energy and mass transport, it is necessary to transition from these theoretical formulations to the specific computational datasets that serve as the foundation for our SciML analyses. The training and validation of robust surrogate models require high-fidelity data that accurately captures the complex nonlinear dynamics of two-way coupled fluid flows. In this chapter, we focus on two canonical problems to evaluate the proposed data-driven methodologies.

\begin{figure}[ht]
    \centering
    \begin{subfigure}[b]{\textwidth}
    		\centering
    		\includegraphics[width=\linewidth]{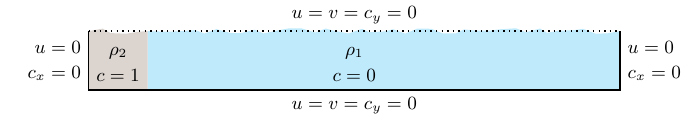}
    		\subcaption{Tank sketch with the boundary conditions for concentration and velocities. The density of the heavier and lighter fluids are denoted by $\rho_2$ and $\rho_1$, respectively.}
    		\label{fig:tank_scheme}
    	\end{subfigure}
    	\hfill
    	\begin{subfigure}[b]{\textwidth}
    		\centering
    		\includegraphics[width=\textwidth]{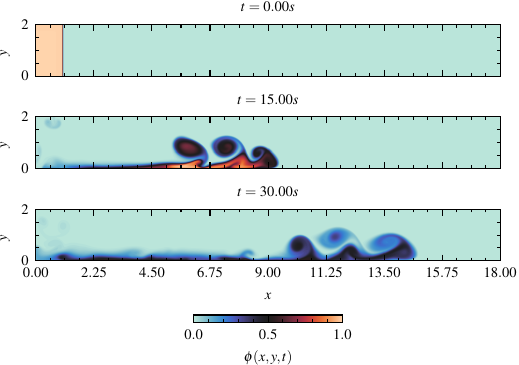}
    		\subcaption{Evolution of the concentration over time. At time $t=0$, we allow the two fluids to mix, driven by gravity. }
    		\label{fig:phi_fulldomain}
	\end{subfigure}
	\label{fig:setup_solution}
	\caption{Lock-exchange problem at $Sc = 1.0$ and $Gr=
5\times10^6$. (a) Schematic of the computational domain showing the boundary conditions: no-slip walls for velocity ($\mathbf{u}=\mathbf{0}$) on all boundaries and zero-flux conditions for sediment concentration. The heavier fluid ($\rho_2$, with $c=1$) initially occupies the left column of dimensions $H \times S$, while the lighter fluid ($\rho_1$, with $c=0$) fills the remainder. (b) Temporal evolution of the sediment concentration field: upon the lock release, the denser fluid collapses and propagates rightward along the bottom of the tank, while the lighter fluid propagates leftward along the top. Kelvin--Helmholtz instabilities develop at the interface due to the velocity shear between the two counter-flowing layers.}
\end{figure}

To investigate mass transport mechanisms, we analyze the lock-exchange problem, a standard benchmark for turbidity and density-driven gravity currents. In this case, we have a 2D tank filled with water, with a vertical column of sediment in the leftmost region. That is, a rectangular domain with length $L = 18$, height $H = 2$. The domain boundary $\partial \Omega$ is partitioned into two complementary subsets:$\Gamma_{in}$, representing segments where Dirichlet (essential) boundary conditions are applied, and $\Gamma_h$, representing segments where Neumann (natural) boundary conditions are imposed, such that $\partial \Omega = \Gamma_{in} \bigcup \Gamma_h$ and $\Gamma_{in} \bigcap \Gamma_h = \emptyset$. For the momentum equations, the Dirichlet condition is $\mathbf{u} = \mathbf{g}$ on $\Gamma_{in}$, where $\mathbf{g}$ is a prescribed velocity, and the Neumann condition is $(-p \mathbf{I} + \frac{1}{\sqrt{Gr}} \nabla \mathbf{u}) \cdot \mathbf{n} = \mathbf{h}$ on $\Gamma_h$, where $\mathbf{h}$ is the prescribed traction and $\mathbf{n}$ is the outward unit normal. For the transport equation, the boundary conditions are the Dirichlet condition $c = c_{in}$ on $\Gamma_c$, which describes the quantity of sediment entering in the flow domain, and the no-flux boundary condition $(u_s \mathbf{g} c - \frac{1}{Sc \sqrt{Gr}} \nabla c) \cdot \mathbf{n} = \mathbf{0}$ on $\Gamma_h$. In the lock-exchange configuration, the tank is a closed system: all walls are solid surfaces with no-slip conditions for velocity ($\mathbf{g} = \mathbf{0}$, meaning the fluid is at rest at the walls) and zero-flux conditions for the sediment concentration ($c_{in} = 0$, meaning no sediment enters or leaves through the boundaries). No external injection of fluids or sediments is considered. For the initial conditions, we assume a divergence-free velocity field. For the transport equation, the initial conditions are such that the heavy fluid ($c = 1$) is represented as a column with dimensions $S \times H = 1$ $\times$  $2$ area units located at the left border of the tank, and the light fluid ($c = 0$) fills the rest of the domain. Figure \ref{fig:tank_scheme} illustrates the domain, boundary conditions, and the initial conditions. For this problem, we consider a turbulent flow such that $Gr = 5 \times 10^6$ and $Sc = 1.0$. The remaining figures in Fig. 1 cover some snapshots. The high-fidelity data for this configuration is generated through dedicated numerical simulations. We employ a fully coupled finite element  code that solves the transient mass and momentum equations simultaneously. To guarantee numerical stability and capture the intricate subgrid-scale physics characteristic of turbulent mixing fronts, the solver utilizes a Residual-Based Variational Multiscale formulation~\cite{lins2010residual, guerra2013numerical}. The RB-VMS approach is a stabilized finite element method that models the effect of unresolved fine scales on  the resolved coarse scales by adding residual-based stabilization terms to the variational formulation. This is particularly important for convection-dominated flows at high Grashof numbers, where standard Galerkin methods suffer from spurious oscillations near sharp fronts and boundary layers. The method provides consistent stabilization without introducing excessive numerical diffusion, thereby preserving the physical sharpness of the sediment concentration fronts characteristic of turbidity currents. This computational framework is implemented within the open-source FEniCS computing platform \cite{logg2012automated}. The spatial discretization uses a $700$ by $100$ cell mesh, where each cell is divided into two linear triangles. We simulate the flow evolution up to $30$ seconds.

For the thermal fluid flow regime, we analyze the Rayleigh-Bénard convection problem. The high-fidelity snapshot data used for this energy transport case is sourced directly from The Well dataset\footnote{https://polymathic-ai.org/the\_well/datasets/rayleigh\_benard/}, a repository curated by Polymathic AI \cite{ohana2024well} for benchmarking machine learning models in scientific domains. This dataset provides rigorously resolved fields for velocity, pressure, and temperature across a spectrum of Rayleigh numbers. Utilizing this standardized data allows for a transparent evaluation of deep learning architectures tasked with capturing buoyancy-driven thermal instabilities.

The computational domain is a two-dimensional box with horizontal length $L = 4$ and vertical height $H = 1$. The spatial discretization uses a $512 \times 128$ grid. To accurately capture the boundary-layer physics near the solid walls, the grid is uniformly sampled along the horizontal axis and uses Chebyshev nodes along the vertical axis. The physical boundary conditions applied to this domain are periodic in the horizontal direction, simulating an infinitely wide fluid layer, while Dirichlet boundary conditions are prescribed at the top and bottom walls. The fluid is assumed to be initially at rest, imposing zero fields for velocity and pressure at $t=0$. To trigger the thermal instability, the initial buoyancy field is defined by a linear background profile superimposed with a damped noise. The full dataset encompasses an ensemble of $1750$ simulations generated by varying the governing dimensionless numbers and initial noise seeds. Specifically, the Rayleigh number is evaluated across a spectrum from $10^6$ to $10^{10}$, and the Prandtl number ranges from $0.1$ to $10.0$. In this chapter, we focus on two cases: the case where $Ra = 10^6$ and $Ra = 10^{ 10}$, both with $Pr = 1$. The temporal evolution of the flow is captured over a total time range from $t=0$ to $t=50$ with a simulation time step of $0.25$, yielding $200$ recorded snapshots. 

These snapshots constitute the high-fidelity dataset used throughout the subsequent SciML analyses. Specifically, in Section \ref{sec:svd}, the snapshot matrices are assembled from these temperature fields to perform SVD-based analyses and assess the suitability of linear reduced-order models for different Rayleigh number regimes. In Section \ref{sec:betavaes}, the same snapshots serve as the training and validation data for the $\beta$-VAE framework, where the convolutional encoder processes each $512 \times 128$ temperature field as an input image to learn a nonlinear latent representation.

Figure \ref{fig:rb-setup} shows the solution of two simulations dictated by different values of $Ra$, which are used in this study. These values characterize two regimes,  sometimes referred to in the literature as "soft" and "hard" turbulence \cite{Castaing_Gunaratne_Heslot_Kadanoff_Libchaber_Thomae_Wu_Zaleski_Zanetti_1989}. Soft turbulence, illustrated here with $Ra = 10^6$, is characterized by coherent structures and moderate thermal boundary-layer interactions. The bottom row corresponds to $Ra=10^{10}$, where the flow transitions to a hard turbulent regime in which intense small-scale fluctuations dominate, thermal plumes become highly fragmented, and the boundary layers are significantly thinner.

\begin{figure}[ht]
    \centering
    \includegraphics[width=0.9\linewidth]{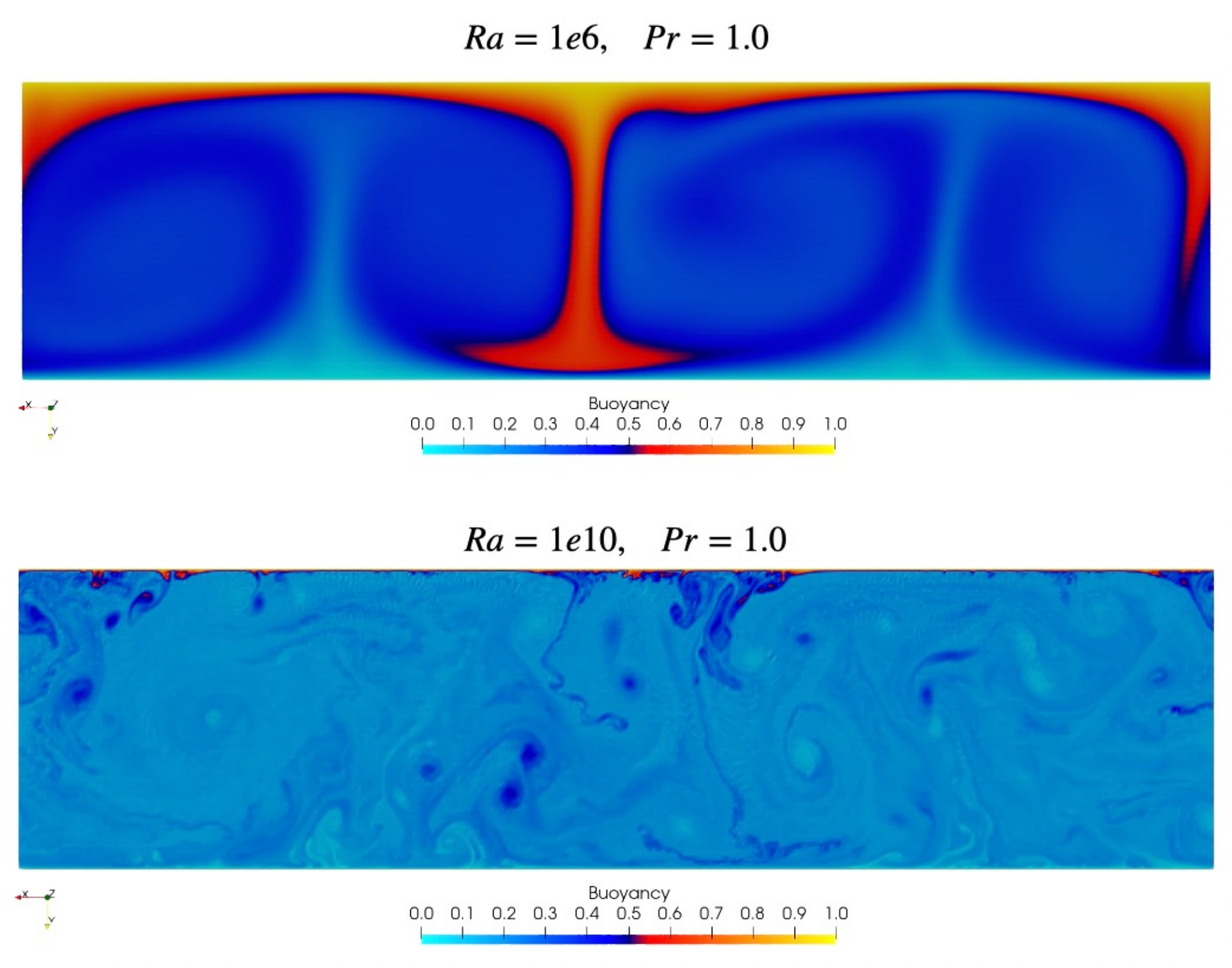}
    \caption{Buoyancy plots for Rayleigh-Bénard convection at different values of $Ra$ and $Pr=1.0$. The top row shows a late-time snapshot at $Ra=10^6$, where the flow exhibits organized large-scale convective rolls with relatively smooth temperature gradients. The distinction between these values is important because it directly impacts the feasibility of reduced-order modeling: the slow singular-value decay observed at $Ra=10^{10}$ (Section~\ref{sec:svd}) reflects this increased dynamical complexity.}
    \label{fig:rb-setup}
\end{figure}

\paragraph*{\textbf{Section Summary}} This section presented the mathematical framework underpinning coupled fluid flow and transport problems. Starting from the incompressible Navier--Stokes and scalar transport equations, we showed how the Boussinesq approximation introduces a two-way coupling through the buoyancy body force term, linking the scalar field (temperature or concentration) to the fluid momentum. The non-dimensionalization of these equations was carried out in detail for two canonical configurations: Rayleigh--B\'{e}nard convection, governed by the Prandtl and Rayleigh numbers, and turbidity currents, governed by the Grashof and Schmidt numbers. For each case, we defined the computational domain, boundary conditions, initial conditions, and the high-fidelity numerical datasets that serve as the foundation for the SciML analyses in subsequent sections. The key takeaway is that these coupled systems are computationally expensive to simulate at high fidelity, particularly in turbulent regimes, which motivates the development of efficient surrogate models.


\section{The Emergence of Scientific Machine Learning}
\label{sec:sciml}

The staggering computational demands of resolving coupled fluid systems necessitate a paradigm shift in how these problems are approached \cite{Chen31122024, chen2026aihw2035shapingdecade}. Traditional numerical approaches, hereafter referred to as High-Fidelity Simulations or Full-Order Models (FOMs), solve the discretized Navier-Stokes and transport equations directly. While FOMs provide accurate solutions to complex coupled physics, their extreme computational cost poses a severe limitation in practical engineering and research workflows.

This limitation becomes a critical bottleneck in many-query scenarios. Advanced engineering applications, encompassing active flow control, Uncertainty Quantification (UQ), and design optimization, inherently require evaluating the system's response across a vast parameter space. Executing hundreds or thousands of high-fidelity simulations to map these variations, such as fluctuating inlet conditions or varying geometric parameters, is computationally prohibitive and often infeasible within standard project timeframes.

Scientific Machine Learning (SciML) emerges as a promising methodology to overcome these computational barriers by enabling the construction of surrogate models. A surrogate model \cite{alizadeh2020managing} functions as a computationally efficient proxy for the full-order model. Rather than iteratively solving massive systems of algebraic equations, SciML leverages deep learning architectures to approximate the nonlinear mappings between the input parameters and the resulting coupled flow fields. Once trained, these neural network-based surrogates predict velocity, pressure, and scalar distributions in a fraction of the time required by a high-fidelity simulation. It is important to note, however, that the construction of reliable surrogates comes with its own challenges. The training phase itself can be computationally expensive, requiring large volumes of high-fidelity data and careful hyperparameter tuning. Furthermore, neural network-based surrogates may exhibit poor generalization when evaluated outside the training distribution and their sensitivity to network architecture, learning rate, and regularization strategies demands systematic validation before deployment in safety-critical or design-oriented workflows.

The rapid inference capabilities of these surrogate models directly unlock the potential for exhaustive many-query analyses. In the realm of design optimization, engineers can rapidly evaluate design iterations to maximize heat transfer efficiency or minimize pollutant dispersion \cite{dbouk2017review}. For Uncertainty Quantification, the computational speed of surrogates allows for extensive statistical sampling \cite{wang2022recent}. This enables researchers to propagate input uncertainties through the coupled system to robustly quantify the confidence intervals of the predicted flow metrics. Furthermore, the inherent differentiability of neural networks enables the development of sophisticated real-time control algorithms, allowing dynamic adjustment of system parameters to maintain targeted flow conditions \cite{zhao2025applications}.


To effectively construct surrogate models for complex coupled fluid flows, Scientific Machine Learning methodologies can be broadly categorized into two primary branches. The first branch encompasses data-driven techniques rooted in Singular Value Decomposition, which excel at extracting dominant coherent structures from high-dimensional data. The second branch utilizes the vast representational capacity of neural networks to approximate highly nonlinear mappings and embed physical constraints. The following subsections detail these foundational architectures, setting the theoretical stage for specific methodological contributions and targeted applications in coupled mass and energy transport.

We split this section into two topics of interest: 
\begin{itemize}
    \item Some SciML methods can predict unseen scenarios without iterative training, usually through linear manifold learning such as Singular Value Decomposition (SVD). That is, in these cases, the optimization of the model parameters is performed by minimizing a convex function, enabling straightforward convergence to the global minimum. Methods such as Proper Orthogonal Decomposition (POD), Reduced Basis (RB), and Dynamic Mode Decomposition (DMD) are SciML models that can predict unseen scenarios using bases constructed solely from SVD calculations of the available spatio-temporal data, with fast training and reasonable explainability \cite{sreekumar2026snapshots}. We cover these models in Section \ref{sec:svd}.
    \item On the other hand, other SciML methods require iterative training, especially given that their optimization processes are non-convex. With that, the gradients of the model parameters need to be updated iteratively. Methods based on neural networks, such as multilayer perceptrons (MLPs), autoencoders (AEs), Physics-Informed Neural Networks (PINNs), and $\beta-$Variational AutoEncoders ($\beta-$VAEs) are SciML models that require iterative training. We cover these models in Section \ref{sec:nn}.
\end{itemize}

\paragraph*{\textbf{Section Summary}} This section motivated the need for Scientific Machine Learning as a response to the computational bottlenecks of high-fidelity simulations in many-query scenarios such as design optimization, uncertainty quantification, and real-time control. We introduced the concept of surrogate models and discussed both their advantages---rapid inference, differentiability, and flexibility---and their limitations, including sensitivity to hyperparameters, generalization challenges, and non-trivial training costs. The SciML landscape was organized into two complementary families: SVD-based methods, which rely on linear dimensionality reduction and convex optimization, and neural network-based methods, which learn nonlinear mappings through iterative training. The following subsections detail each family and demonstrate their application to the coupled flow problems introduced in Section~\ref{sec:eqs}.

\subsection{SVD-Based Methods}
\label{sec:svd}

Singular Value Decomposition serves as the mathematical cornerstone for numerous dimensionality reduction techniques in fluid mechanics. By factorizing the snapshot matrix obtained from a high-fidelity simulation, SVD identifies an orthogonal basis that optimally represents the spatial variance of the flow field. Several methods, such as Proper Orthogonal Decomposition (POD) and Dynamic Mode Decomposition (DMD), are built under the Eckart-Young-Mirsky theorem, which states that the optimal low-rank approximation of a matrix is obtained by using the truncated SVD \cite{chipman2020proofs} illustrated in Figure \ref{fig:svd}. In this case, notice that part of the information is discarded, and a smaller number of modes (in comparison with the original matrix rank) is used to build the basis for POD and DMD.

\begin{figure}[ht!]
    \sidecaption[t]
    \centering
    \begin{minipage}[t]{0.6\textwidth}
        \vspace{-7.9cm} 
        \begin{subfigure}{\linewidth}
            \centering
            \includegraphics[width=\linewidth]{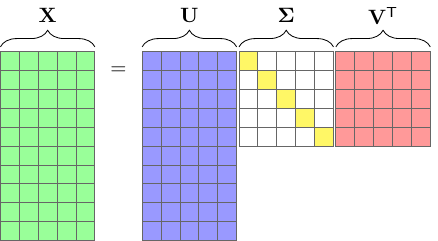}
            \subcaption[]{Singular Value Decomposition of a given matrix $\mathbf{X}$.}        
        \end{subfigure} \\
        \begin{subfigure}{\linewidth}
            \centering
            \includegraphics[width=\linewidth]{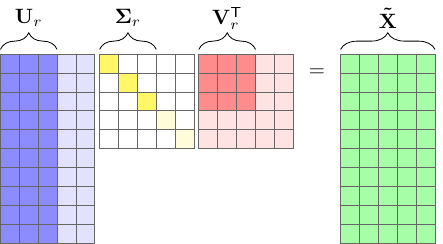}
            \subcaption[]{Low-rank approximation of $\mathbf{X}$ based on singular values and singular vectors.}
        \end{subfigure}
    \end{minipage}
    \setcounter{figure}{2}
    \caption{Illustration of the Singular Value Decomposition (SVD) and its low-rank truncation. (a) The full SVD decomposes a snapshot matrix $\mathbf{X} \in \mathbb{R}^{n \times m}$ into the product $\mathbf{U}\boldsymbol{\Sigma}\mathbf{V}^T$, where $\mathbf{U}$ contains the left singular vectors (spatial modes), $\boldsymbol{\Sigma}$ is a diagonal matrix of singular values in decreasing order, and $\mathbf{V}^T$ contains the right singular vectors (temporal coefficients). (b) The truncated SVD retains only the first $r$ modes, yielding the best rank-$r$ approximation of $\mathbf{X}$ in the Frobenius and spectral norms (Eckart--Young--Mirsky theorem). The discarded singular values and their associated modes represent fine-scale or low-energy features of the data.}
    \label{fig:svd}
\end{figure}

\par 
For instance, in POD, the modes are the columns of $\mathbf{U_r}$, which is a matrix composed of the first $r$ columns of $\mathbf{U}$. When spatially normalized and redistributed, we can extract information from the system dynamics. For instance, on Fig. \ref{fig:lock2d_modes}, we can see a snapshot of the concentration field of the turbidity flow simulation, its time-averaged quantity in space, and the visualization of the first four (normalized) POD modes, where the red color indicates a positive value and blue values indicate negative values. We can see oscillatory behavior in the modes as we increase the associated frequency (the singular values of the POD modes). That is, removing the last columns of $\mathbf{U}$ to generate $\mathbf{U_r}$ removes the high-frequency modes of the data representation.

\begin{figure}[ht!]
    \sidecaption
    \centering
    \includegraphics[width = \linewidth]{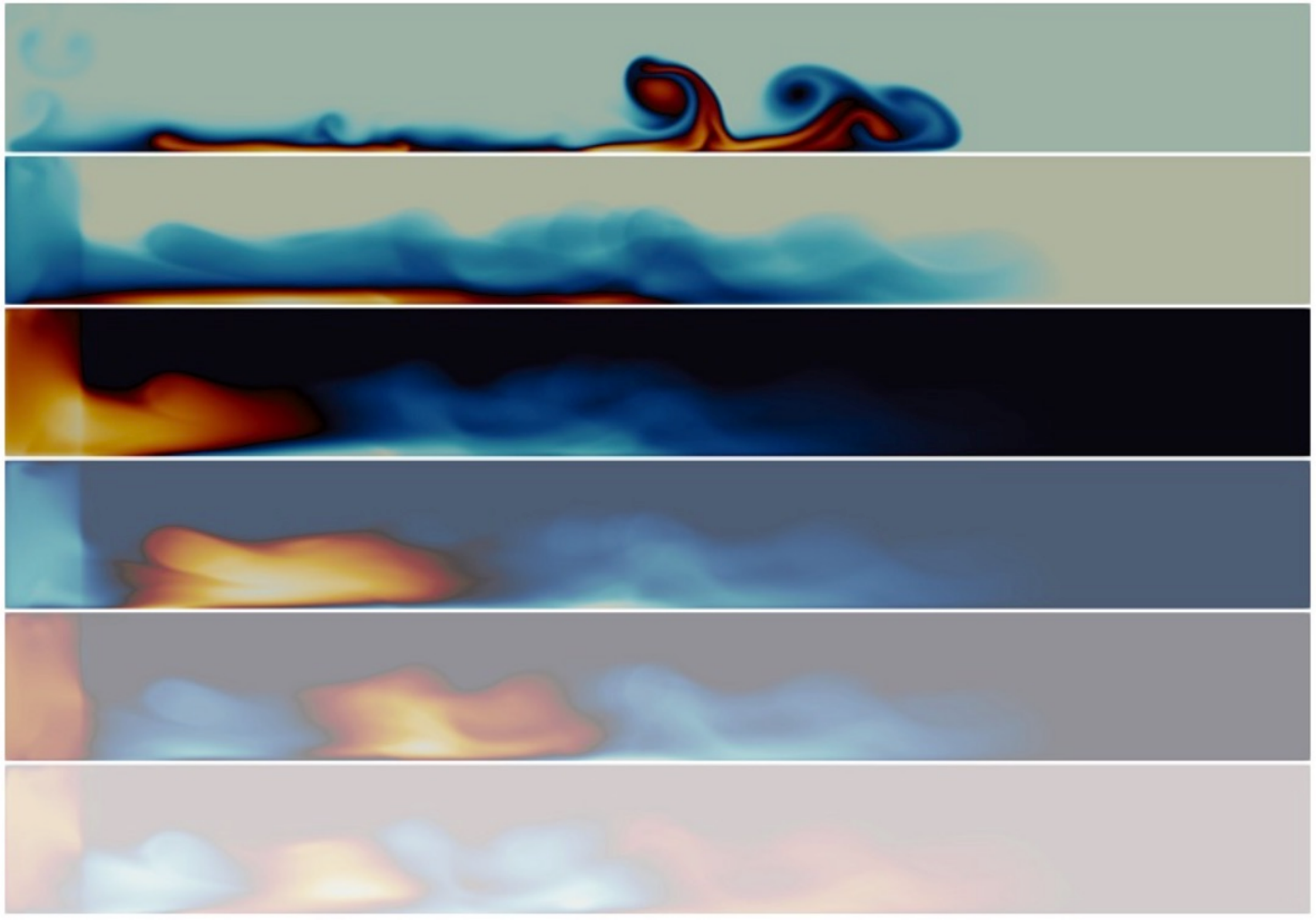}
    \setcounter{figure}{3}
    \caption{Visualization of POD spatial modes for the 2D lock-exchange problem. From top to bottom: a representative snapshot of the concentration field at an intermediate time, the time-averaged concentration (mean field), and the first four normalized POD modes (columns of $\mathbf{U}_r$). Red and blue regions indicate positive and negative mode values, respectively. The progressive increase in spatial oscillation frequency from Mode~1 to Mode~4 reflects the hierarchical structure of the SVD: lower modes capture large-scale coherent structures, while higher modes represent increasingly fine-scale features. Discarding higher-order modes in the truncated SVD thus removes high-frequency spatial content.}
    \label{fig:lock2d_modes}
\end{figure}

To reveal the difference between POD and DMD, we must first describe Reduced Order Models (ROMs) \cite{sreekumar2026snapshots}. This is a class of SciML models that extract coherent structures from high-dimensional, high-fidelity simulations and represent these patterns in a low-dimensional space. The ROM literature began with SVD-based models but later embraced NN-based models, such as Autoencoders, with the advance of deep learning \cite{Velho2025}. One of the most relevant taxonomies of ROMs concerns their intrusiveness on the high-fidelity simulation code. Methods such as POD require a projection step (often the Galerkin method), in which the high-dimensional discrete operator is projected onto the low-dimensional subspace spanned by $\mathbf{U}_r$. That is, POD is categorized as an \textbf{intrusive ROM}, given that it requires access to the same high-fidelity simulation code and environment. On the other hand, methods such as DMD are considered \textbf{non-intrusive ROMs}, given that they are completely data-driven and equation-free. In other words, to have a working POD model, you need access to both the data and the numerical simulation code, while for DMD, you only need access to the high-fidelity data. It is worth noting that this taxonomy is not always sharply defined in practice. Operator Inference (OpInf), for example, learns a reduced operator from data but requires knowledge of the equation structure, placing it somewhere between intrusive and non-intrusive. Similarly, PINNs embed the governing equations directly into the loss function, yet require no access to a simulation code. The boundary therefore depends on what 'prior knowledge' is counted as intrusive. For a comprehensive discussion of this spectrum, the reader is referred to Padula et al. \cite{padula2024} and the references therein.

\par 
With that, building upon this mathematical foundation, Dynamic Mode Decomposition stands out as a prominent data-driven technique for analyzing fluid systems. DMD computes a best-fit linear approximation of the underlying nonlinear operator dictating the temporal evolution of the coupled flow. The exact DMD algorithm is described in Algorithm \ref{algdmd}. By applying SVD to a sequential set of flow snapshots, DMD successfully isolates dynamically significant spatial modes, along with their distinct growth rates and temporal frequencies. This capability to decouple complex spatio-temporal dynamics into discrete, actionable modes provides a robust mathematical framework for constructing computationally efficient, reduced-order surrogate models. Recent advancements \cite{barros2023phd} have expanded the applicability of DMD to more complex computational scenarios, including adaptive mesh refinement and coarsening (AMR/C) \cite{Barros2022} and data compression workflows \cite{Barros2023}.

\begin{algorithm}
    \caption{Reconstruction of dynamical systems using the exact DMD method}
    \begin{algorithmic}
        \STATE \textbf{INPUT:} Snapshots $\mathbf{Y} = \{\mathbf{y}_0, \hdots , \mathbf{y}_{m-1}\}$
        \STATE \textbf{OUTPUT:} Signal reconstruction $\mathbf{\tilde{y}}(t) \approx \mathbf{y}(t)$
        \STATE 1: Set $\mathbf{Y}_1 = \{\mathbf{y}_0, \hdots , \mathbf{y}_{m-2}\}$ and $\mathbf{Y}_2 = \{\mathbf{y}_1, \hdots , \mathbf{y}_{m-1}\}$.
        \STATE 2: Compute the SVD of  $\mathbf{Y}_1$,  $\mathbf{Y}_1 = \mathbf{U\Sigma V}^T$.
        \STATE 3: Define the truncation rank $r$ and compute  $\mathbf{\tilde{A}} \coloneqq \mathbf{U}_r^T \mathbf{Y}_2 \mathbf{V}_r\mathbf{\Sigma}_r^{-1}$.
        \STATE 4: Compute eigenvalues and eigenvectors of $\mathbf{\tilde{A}W = W\Lambda}$.
        \STATE 5: Compute continuous eigenvalues $\mathbf{\Omega}$ from discrete eigenvalues $\mathbf{\Lambda}$ obtained, $\omega_i = \ln(\lambda_i) / \Delta t_i$, where $\Delta t_i = t_{i+1} - t_i$.
        \STATE 6: Compute DMD modes $\mathbf{\Psi}^{DMD} = \mathbf{Y}_2 \mathbf{V}_r \mathbf{\Sigma}^{-1}_r \mathbf{W}$.
        \STATE 7: Compute vector $\mathbf{b}$, $\mathbf{b} = (\mathbf{\Psi}^{DMD})^\dagger \mathbf{y}_0$.
        \STATE 8: Reconstruct approximation $\mathbf{\tilde{y}}(t)$, $\mathbf{\tilde{y}}(t) = \sum_{i = 1}^{k}b_i \psi_i\exp(\omega_i t)$
    \end{algorithmic}
    \label{algdmd}
\end{algorithm}

\par 
\subsubsection*{Fixed Mesh Snapshots}

First, we consider an initial dataset of the simulations performed as described in Section \ref{sec:le_rb}. Starting with the lock-exchange simulation, we perform the SVD on the concentration dataset. Figure \ref{fig:lock2d_sing_values} shows the singular value decay of the dataset. We notice a swift decay of around one order of magnitude for the early modes (approximately $r \leq 25$). After that, the modes still decay rapidly, though with a shallower slope, until around $r=380$. At this point, the singular values are negligible ($\mathcal{O}(10^{-3})$) in comparison with the first singular values. Several values of $r$ are assessed. In practice, selecting $r$ involves balancing three competing considerations: the retained variance $\kappa = \sum_{i=1}^{r}\sigma_{i}^{2}/\sum_{i=1}^{m-1}\sigma_{i}^{2}$, the reconstruction error $\eta$ measured against a held-out snapshot, and the conditioning of the pseudoinverse — since increasing $r$ beyond a threshold raises the spectral radius and introduces instability \cite{heath2018scientific}. A pragmatic approach is to sweep $r$ over a candidate set, evaluate $\eta$ on a validation snapshot, and select the smallest $r$ for which $\eta$ falls below an application-specific tolerance. This is precisely the procedure followed in Fig. \ref{fig:lock2d_sing_values}, where $r \in \{50, 100, 150, 200, 250\}$ are assessed rather than a single value being prescribed \textit{a priori}.

Considering $m$ as the rank of the snapshots matrix, when analyzing the retained variance \cite{Lumley1967, Taira2017}, that is, how much of the information is being retained by the truncated modes, we notice that from $r=50$ to $r=250$ the difference in $\kappa$ is around three orders of magnitude. Although it seems a substantial difference, even for $r = 50$, the discarded information is less than $0.5\%$. Although this metric is not a direct assessment of the quality of the DMD approximation \cite{Taira2017}, it is an adequate \textit{a priori} estimate and we proceed to assess the DMD approximations for all the proposed values of $r$. It is important to note that $\kappa$ measures the fraction of the total variance captured by the retained modes, but this does not directly translate to pointwise or temporal accuracy of the reconstructed fields. A high $\kappa$ indicates that the dominant spatial structures are well represented, but the dynamical accuracy of the DMD reconstruction also depends on the quality of eigenvalue estimation and the temporal coherence of the retained modes. Therefore, $\kappa$ should be interpreted as a necessary but not sufficient condition for model fidelity.

\begin{figure}[ht!]
    \centering
    \includegraphics[width = 0.55\linewidth]{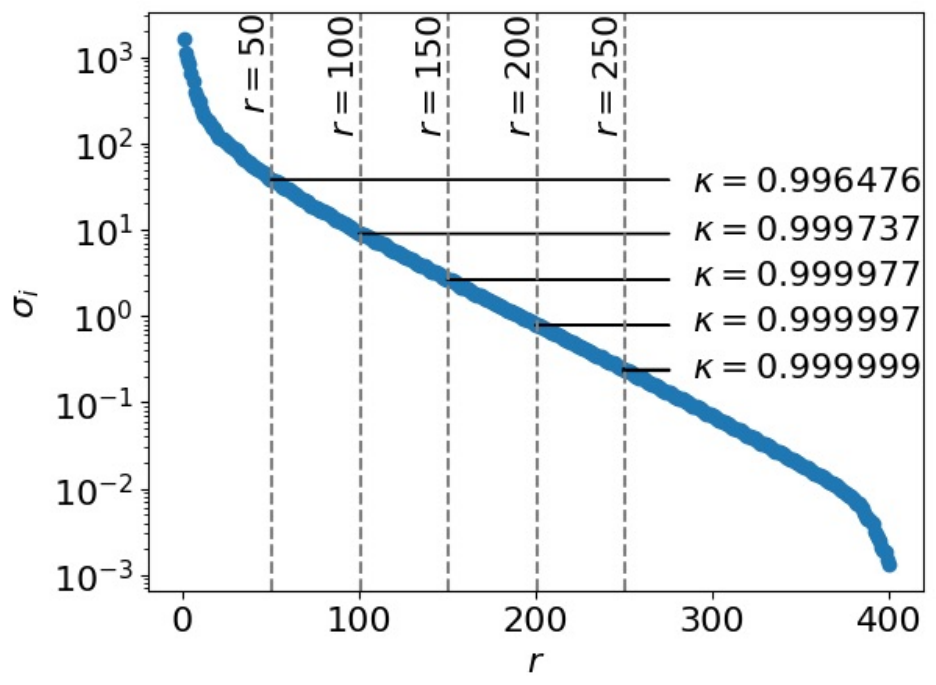}
    \caption{Singular values for the $\mathbf{Y}_1$ matrix containing the 2D lock-exchange snapshots and the values for $\kappa$ for each $r = \{50, 100, 150, 200, 250\}$. Singular values were limited to a maximum of $r = 400$ for proper visualization. The rapid initial decay (roughly one order of magnitude for $r \leq 25$) indicates that most of the flow's energy is concentrated in a small number of dominant modes. The retained energy $\kappa$ quantifies the fraction of total variance captured: values very close to 1 (e.g., $\kappa > 0.995$ for $r \geq 50$) suggest that the truncated basis adequately represents the dominant spatial structures, though dynamical accuracy of the DMD reconstruction must be verified separately (see Fig.~\ref{fig:lock2d_rel_error}). Taken from \cite{Barros2020}.}
    \label{fig:lock2d_sing_values}
\end{figure}

Now we apply DMD to reconstruct the problem's dynamics and compare the results with the ground truth. We use the relative 2-norm such that $\eta = ||\hat{\mathbf{y}} - \mathbf{y}||_2 / ||\mathbf{y}||_2$, being $||\cdot||_2$ the 2-norm of a vector, $\mathbf{y}$ is the ground truth solution vector and $\hat{\mathbf{y}}$ its DMD approximation. Figure \ref{fig:lock2d_rel_error} shows how the error behaves across the simulated time and the evolution of the quantities of interest over time. We observe that the error follows the problem dynamics and that, as we increase $r$, we obtain better approximations of the DMD proxy. It is important to mention that increasing $r$ indiscriminately may improve prediction quality until a certain point, but the associated spectral radius of the pseudoinverse also increases in orders of magnitude, making it ill-conditioned and prone to instability \cite{heath2018scientific}.  Regarding Quantities of Interest (QoIs), we note that the quantities are conserved across all approximations.

\begin{figure}[ht!]
    \sidecaption
    \centering
    \begin{minipage}[t]{0.6\textwidth}
        \vspace{-6.8cm} 
            \begin{subfigure}{\textwidth}
            \centering
            \includegraphics[width = 0.8\linewidth]{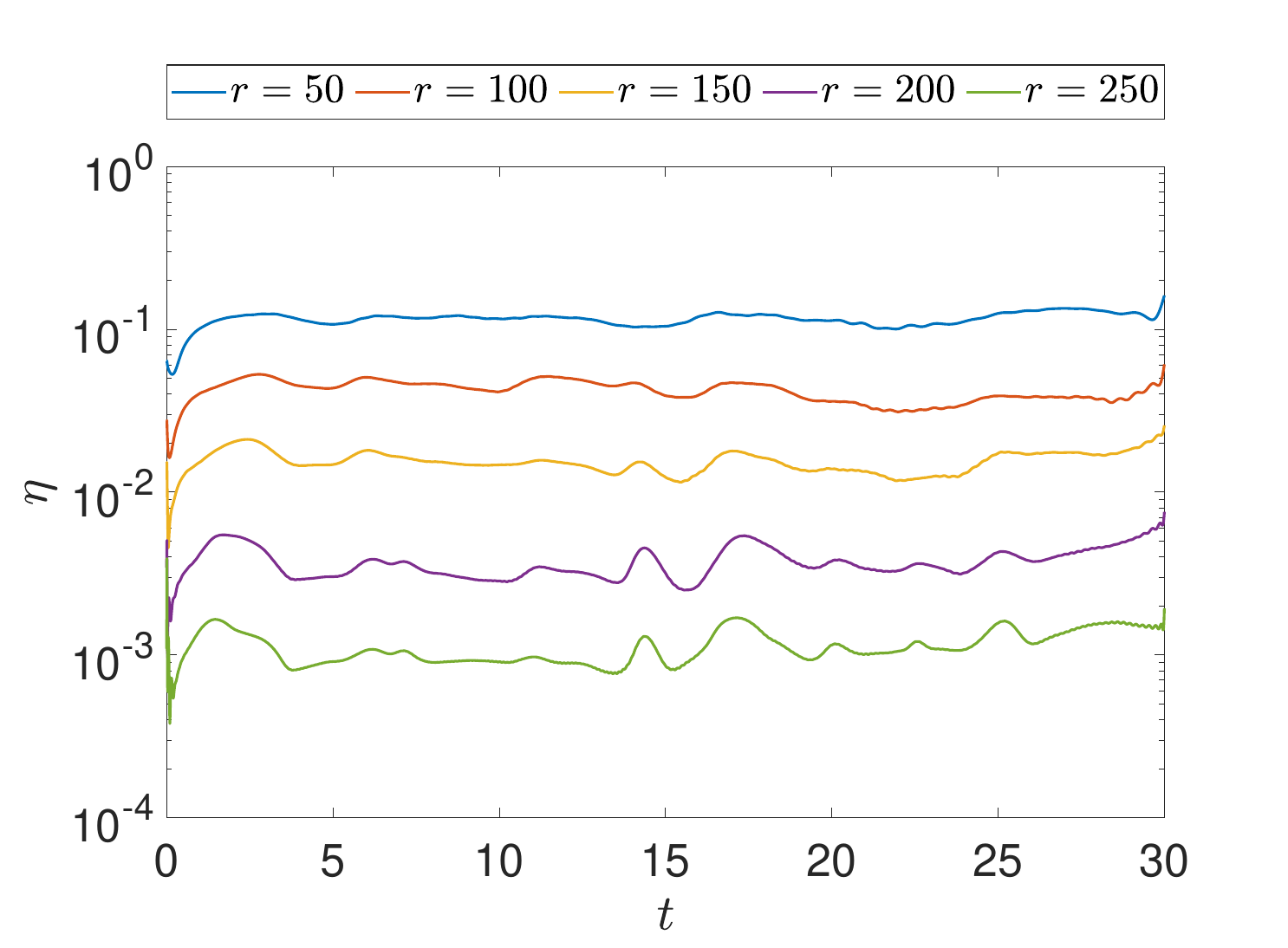}
            \subcaption[]{Relative $2$-norm over time for all approximations.}       
            \end{subfigure} \\
            \begin{subfigure}{\textwidth}
            \centering
            \includegraphics[width = \linewidth]{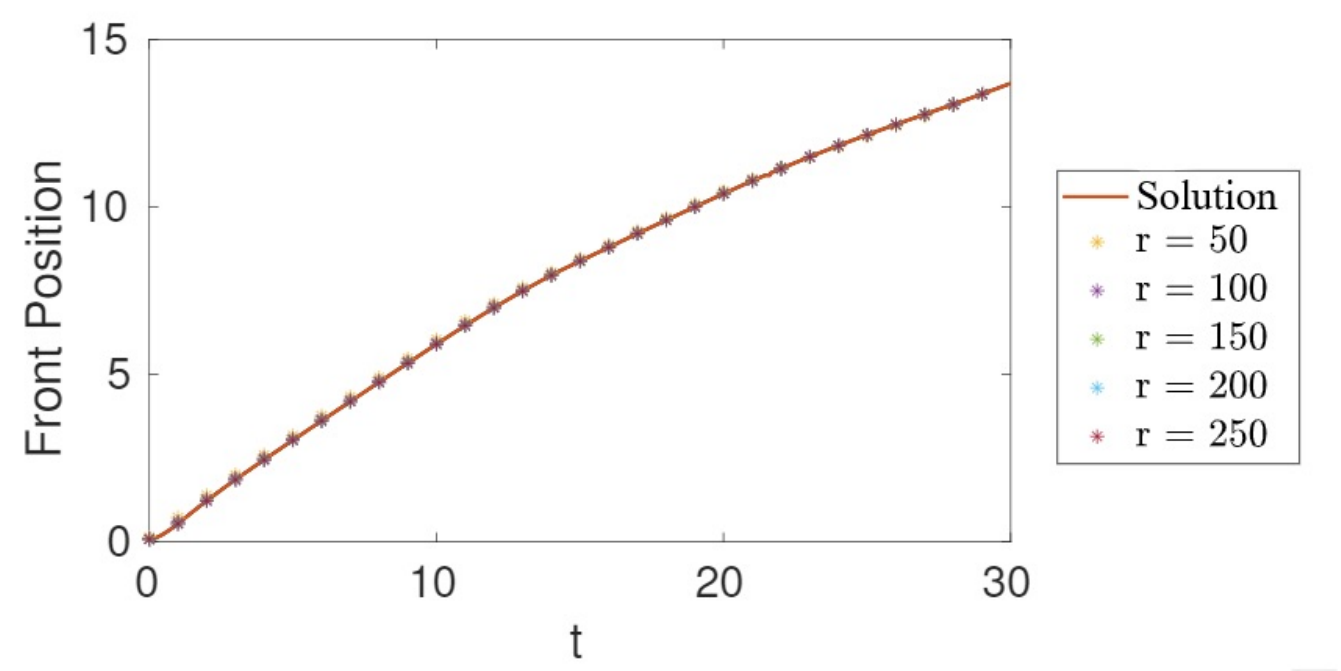}
            \subcaption[]{Quantity of Interest: Front position.}       
            \end{subfigure} \\
            \begin{subfigure}{\textwidth}
            \centering
            \includegraphics[width = \linewidth]{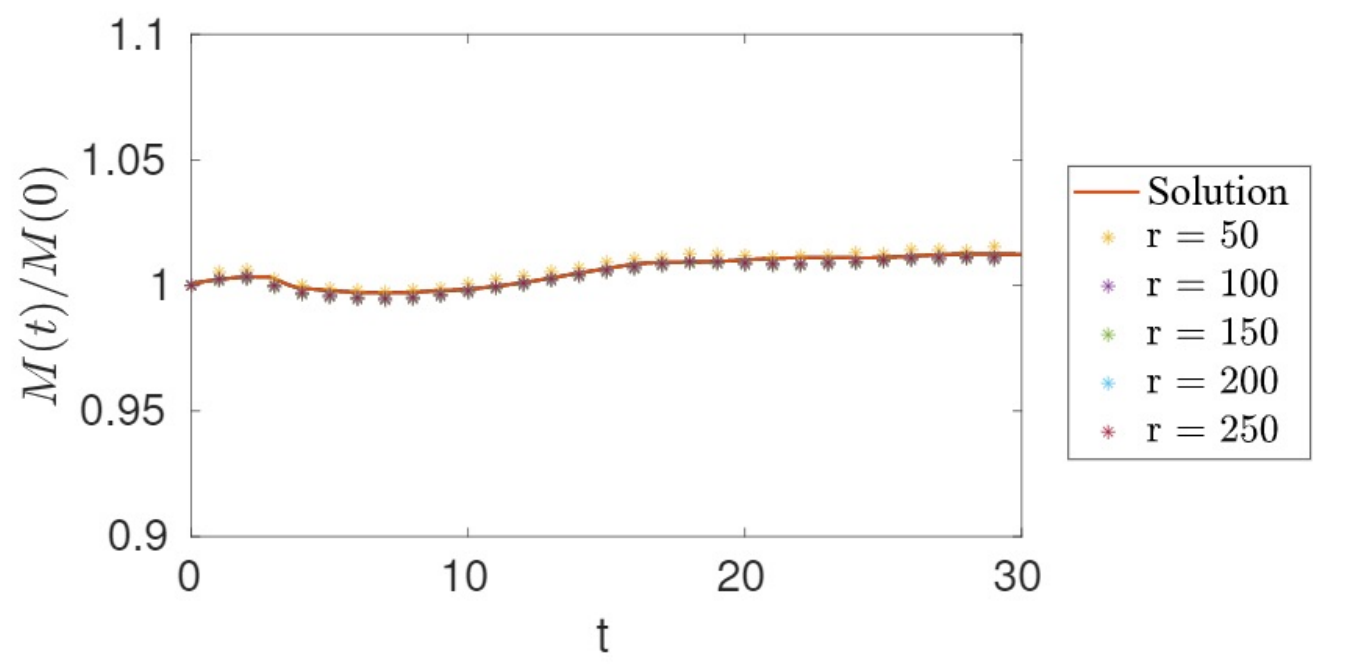}
            \subcaption[]{Quantity of Interest: Mass conservation.}       
            \end{subfigure} 
    \end{minipage}
    \setcounter{figure}{5}
    \caption{DMD reconstruction performance for the 2D lock-exchange problem at different truncation ranks $r$. (a) Temporal evolution of the relative 2-norm error $\eta$: the error tracks the physical dynamics, peaking during highly transient phases and decreasing as the flow stabilizes. Increasing $r$ consistently reduces the reconstruction error. (b--c) Quantities of interest (front position and sediment mass) are well conserved across all approximations, confirming the physical consistency of the DMD reconstruction. Taken from \cite{Barros2020}}
    \label{fig:lock2d_rel_error}
\end{figure}

\par 
For the Rayleigh-Bénard instability, we perform the same steps. First, we perform the SVD decomposition, and we analyze the singular value decay for the temperature dataset, as illustrated in Figure \ref{fig:sing_val_rb}. We observe that for $Ra = 10^{10}$, the singular values exhibit a slow decay, indicating that the flow energy is not concentrated in a few dominant modes. Conversely, the $Ra = 10^6$ case displays a significantly more rapid decay. Evaluating the number of modes required to capture a sufficient amount of the retained variance \cite{Peherstorfer2018SurveyOptimization}, we note that the case where $Ra = 10^{10}$ requires approximately $100$ modes, whereas the $Ra = 10^6$ case requires only around $10$ modes. Finally, regarding the low-rank approximation $\mathbf{\Tilde{X}}$ of $\mathbf{X}$, we observe that the $Ra = 10^{10}$ case requires nearly the full rank of $\mathbf{X}$ to achieve Frobenius relative errors ($\eta_F = ||\hat{\mathbf{X}} - \mathbf{X}||_2 / ||\mathbf{X}||_2$ below $10^{-2}$, a threshold considered acceptable for this application. In contrast, for $Ra = 10^6$, this accuracy is achieved with a truncation rank of $r=50$. With that, we observe that the singular-value decay of the data is undesirable for linear models, making DMD an unsuitable choice for this purpose. We revisit this example in Section \ref{sec:betavaes}, noting that a more robust, nonlinear ROM might be a better choice for this data. Until then, for the remainder of this chapter, we cover results for the lock-exchange simulation only.

\begin{figure}[ht!]
    \centering
    \includegraphics[width=0.99\linewidth]{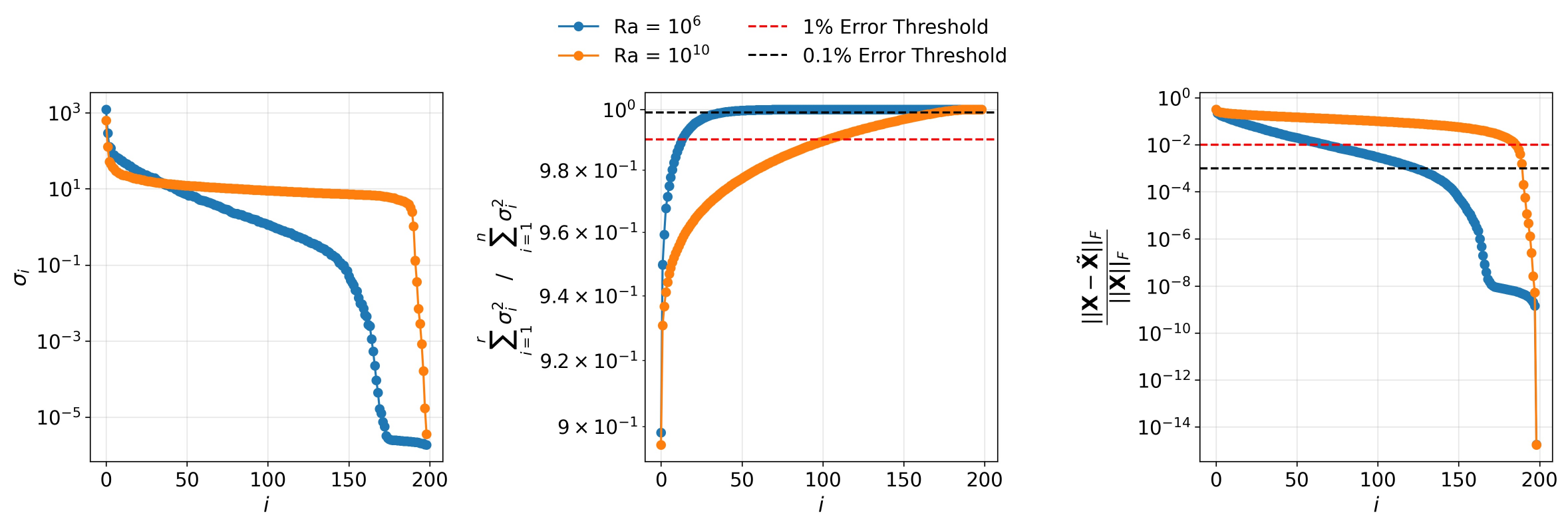}
    \setcounter{figure}{6}
    \caption{Singular values, $\kappa$ and relative Frobenius norm between the truncated approximation and the ground truth data for the Rayleigh-Bénard dataset.}
    \label{fig:sing_val_rb}
\end{figure}

\subsubsection*{AMR/C Snapshots} 
It is natural to use adaptivity in numerical simulations to reduce the computational burden, and this is true for both spatial (fewer equations in the nonlinear systems) and temporal (fewer time steps evaluated) adaptive schemes. Traditional DMD relies on snapshots possessing the same spatial dimensionality. However, in highly dynamic simulations using AMR/C, spatial adaptivity results in computational meshes that change over time, rendering standard snapshot matrices incompatible. For instance, see Fig. \ref{fig:amrcdmd}, which shows a snapshot of a turbidity current and the mesh used at this time step. Here, an adaptive mesh scheme is used to track sediment concentration gradients and refine the mesh in the specific region of interest for the problem. As the current moves, the mesh updates its topology and number of elements and nodes. With that, it is not possible to create the snapshot matrix for the SVD factorization.

\begin{figure}[ht!]
    \sidecaption[t]
    \centering
    \begin{minipage}[t]{0.6\textwidth}
        \vspace{-3.2cm} 
            \begin{subfigure}{\textwidth}
            \centering
            \includegraphics[trim={0 0 0 5.5cm}, clip, width=0.99\linewidth]{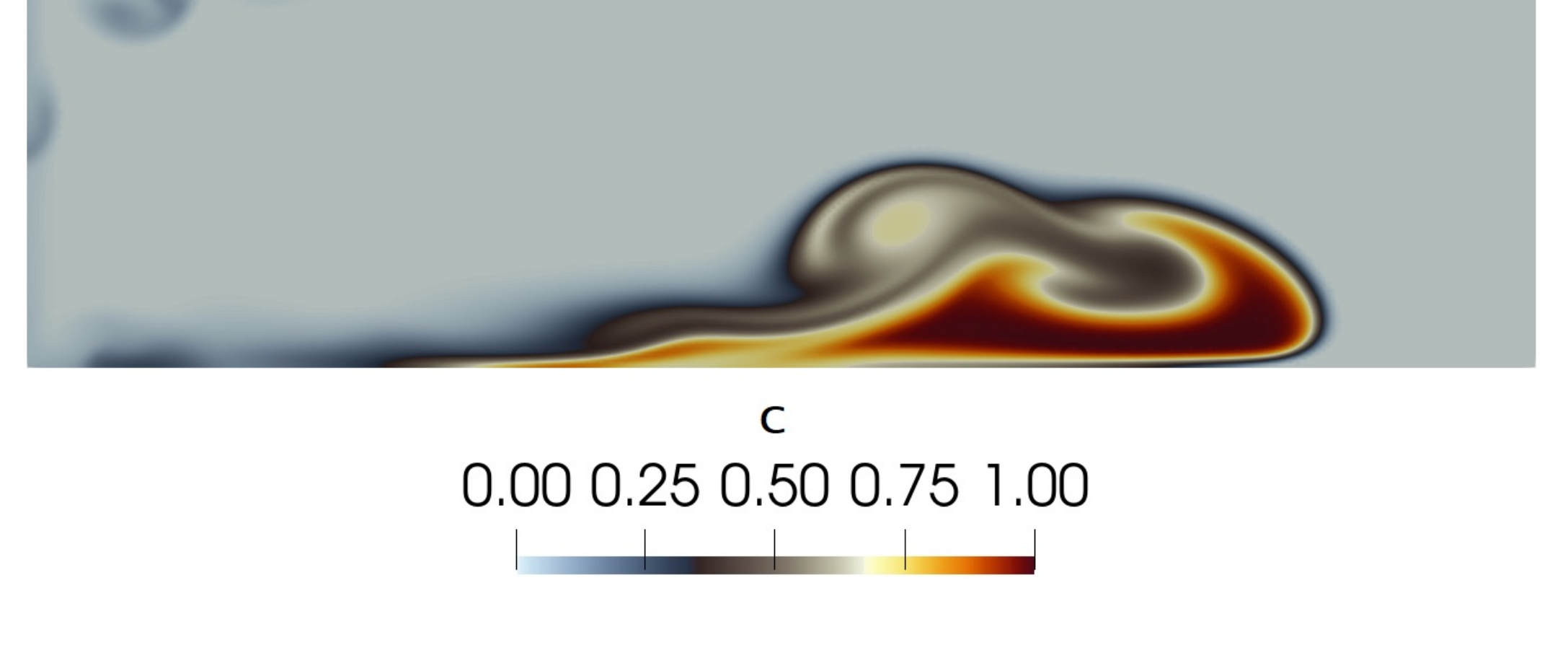}
            \subcaption[]{Adaptive mesh solution.}       
            \end{subfigure} \\
            \begin{subfigure}{\textwidth}
            \centering
            \includegraphics[trim={0 8cm 0 5.5cm}, clip, width=0.99\linewidth]{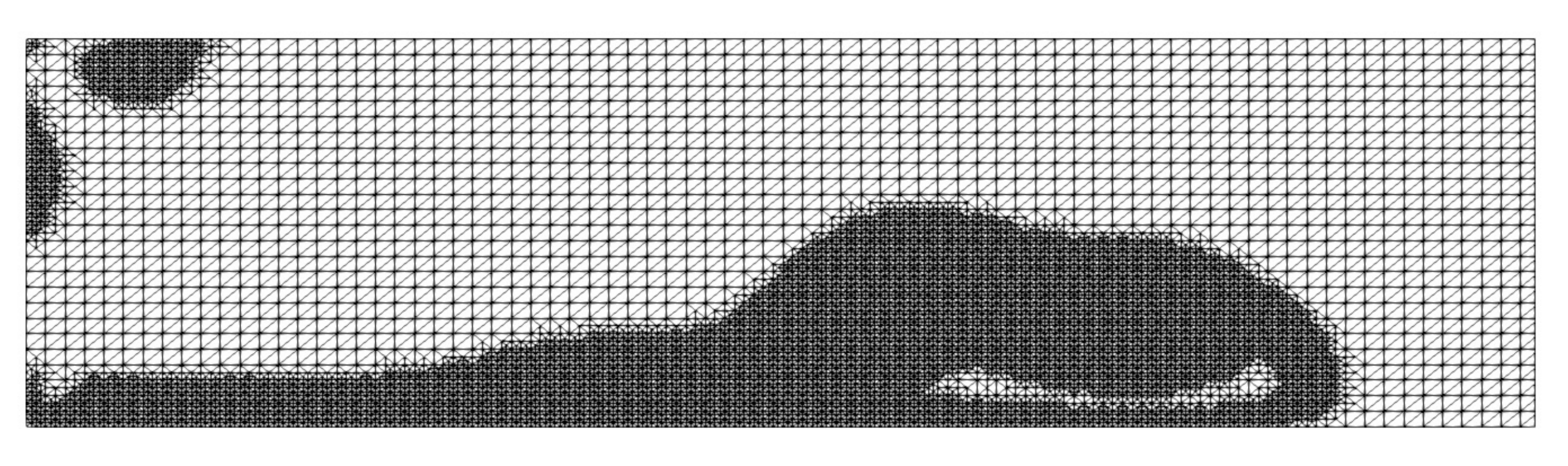}
            \subcaption[]{Adaptive mesh.}       
            \end{subfigure}
    \end{minipage}
    \setcounter{figure}{7}
    \caption{Results and mesh for the first 8 length units of the domain at $t = 10$ time units, from \cite{Barros2022}.}
    \label{fig:amrcdmd}
\end{figure}

To overcome this, a robust strategy involves projecting adaptive snapshots onto a common reference function space using an $\mathcal{L}^{2}$-projection. This enables SVD-based snapshot methods to extract coherent structures from observations with varying mesh topologies. In this case, a suitable reference mesh must be defined to properly capture the problem. Then, considering that the concentration field is the target variable for the projection, the following problem is solved:

\begin{equation}
    (\mathbf{c}^h - \mathbf{c}^{p},\mathbf{v})_{\mathcal{L}^2} = 0 \quad \forall \mathbf{v} \in V^{\psi},
\end{equation}
\noindent where $\mathbf{c}^h$ is the discrete vector for the concentration, $\mathbf{c}^{p}$ is the projected (and desired) solution and $\mathbf{v}$ is a test function residing in function space $V^{\psi}$, which is the same used in the variational formulation for the trial function. With that, results become identical to the ones posed in Fig. \ref{fig:lock2d_rel_error}, given that the AMR/C simulations are identical to the fixed mesh solutions and the $\mathcal{L}^{2}$-projection, when done in adequate meshes, leads to negligible errors. Further discussion of this matter can be found in \cite{Barros2022}.

\subsubsection*{Compressed Snapshots}

A key practical consideration when deploying SVD-based SciML methods in large scale computational workflows is data management. The quality of the reduced-order model depends directly on the fidelity of the snapshot data, yet high-fidelity simulations generate enormous volumes of floating-point data that strain storage and I/O systems. This subsection examines whether lossy data compression, which inevitably introduces small perturbations to the stored fields, is compatible with the SVD-based DMD workflow presented above. The goal is to demonstrate that, under appropriate compression tolerances, the dominant spectral structure of the data is preserved, and the resulting DMD surrogate remains accurate. This finding has direct implications for SciML practitioners who must balance data fidelity against storage constraints in production environments.

Another issue when generating high-dimensional snapshots for these complex problems is the severe storage and I/O constraints. It is not uncommon to see floating point compression strategies in highly intensive computing tasks \cite{cappello2019use, di2025survey}. In this section, we discuss integrating data compression techniques directly into the DMD workflow to relieve disk pressure, as discussed in \cite{Barros2023}. With the increased interest in data compression techniques for machine learning (see, for instance, Google's latest TurboQuant\footnote{https://research.google/blog/turboquant-redefining-ai-efficiency-with-extreme-compression/} algorithm \cite{zandieh2026turboquant}), we go through our results in this section in more detail.

In modern scientific workflows, the enormous volume of data generated by high-fidelity simulations creates severe storage and input/output bottlenecks. Lossy data compression has emerged as an essential strategy to mitigate these constraints, allowing researchers to significantly reduce file sizes while retaining the critical physical features required for downstream analysis \cite{cappello2019use, di2025survey}. The ZFP compression algorithm \cite{lindstrom2014fixed} is specifically engineered to handle multi-dimensional arrays of floating-point data prevalent in scientific computing. Instead of relying on traditional dictionary methods, ZFP operates by partitioning the spatial domain into small, independent blocks. Within each block, the floating-point values are aligned to a common exponent, transforming them into a block-floating-point representation. A custom orthogonal transform is then applied to decorrelate the spatial data, converting the physical field values into a set of transform coefficients. Because neighboring data points in physical simulations are often highly correlated, most of the signal energy is concentrated in a few low-frequency coefficients. ZFP encodes these coefficients using an embedded coding scheme that prioritizes the most significant bits, providing granular control over the trade-off between compression rate and required data fidelity. To achieve this granular control, ZFP offers four distinct operational modes tailored to different scientific requirements. 
\begin{itemize}
    \item The fixed-rate mode allows the user to specify an exact number of compressed bits per block, guaranteeing a predictable compressed file size and facilitating read and write access patterns.
    \item The fixed-precision mode dictates the number of uncompressed bits used to represent each floating-point value, effectively controlling the relative error.
    \item  For applications demanding strict error bounds, the fixed-accuracy mode specifies an absolute error tolerance, ensuring that the maximum difference between the original and reconstructed data never exceeds a prescribed threshold.
    \item Finally, the expert mode provides complete control over the underlying algorithm parameters, allowing advanced users to fine-tune the minimum and maximum bit allocations, precision levels, and exponent thresholds for highly specialized compression tasks.
\end{itemize}

For this example, exceptionally, we add a third spatial dimension to our lock-exchange model. Figure \ref{fig:scheme3d} shows the 3D tank and a snapshot of the turbidity current. We made this problem 3D so that the effects of data compression could be better explored. The new domain is a $20 \times 2 \times 0.1$ block, which consists of sediment columns with $S = 0.75$. We use a hexahedral structured mesh with a 0.025 grid spacing. The Navier–Stokes and sediment concentration systems of linear equations are solved in 16 cores using Block-Jacobi + GMRES(35) with local  ILU(0) preconditioning. GMRES tolerance is  $10^{-6}$. For the nonlinear solver, the tolerance is $10^{-3}$. The time step size is $0.001$, and the simulation runs until the maximum time of 20 time units is reached. XDMF/HDF5 raw data files are written every $10$ time steps, producing 2,000 snapshots. Each file stores the variables $u$, $v$, $w$, and $p$ from the Navier-Stokes equations, the variable $c$, and the deposition, $d$, that is, the integral over time of the deposition rate, $r=u_sc_b$, where $c_b$ is the sediment concentration at the lower boundary. Mesh data is stored in a single HDF5 file. 

\begin{figure}[ht!]
    \sidecaption[t]
    \centering
    \begin{minipage}[t]{0.99\textwidth}
        \vspace{0cm} 
            \begin{subfigure}{\textwidth}
            \centering
            \includegraphics[width=0.9\linewidth]{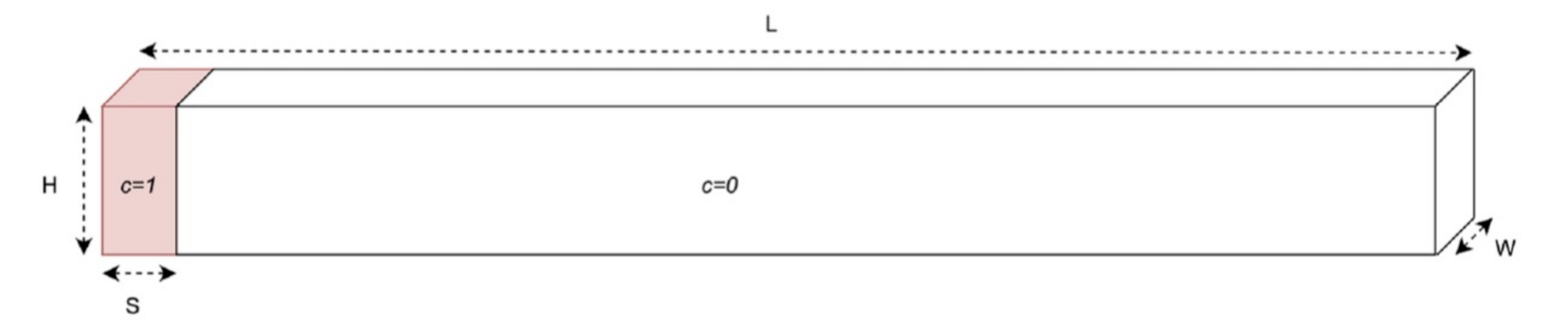}
            \subcaption[]{Geometry and initial conditions for the 3D lock-exchange simulation, showing the sediment column (orange) within the rectangular tank.}
            \end{subfigure} \\
            \begin{subfigure}{\textwidth}
            \centering
            \includegraphics[width=0.9\linewidth]{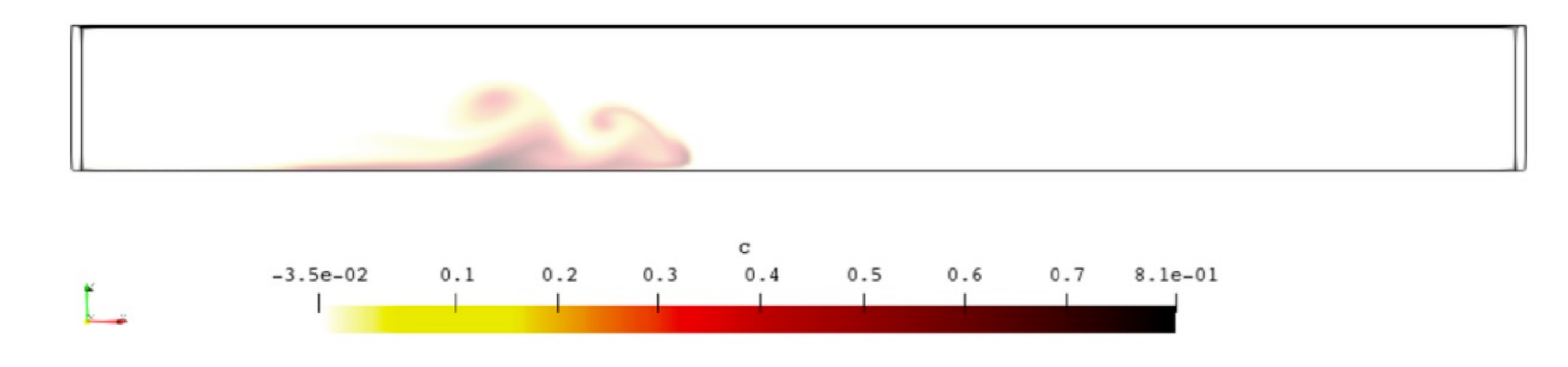}
            \subcaption[]{Sediment concentration volume rendering at $t=10$ time units, showing the propagating gravity current and the development of three-dimensional mixing structures.}
            \end{subfigure}
    \end{minipage}
    \setcounter{figure}{8}
    \caption{Geometry and initial conditions for the 3D lock-exchange simulation (top) and Sediment concentration snapshot at t = 10 time units (bottom). The figure, from \cite{Barros2023}, shows the sediment concentration 3D volume rendering.}    
    \label{fig:scheme3d}
\end{figure}

As a test, we assess how data is affected under three different data compression strategies: a lossless compression (that is, where data is reversible in its full integrity) and two lossy compression formats: using fixed-accuracy mode on ZFP, we compress the simulation data preserving accuracies of $10^{-6}$ and $10^{-3}$. In this case, the larger the threshold, more information is lost during the compression and more space is saved in disk. Using lossy compression algorithms, such as ZFP, achieves compression rates (CRs) of up to two orders of magnitude for this kind of scientific data, according to Table \ref{tab:data_comp}.

\begin{table}\centering
\caption{Storage requirements (in MBytes) for each quantity of interest and the compression rates for the lock-exchange simulation, taken from \cite{Barros2023}.}\label{tab:data_comp}
\small
\begin{tabular}{lcccccccccccccc}\toprule
&\multicolumn{2}{c}{$u$} &\multicolumn{2}{c}{$v$} &\multicolumn{2}{c}{$w$} &\multicolumn{2}{c}{$p$} &\multicolumn{2}{c}{$c$} &\multicolumn{2}{c}{$d$} \\\cmidrule{2-13}
&Storage &CR &Storage &CR &Storage &CR &Storage &CR &Storage &CR &Storage &CR \\\midrule
Raw Data &1384.9 & &2737.3 & &2737.3 & &2737.3 & &2737.3 & &2737.3 & \\
ZFP Lossless &1275.5 &1.1 &2480.8 &1.1 &2534.7 &1.1 &2348.9 &1.2 &2631.0 &1.04 &194.3 &14.1 \\
ZFP $10^{-6}$ &295.4 &4.7 &142.9 &19.2 &556.7 &4.9 &654.6 &4.2 &189.9 &14.41 &35.0 &78.2 \\
ZFP $10^{-3}$ &120.9 &11.5 &12.3 &221.7 &222.8 &12.3 &323.3 &8.5 &90.7 &30.18 &23.4 &116.8 \\
\bottomrule
\end{tabular}
\end{table}

\begin{figure}[ht!]
    \centering
    \includegraphics[width=0.59\linewidth]{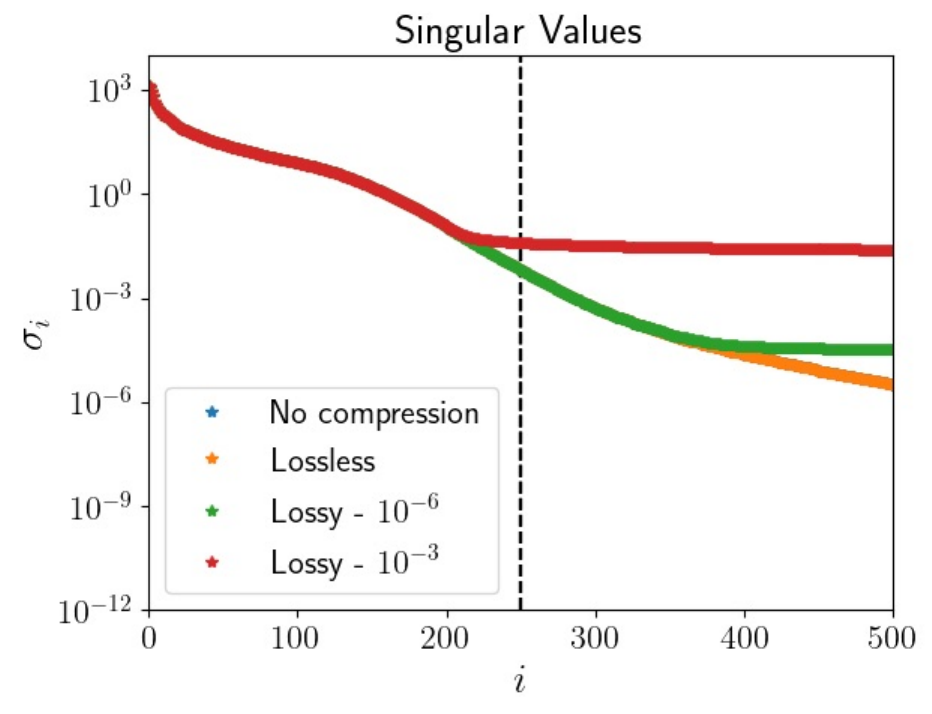}
    \caption{Singular values for the tested cases. The dashed line represents the number of $r = 250$ singular values and vectors selected for the DMD basis. The key observation is that the lossless and high-fidelity lossy ($10^{-6}$) compressed datasets closely track the uncompressed baseline throughout the retained spectrum. The more aggressive lossy compression ($10^{-3}$) introduces a plateau in the singular values that begins before the truncation threshold, indicating that some physically relevant modal information has been altered by the compression. This spectral analysis provides a practical diagnostic for assessing whether a given compression level is compatible with a target truncation rank. From \cite{Barros2023}.}
    \label{fig:singval_necker}
\end{figure}

In the context of mass transport and turbidity currents, this compression strategy proves highly effective for processing concentration snapshots with CR of order $O(10^2)$. The scalar concentration field typically exhibits regions of smooth variations intertwined with localized, sharp moving fronts. When ZFP acts on these snapshots, it efficiently compresses the smooth regions where spatial correlation is highest. For blocks containing sharp concentration gradients, the bounded error tolerances ensure that crucial physical features are accurately preserved. By directly compressing the floating-point arrays before disk storage, severe input and output bottlenecks are mitigated. Now, by applying SVD to the four datasets, we can extract interesting insights, as illustrated in Fig. \ref{fig:singval_necker}. This figure contrasts the singular value decay for the uncompressed baseline data against datasets subjected to lossless compression and lossy compression at two distinct accuracy tolerances. For the subsequent solution reconstruction, the basis is truncated to retain the $r = 250$ most energetically dominant dynamic modes.The snapshot matrices stemming from the uncompressed and lossless compression simulations exhibit perfect agreement, confirming that the structural integrity of the data is entirely preserved. Conversely, the spectral decay for the lossy compressed datasets eventually flattens into a plateau. When configured with a fixed-accuracy tolerance of $10^{-6}$, the compression algorithm maintains the vast majority of the leading singular values, closely tracking the uncompressed baseline. Because the dashed line in the figure denotes the truncation threshold utilized for the DMD basis construction, the data discarded by this specific lossy compression configuration does not alter the reduced-order model. The deviations in the singular spectrum occur strictly beyond the designated truncation rank. In contrast, imposing a more relaxed accuracy tolerance of $10^{-3}$ results in an aggressive data reduction that perturbs the singular spectrum well before the truncation cutoff. This modification directly alters the dominant singular values retained for the DMD basis, inherently impacting the physical information available for the surrogate reconstruction.

The next step is to assess the $\mathbf{\tilde{A}}$ eigenvalues spectrum. To simplify the discussion, we refer to the following nomenclature:

\begin{itemize}
    \item Case A: Raw data;
    \item Case B: Lossless compression;
    \item Case C: Lossy compression with a fixed-accuracy of $10^{-6}$;
    \item Case D: Lossy compression with a fixed-accuracy of $10^{-3}$.
\end{itemize}

Figure \ref{fig:lock3d_eigenvalues} illustrates these eigenvalues mapped onto the complex plane. A detailed view of the entire eigenvalue spectrum for cases A, B and C is provided in Figures \ref{fig:lock3d_eig_250} and \ref{fig:lock3d_eig_250_zoom}, whereas Figures \ref{fig:lock3d_eig_lossy} and \ref{fig:lock3d_eig_lossy_zoom} contrast the results between cases A and D. Because the underlying snapshot matrices consist entirely of real-valued physical data, all extracted eigenvalues naturally emerge as complex conjugate pairs \cite{Krake2021}. The spectral distributions for Cases A, B, and C are nearly identical, suggesting that the application of lossless or high-fidelity lossy compression preserves the essential flow dynamics. In contrast, the comparison between Cases A and D highlights that adopting a more aggressive lossy compression configuration fundamentally shifts the resulting eigenvalue locations. However, relying exclusively on the visual distribution of DMD eigenvalues across the complex plane can obscure the analysis by displaying redundant modal components. It is crucial to emphasize that determining the physical significance of individual DMD modes and their paired eigenvalues is not a trivial procedure when depending solely on the exact DMD formulation and standard complex plane plots \cite{Krake2021}.

\begin{figure}[ht!]
	\begin{center}
            \begin{subfigure}{0.45\textwidth}
            \centering
            \includegraphics[width=\linewidth]{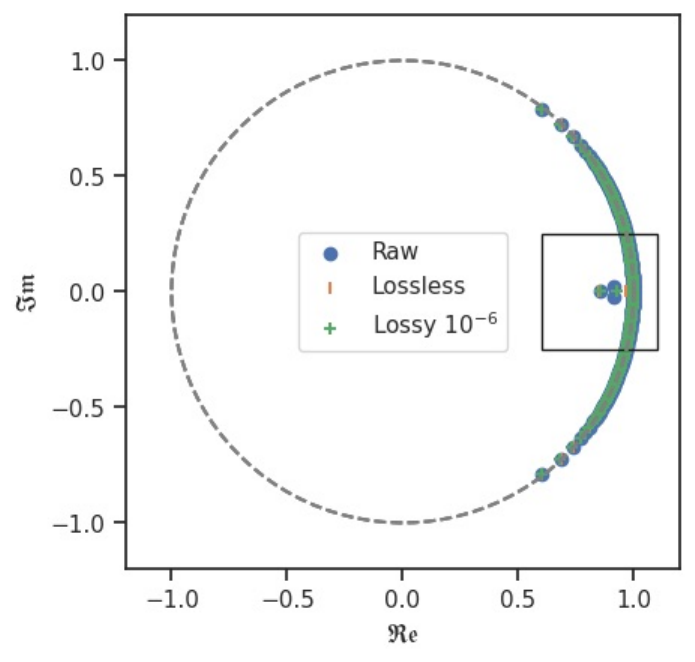} 
            \subcaption[]{Eigenvalues of $\mathbf{\Tilde{A}}$.}      
            \label{fig:lock3d_eig_250}
            \end{subfigure} 
            \begin{subfigure}{0.45\textwidth}
            \includegraphics[width=\linewidth]{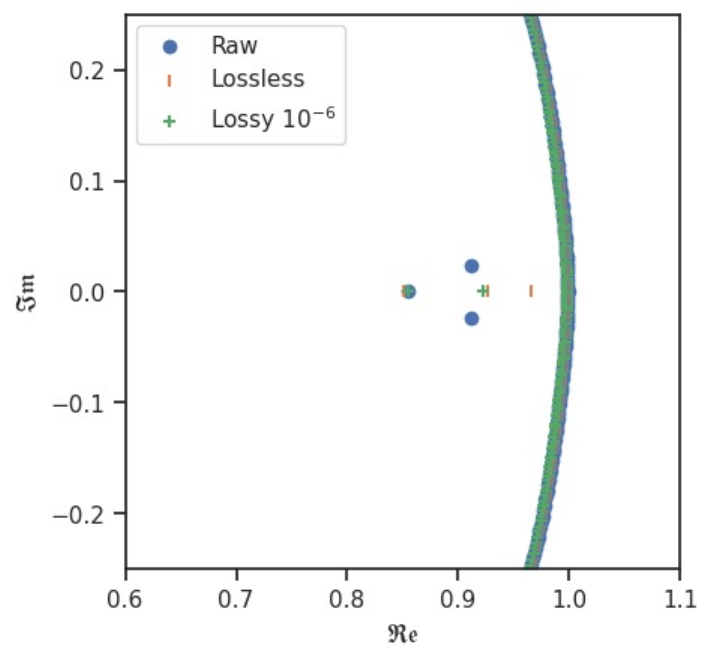} 
            \subcaption[]{Zoom at rectangular region of interest.}
            \label{fig:lock3d_eig_250_zoom}
            \end{subfigure} \\
            \begin{subfigure}{0.45\textwidth}
            \centering
            \includegraphics[width=\linewidth]{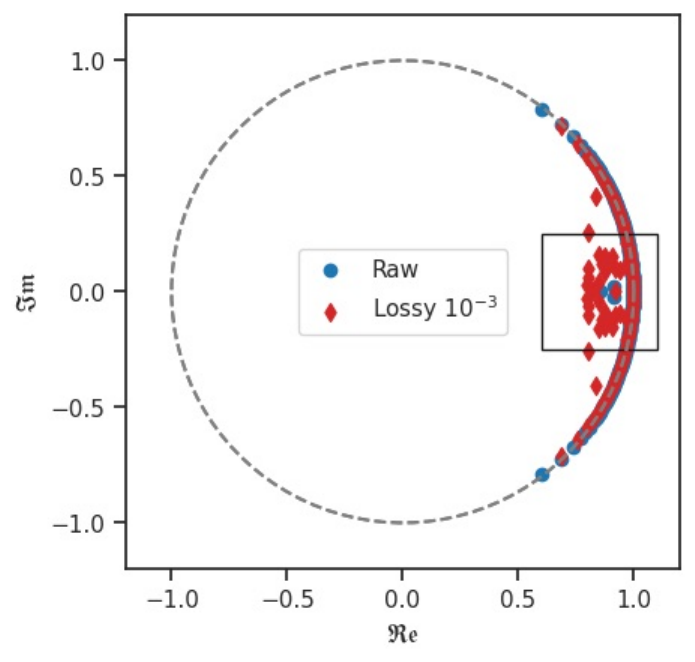} 
            \subcaption[]{Eigenvalues of $\mathbf{\Tilde{A}}$.}   
            \label{fig:lock3d_eig_lossy}
            \end{subfigure} 
            \begin{subfigure}{0.45\textwidth}
            \includegraphics[width=\linewidth]{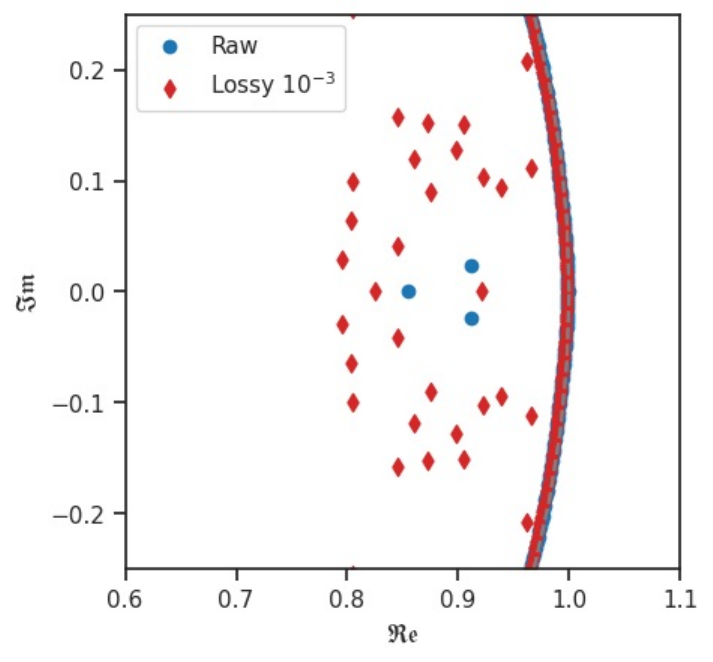} 
            \subcaption[]{Zoom at the rectangular region of interest.}       
            \label{fig:lock3d_eig_lossy_zoom}
            \end{subfigure}
		\caption{DMD eigenvalues for the 3D lock-exchange simulation plotted in the complex plane. (a--b) Comparison of Cases A (raw), B (lossless), and C (lossy $10^{-6}$), showing near-identical spectral distributions. (c--d) Comparison of Cases A and D (lossy $10^{-3}$), showing significant eigenvalue displacement due to aggressive compression. The right panels zoom into the boxed region near the unit circle. Eigenvalues appear as complex conjugate pairs because the snapshot data is real-valued. The proximity of eigenvalues to the unit circle indicates nearly neutral stability of the corresponding modes.}
		\label{fig:lock3d_eigenvalues}
	\end{center}
\end{figure}

Finally, we assess the DMD approximations using various compression strategies with the ground truth (raw data). An examination of Figure \ref{fig:rel_error_necker} reveals that the magnitude of the reconstruction error remains within the same order across all evaluated compression schemes. As anticipated, the baseline uncompressed data and the lossless compression approach yield completely indistinguishable error profiles. Regarding the lossy compression configurations, the relative discrepancies exhibit a marginal increase during the final phases of the simulated timeframe. During the initial flow development up to $t < 8$s, the approximation utilizing a lossy tolerance of $10^{-6}$ produces outcomes that closely mirror the uncompressed baseline. Furthermore, the temporal evolution of the relative error closely tracks the underlying physical dynamics of the system. Specifically, the peak error magnitudes coincide with periods of highly transient and rapid flow variations, such as around $t = 2$s, contrasting with the lower error levels observed as the flow regime stabilizes later in the simulation. Overall, the error trajectories demonstrate consistent behavioral trends across all four tested configurations, confirming that the integration of these data reduction strategies preserves the fundamental physical accuracy of the reduced-order model.

\begin{figure}[ht!]
    \sidecaption
    \centering
    \begin{minipage}[t]{0.6\textwidth}
        \vspace{-6.1cm} 
            \begin{subfigure}{\textwidth}
            \centering
            \includegraphics[width = 0.8\linewidth]{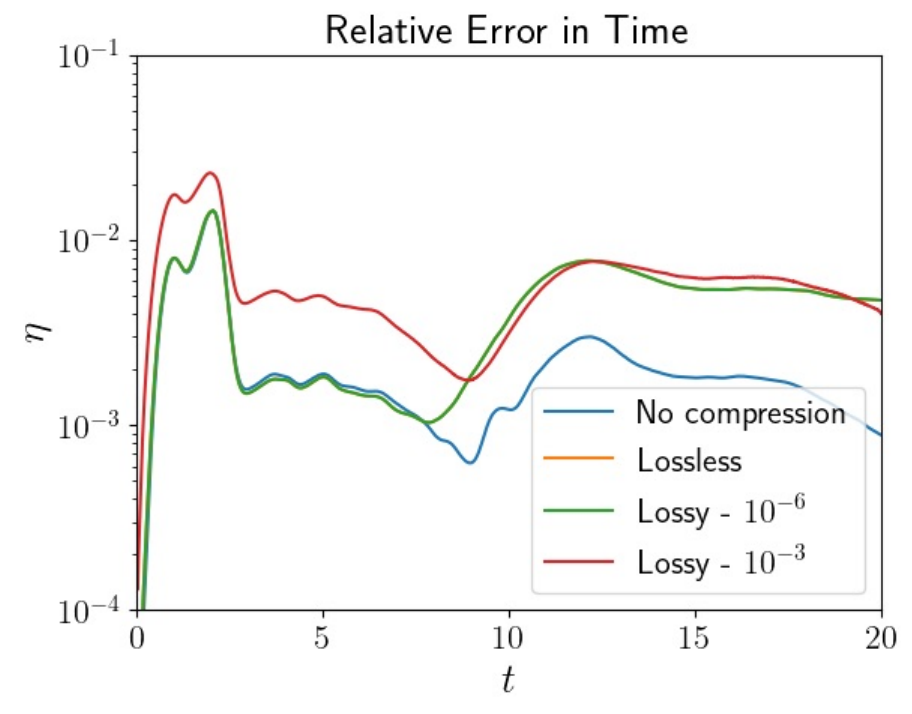}
            \subcaption[]{Relative error in time for the solution reconstruction.}       
            \end{subfigure} \\
            \begin{subfigure}{\textwidth}
            \centering
            \includegraphics[width = \linewidth]{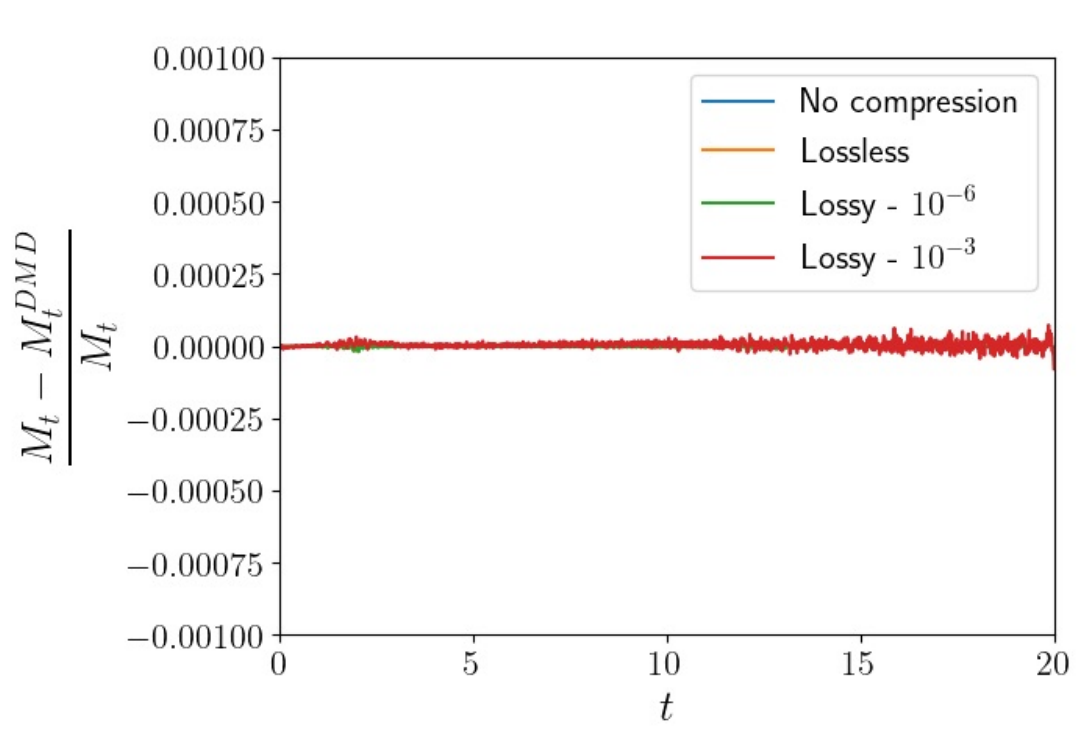}
            \subcaption[]{Time history of the relative sediment mass error for all cases. For this problem, $M_t = \int_\Omega c d\Omega$ and the superscript $DMD$ refers to the same expression but taking the DMD reconstruction of $c$.}       
            \end{subfigure} 
    \end{minipage}
    \setcounter{figure}{11}
    \caption{Reconstruction error and QoIs between DMD approximations using compressed data and ground truth (raw data). Taken from \cite{Barros2023}}
    \label{fig:rel_error_necker}
\end{figure}

\paragraph*{\textbf{Section Summary}}

This section demonstrated that SVD-based methods, particularly DMD, provide effective reduced-order surrogates for coupled fluid flow problems when the singular-value spectrum decays sufficiently fast. For the lock-exchange benchmark, DMD accurately reconstructed the spatio-temporal dynamics with modest truncation ranks, and this performance was maintained under adaptive mesh refinement and moderate levels of lossy data compression. However, for the highly turbulent Rayleigh-Bénard case at $Ra = 10^{10}$, the slow singular-value decay revealed the inherent limitation of linear dimensionality reduction for complex, multi-scale flows. This motivates the neural network-based methods discussed in the following section, which can learn nonlinear manifolds and address regimes where linear approaches are insufficient.


\subsection{Neural Network-Based Methods}
\label{sec:nn}

While SVD-based methods provide efficient and interpretable reduced-order models for problems with rapidly decaying singular-value spectra, their linear nature limits their applicability in highly nonlinear or multi-scale regimes, as demonstrated by the Rayleigh-Bénard case at $Ra = 10^{10}$ in the previous section. To overcome these limitations, we now turn to neural network-based approaches, which can learn nonlinear mappings and capture complex dynamical features that lie outside the span of any linear subspace.

Deep learning architectures offer considerable flexibility in modeling the severe nonlinearities inherent in the Navier-Stokes and scalar transport equations. These neural network-based methods can be tailored to perform nonlinear dimensionality reduction, solve inverse problems, or construct highly parameterized surrogate models.




\subsubsection{Physics-Informed Neural Networks for Density-driven Gravity Flows}
\label{sec:pinns}

Following the neural-network-based reduced-order and surrogate strategies discussed so far, we now consider a different class of models in which the governing equations are enforced directly during training. In this setting, Physics-Informed Neural Networks (PINNs) \cite{raissi2019physics, Cuomo2022, PSAROS2023111902, oberainotes} are particularly attractive for density-driven gravity flows because they combine sparse measurements with the physical constraints \cite{romulo2023phd}. This makes them suitable for inverse settings in which only partial observations of the flow are available, while the goal remains the reconstruction of the full spatio-temporal evolution of velocity, pressure, and concentration fields.

In general, PINNs approximate the unknown solution of a PDE with a neural network and train it by penalizing the residual of the governing equations at a set of collocation points. If $\hat{u}(\mathbf{x};\mathbf{\theta})$ denotes the network approximation of a field $u(\mathbf{x})$, and if the problem is written as
\begin{equation}
	\mathcal{N}[u](\mathbf{x}) = 0 \;\; \text{in } \Omega,
	\qquad
	\mathcal{B}[u](\mathbf{x}) = 0 \;\; \text{on } \partial\Omega,
\end{equation}
\noindent then the corresponding PDE and boundary residuals are
\begin{equation}
	r_{\Omega}(\mathbf{x};\mathbf{\theta}) = \mathcal{N}[\hat{u}](\mathbf{x}),
	\qquad
	r_{\partial\Omega}(\mathbf{x};\mathbf{\theta}) = \mathcal{B}[\hat{u}](\mathbf{x}).
\end{equation}
These quantities are evaluated through automatic differentiation, allowing the spatial and temporal derivatives of the network output to be computed directly with respect to the input. The training objective is then constructed as a weighted combination of physics and data terms,
\begin{equation}
    \label{eq:weights}
	\mathcal{L}(\mathbf{\theta}) =
	w_{\Omega}\mathcal{L}_{\Omega}(\mathbf{\theta}) +
	w_{\partial\Omega}\mathcal{L}_{\partial\Omega}(\mathbf{\theta}) +
	w_{d}\mathcal{L}_{d}(\mathbf{\theta}),
\end{equation}
\noindent where $\mathcal{L}_{\Omega}$ penalizes the PDE residual at interior collocation points, $\mathcal{L}_{\partial\Omega}$ enforces the boundary conditions, and $\mathcal{L}_{d}$ measures mismatch with available observations, weighted by $w_{\Omega}$, $w_{\partial\Omega}$ and $w_{d}$, respectively. In this sense, PINNs may be interpreted as residual-based surrogate models that combine collocation and data assimilation within a single optimization problem.

In practice, however, PINN training is often sensitive to the relative scale of these loss components. In coupled systems, some residual terms may decrease much faster than others, potentially biasing the optimization toward the easiest constraints. A common remedy for this multi-objective optimization problem is to use dynamic loss weights, in which the coefficients $w_{\Omega}$, $w_{\partial\Omega}$, and $w_{d}$ in Eq.~\eqref{eq:weights} are updated during training based on gradient-based statistics \cite{relobralo, gradnorm, softadapt}, so that the data, boundary, and PDE terms remain more balanced throughout optimization. For example, the ReLoBRaLo strategy~\cite{relobralo} adjusts each weight based on the relative rate of change of its corresponding loss component, while SoftAdapt~\cite{softadapt} uses the exponentially weighted moving average of the loss histories to balance the terms. A concrete illustration of how these strategies affect convergence in the context of the gravity-current reconstruction considered here is provided in Fig.~\ref{fig:loss_err_dynweights}. This is especially relevant for multi-physics problems such as density-driven gravity flows, where momentum, continuity, and transport residuals may have markedly different magnitudes.
The second key ingredient is the choice of collocation points used to evaluate $\mathcal{L}_{\Omega}$. Uniformly fixed points may undersample sharp fronts and localized flow structures, whereas adaptive or resampled strategies \cite{Wu2023, McClenny2022} improve the coverage of dynamically relevant regions. In particular, Monte Carlo (MC) sampling periodically redraws collocation points over the domain, Residual-based Adaptive Refinement (RAR) enriches the training set with points associated with large residuals, and hybrid MC+RAR strategies combine global exploration with local refinement. These sampling procedures are well-suited to density-driven gravity currents, where steep concentration gradients and interfacial instabilities evolve in space and time.

\begin{figure}[!ht]
	\centering
		\resizebox{\textwidth}{!}{
			\includegraphics[width=\linewidth]{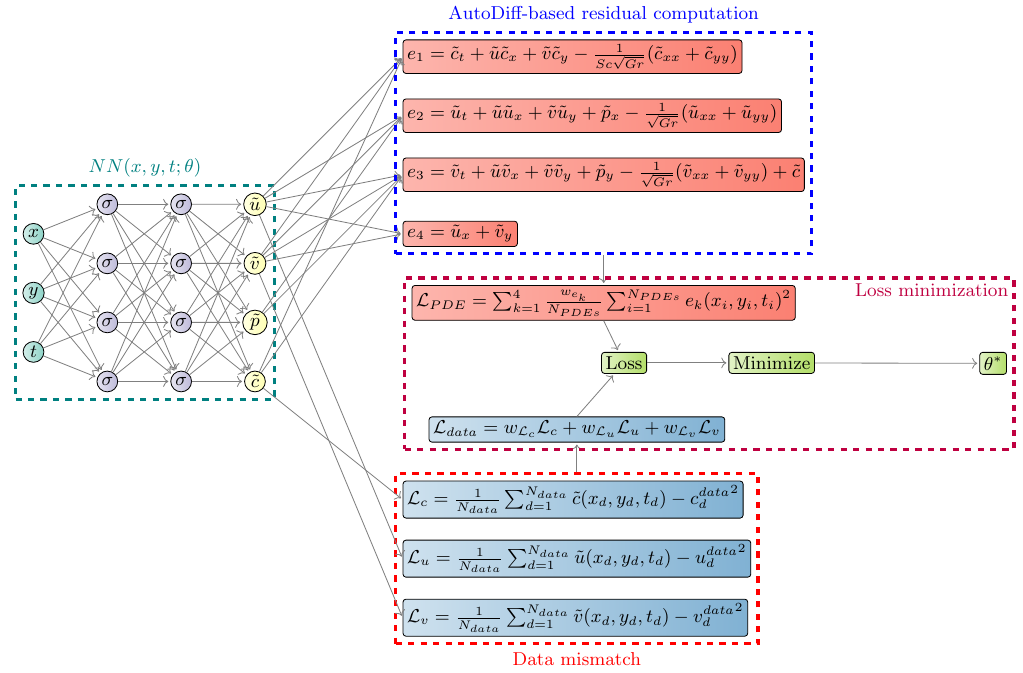}}
	\caption{Schematic of a PINN used to solve a 2D density-driven gravity flow problem. The neural network $NN(x,y,t;\mathbf{\theta})$ takes spatio-temporal coordinates as input and outputs approximations of the velocity components ($\tilde{u}$, $\tilde{v}$), pressure ($\tilde{p}$), and concentration ($\tilde{c}$). The PDE residuals $\{e_k\}_{k=1}^{4}$ corresponding to the continuity, momentum, and transport equations (Eqs.~\eqref{eq:mass_pdgf}--\eqref{eq:transport_pdgf}) are computed via automatic differentiation and penalized at collocation points. For this problem, the settling velocity $u_s = 0$. The loss function $\mathcal{L}$ is the weighted sum of the PDE loss evaluated at the collocation points and the data mismatch, with optional $L_2$-regularization for parameter estimation tasks.}	\label{fig:pinn_scheme_ddgf}
\end{figure}

Figure \ref{fig:pinn_scheme_ddgf} summarizes the reconstruction strategy adopted for sediment transport. A neural network $NN(x, y, t; \mathbf{\theta})$ approximates the velocity components, pressure, and concentration fields, while the residuals of Eqs. \eqref{eq:mass_pdgf}-\eqref{eq:transport_pdgf} are incorporated into the loss function through $\left\{e_k \right\}_{k=1}^{4}$. Available measurements, which in this study are mostly concentration data and, in some cases, velocity data, are added as data-misfit terms. Training then consists of minimizing this combined objective with respect to $\mathbf{\theta}$. In Figure \ref{fig:pinn_scheme_ddgf}, the network outputs are denoted by $\tilde{u} = u_{\mathbf{\theta}}$, $\tilde{v} = v_{\mathbf{\theta}}$, $\tilde{p} = p_{\mathbf{\theta}}$, and $\tilde{c} = c_{\mathbf{\theta}}$.

We first address the reconstruction of concentration and velocity fields from scattered concentration measurements distributed over the spatio-temporal domain. For the velocity components, we impose no-slip boundary conditions exactly by modifying the output of the final layer of the neural network \cite{raissi2019physics}. This modification takes the form

\begin{eqnarray}
	\tilde{u}(\mathbf{x}, t; \mathbf{\theta}) =  g_u(\mathbf{x}, t) + l_u(\mathbf{x}) \mathcal{N}_{u}(\mathbf{x}, t; \mathbf{\theta}), \\
	\tilde{v}(\mathbf{x}, t; \mathbf{\theta}) =  g_v(\mathbf{x}, t) + l_v(\mathbf{x}) \mathcal{N}_{v}(\mathbf{x}, t; \mathbf{\theta})
\end{eqnarray}

\noindent where $\mathcal{N}_{u}(\mathbf{x}, t; \mathbf{\theta})$ and $\mathcal{N}_{v}(\mathbf{x}, t; \mathbf{\theta})$ are the predictions for the $x-$ and $y-$velocity components that come from the neural network, $g_u(\mathbf{x}, t)$ and $g_v(\mathbf{x}, t)$ denote the known boundary conditions for both velocity components, $\tilde{u}(\mathbf{x}, t; \mathbf{\theta})$ and $\tilde{v}(\mathbf{x}, t; \mathbf{\theta})$ are the final approximations for both velocity components, and $l(\mathbf{x})$ is a known function such that \cite{Lu2021_hc}:

\begin{equation}
	\begin{cases}
		l(\mathbf{x}) = 0 & \text{ if } x \in \Gamma, \\
		l(\mathbf{x}) > 0 & \text{ if } x \in \Omega.
	\end{cases}
	\label{eq:l_bd}
\end{equation}

Since the geometry of our 2D problems simply consists of a tank with dimensions $L_x$ and $L_y$, the functions $l_u(\mathbf{x})$ and $l_v(\mathbf{x})$ with $\mathbf{x} = (x, y)$ are chosen to be

\begin{eqnarray}
	l_u(x,y) = \sin{\frac{\pi x}{L_x}} \sin{\frac{\pi y}{L_y}}\\ \label{eq:lu_choice}
	l_v(x,y) = \sin{\frac{\pi y}{L_y}} \label{eq:lv_choice}
\end{eqnarray}

The experiments are performed on three datasets in which the available information consists only of concentration measurements randomly selected across the full spatio-temporal domain. The number of points in each dataset is listed in Table \ref{tab:datasets2d}.

\begin{table}[!ht]
	\centering
	\caption{Dataset identification}
	\begin{tabular}{@{}ll@{}}
		\toprule
		Dataset \#ID & Number of Measurements ($N_{data}$) \\ \midrule
		1            & 15,625                              \\
		2            & 250,000                             \\
		3            & 1,000,000                           \\ \bottomrule
	\end{tabular}
	\label{tab:datasets2d}
\end{table}

Table \ref{tab:hyper_params_2d_hypo} summarizes the hyperparameters adopted in these experiments. The PINN is trained with the MC/random sampling strategy described in \cite{Wu2023}. Multiple runs are performed for each dataset size in order to assess the sensitivity of the results to network initialization.

\begin{table}[!ht]
	\centering
	\caption{Hyper-parameters for the Hypothetical Inverse 2D problem.}
	\begin{tabular}{@{}ll@{}}
		\toprule
		Hyper Parameters                      & Value/Description \\ \midrule
		\# of Hidden Layers                   & 10                \\
		\# of Neurons                         & 200               \\
		Optimizer                             & Adam              \\
		Learning Rate ($\eta$)                 & $0.0008$          \\
		\# of Sampling points ($N_{\mathcal{F}}$)         & $20,000$          \\
		\# of points in the test dataset      & $20,000$          \\
		\# of runs performed for each dataset & $10$              \\
		Activation Function                   & (adaptive) Swish             \\ \bottomrule
	\end{tabular}
	\label{tab:hyper_params_2d_hypo}
\end{table}

\begin{figure}[!ht]
	\centering
	\includegraphics[width=\textwidth]{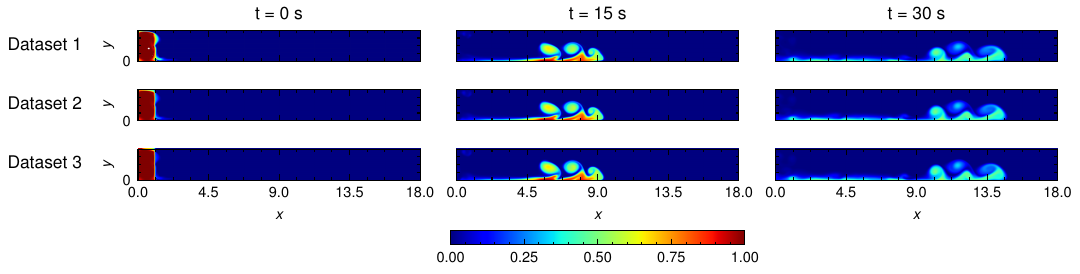}
	\caption{Mean predicted concentration at three different times and datasets.}
	\label{fig:mean_c_pred}
\end{figure}

\begin{figure}[!ht]
	\centering
	\includegraphics[width=\textwidth]{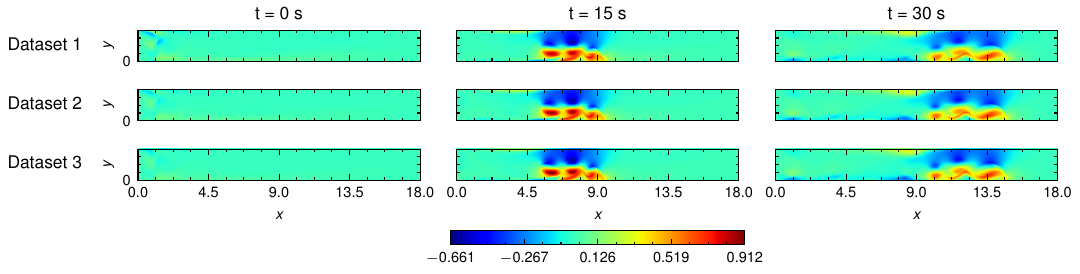}
	\caption{Mean predicted $x-$velocity at three different times and datasets.}
	\label{fig:mean_u_pred}
\end{figure}

\begin{figure}[!ht]
	\centering
	\includegraphics[width=\textwidth]{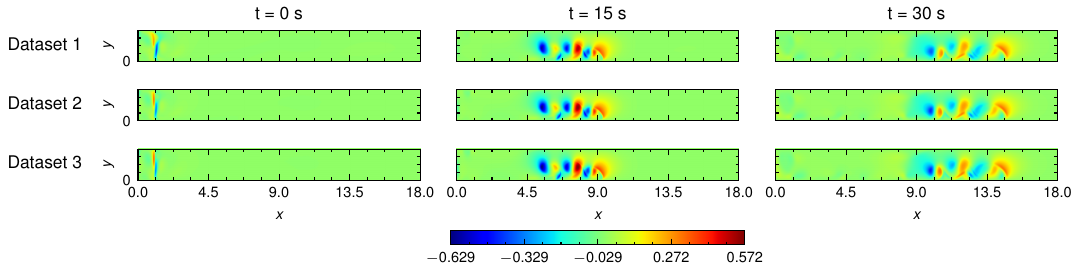}
	\caption{Mean predicted $y-$velocity at three different times and datasets.}
	\label{fig:mean_v_pred}
\end{figure}

\begin{figure}[!ht]
	\centering
	\includegraphics[width=\textwidth]{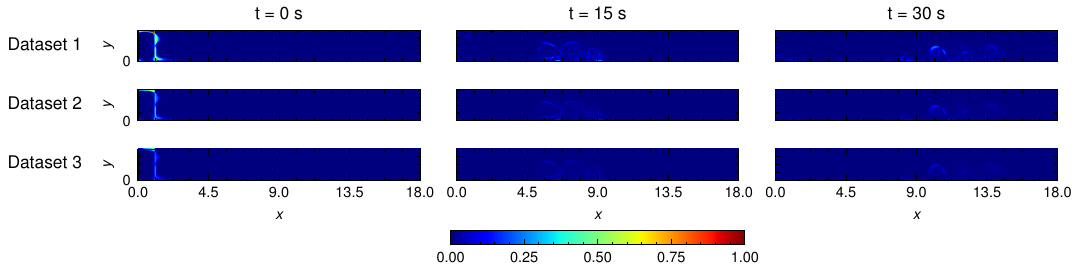}
	\caption{Mean pointwise error for the concentration at three different times and datasets.}
	\label{fig:mean_c_pointwise}
\end{figure}

\begin{figure}[!ht]
	\centering
	\includegraphics[width=\textwidth]{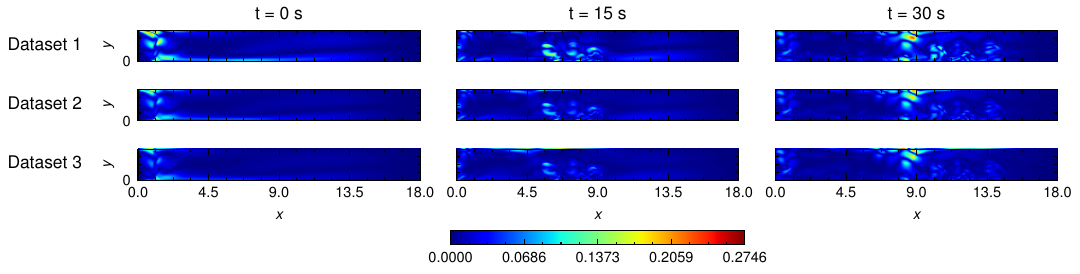}
	\caption{Mean pointwise error for the $x-$velocity at three different times and datasets.}
	\label{fig:mean_u_pointwise}
\end{figure}

\begin{figure}[!ht]
	\centering
	\includegraphics[width=\textwidth]{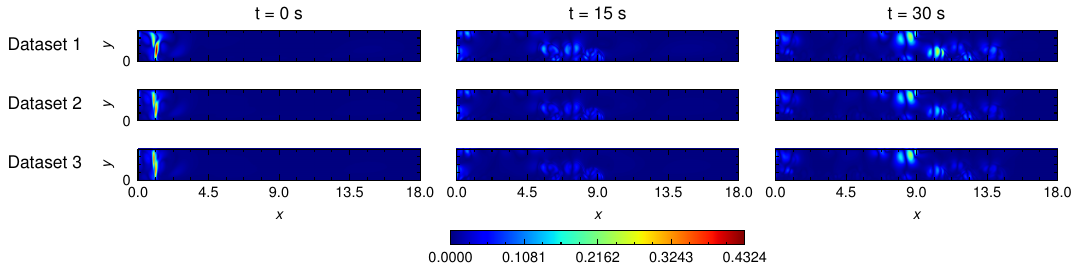}
	\caption{Mean pointwise error for the $y-$velocity at 3 different times and datasets.}
	\label{fig:mean_v_pointwise}
\end{figure}

\begin{figure}[ht!]
    \centering
    \includegraphics[width=\textwidth]{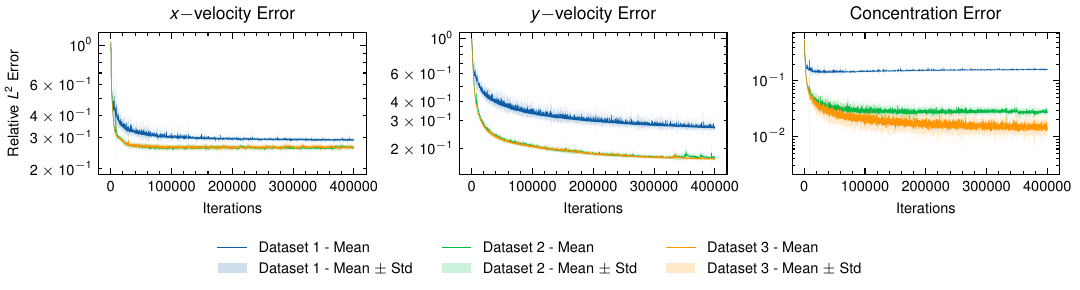}
    \caption{$L^2$ error for multiple runs using the datasets in Table \ref{tab:datasets2d}.}
    \label{fig:l2_err_2d_hypo}
\end{figure}

Figures \ref{fig:mean_c_pred}, \ref{fig:mean_u_pred}, and \ref{fig:mean_v_pred} show the mean predictions for concentration and for the $x$- and $y$-velocity components, while Figures \ref{fig:mean_c_pointwise}, \ref{fig:mean_u_pointwise}, and \ref{fig:mean_v_pointwise} report the corresponding pointwise errors.
From these results, Datasets 2 and 3 provide very similar accuracy. This finding is also confirmed by Fig. \ref{fig:l2_err_2d_hypo}. For the current configuration, $250{,} 000$-concentration measurements already provide sufficient information for the PINN to recover the main flow structures. There is limited additional benefit from increasing the dataset to $1{,}000{,}000$ points.
Building on these findings, even without explicitly provided initial conditions, the PINN recovers the initial state reasonably well, with the largest errors concentrated near the interface between the two fluids at $t=0$. Among the two velocity components, the $x$-velocity is consistently reconstructed more accurately. This is relevant because the streamwise component largely determines how far and how fast the sediment front propagates.

\begin{figure}[!ht]
	\centering
	\includegraphics[width=300pt]{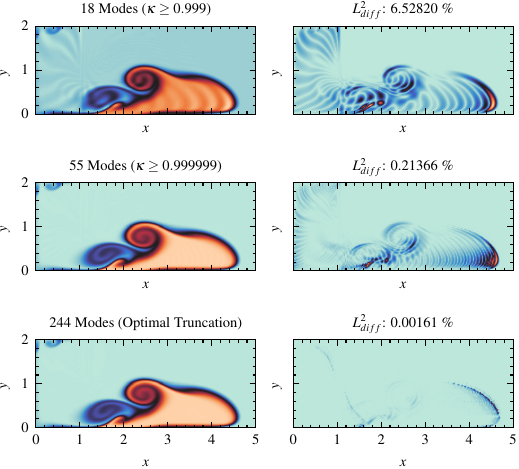}
	\caption{SVD-reconstruction of the concentration using different mode-truncation criteria. Snapshots at $t=6s$. }
	\label{fig:svd_decomp}
\end{figure}

In the sequel, we consider a coupled flow over a small space-time window, where sparse measurements of concentration and velocity are available.
To reduce measurement noise in this smaller observation window, we use the SVD-based filtering strategy already introduced in Section \ref{sec:svd}. Here, the snapshots matrix $\mathbf{X}$ is assembled from concentration or velocity measurements as mentioned above. The filtered reconstruction is obtained by truncating the least informative singular modes. Instead of selecting this truncation visually or by a prescribed explained-variance criterion, we adopt the optimal hard-threshold strategy proposed by \cite{githubGitHubErichsonoptht} and available online\footnote{https://github.com/erichson/optht}.  Figure \ref{fig:svd_decomp} shows the reconstruction (left) of the solution using a given number of modes, and the point-wise difference between the reconstructed and original solution (right) at $t=6s$, with $\kappa$ being the retained variance, and $L^{2}_{diff}$ being the relative 2-norm of the difference between the original and reconstructed solution.

\begin{figure}[!ht]
	\centering
	\begin{subfigure}[b]{0.49\textwidth}
		\centering
		\includegraphics[width=\textwidth]{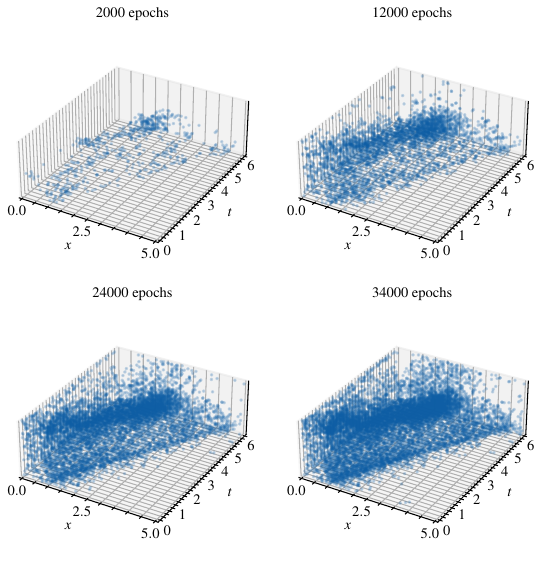}
		\caption{Evolution of the collocation points over the training epochs when using the RAR scheme. Only the new points are shown, and for this particular case, no more points were added after $34,000$ epochs.}
		\label{fig:evo_points_rar}
	\end{subfigure}
	\hfill
	\begin{subfigure}[b]{0.49\textwidth}
		\centering
		\includegraphics[width=\textwidth]{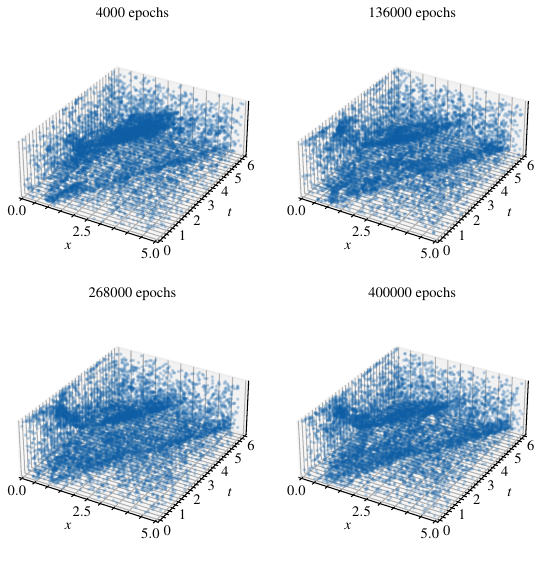}
		\caption{Evolution of the collocation points over the training epochs when using the MC sampling + RAR scheme. The number of RAR-adapted points is fixed and chosen from the beginning. Only the Residual-based Sampled points are shown.}
		\label{fig:evo_points_mcrar}
	\end{subfigure}
	\caption{Evolution of the collocation points for two different sampling techniques.}
	\label{fig:evo_point_cloud}
\end{figure}

The measurements are taken at the center of the tank. After assembling $\mathbf{X}$ from the concentration and velocity observations, we apply the SVD filtering described above and use the filtered data in the PINN reconstruction. The collocation-point distributions for the two adaptive sampling strategies introduced earlier, RAR and MC sampling + RAR, are shown in Fig. \ref{fig:evo_point_cloud}. In the RAR scheme, new points are added after a prescribed number of epochs, whereas in MC sampling + RAR, the number of adaptive points is fixed and their locations are periodically updated. In both cases, the collocation points concentrate near the sediment front and in regions where Kelvin-Helmholtz billows develop.

\begin{figure}[!ht]
	\centering
	\begin{subfigure}[b]{\textwidth}
		\centering
		\includegraphics[width=\textwidth]{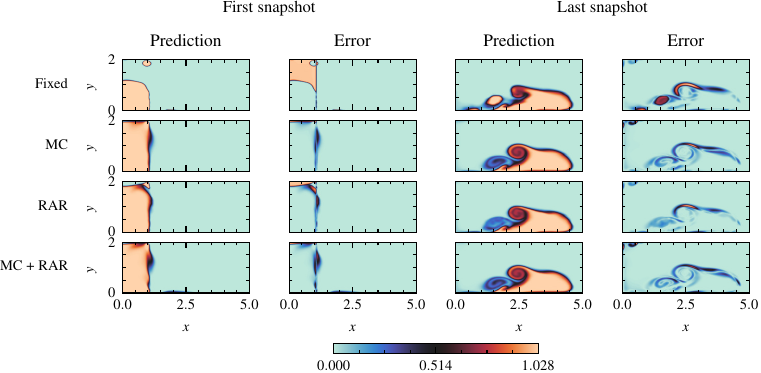}
		\caption{Predictions and errors for the concentration at $t=0.01s$  (first snapshot) and $t=6s$ (last snapshot) when normalizing the \textbf{concentration} output with a sigmoid function. When using fixed collocation points, it becomes harder to recover the initial state of the system.}
		\label{fig:phi_rec_a}
	\end{subfigure}
	\hfill
	\begin{subfigure}[b]{\textwidth}
		\centering
		\includegraphics[width=\textwidth]{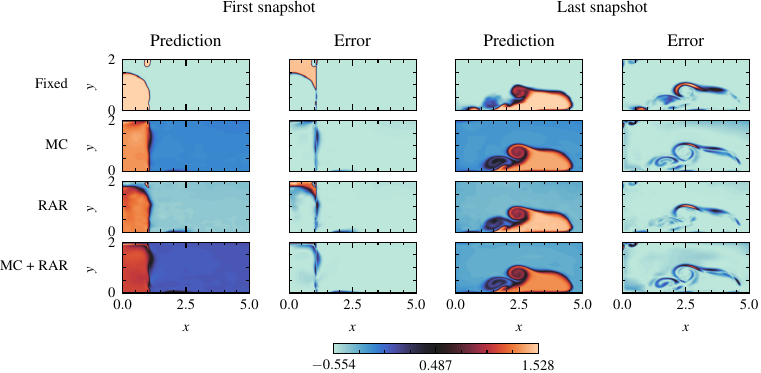}
		\caption{Predictions and errors for the concentration at $t=0.01s$ (first snapshot) and $t=6s$ (last snapshot) without normalizing the concentration output with a sigmoid function. When using fixed collocation points, it becomes harder to recover the initial state of the system.}
		\label{fig:phi_rec_b}
	\end{subfigure}
	\caption{Snapshots of the reconstruction of the entire concentration.}
	\label{fig:phi_rec}
\end{figure}

The results of the reconstruction of the concentration are shown in Fig. \ref{fig:phi_rec}. Two different snapshots are shown, along with the point-wise error plots for each sampling technique. Furthermore, since the concentration is bounded to the interval $[0,1]$, the reconstruction is performed using a sigmoid activation to the concentration predicted by the neural network (Fig. \ref{fig:phi_rec_a}), and without using it (Fig. \ref{fig:phi_rec_b}). Except for the Fixed points collocation scheme, the use of the sigmoid function to normalize the concentration does not seem to affect the solution at later times. Additionally, the Fixed points scheme struggles the most when trying to recover the system's initial state.

\begin{figure}[!ht]
	\centering
	\begin{subfigure}[b]{\textwidth}
		\centering
		\includegraphics[width=\textwidth]{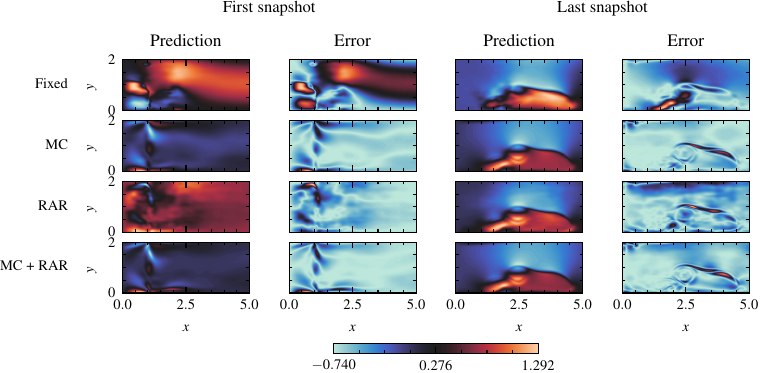}
		\caption{Predictions and errors for the $u-$velocity at $t=0.01s$  (first snapshot) and $t=6s$ (last snapshot) when normalizing the concentration output with a sigmoid function.}
		\label{fig:u_rec_a}
	\end{subfigure}
	\hfill
	\begin{subfigure}[b]{\textwidth}
		\centering
		\includegraphics[width=\textwidth]{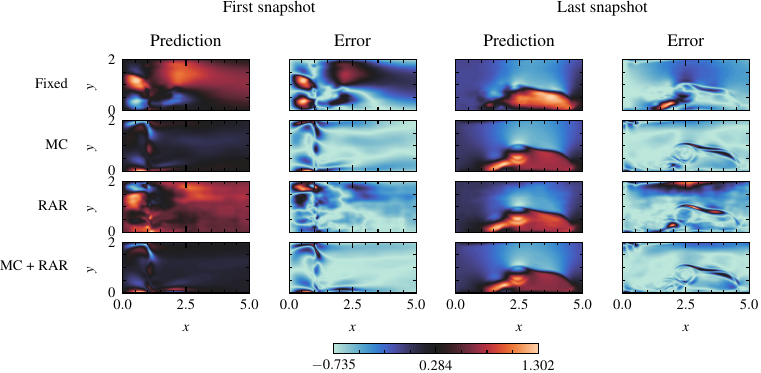}
		\caption{Predictions and errors for the $u-$velocity at $t=0.01s$ (first snapshot) and $t=6s$ (last snapshot) without normalizing the concentration with the sigmoid function.}
		\label{fig:u_rec_b}
	\end{subfigure}
	\caption{Snapshots of the reconstruction of the $u-$velocity.}
	\label{fig:u_rec}
\end{figure}

As for the concentration, Fig. \ref{fig:u_rec} shows the reconstruction of the $x-$velocity component. As it happens for the concentration, the Fixed points collocation scheme not only struggles to approximate the initial state of the system but also shows poor performance when compared to the other schemes in general, especially when close to the Kelvin-Helmholtz billows.

\begin{figure}[!ht]
	\centering
	\includegraphics[width=\textwidth]{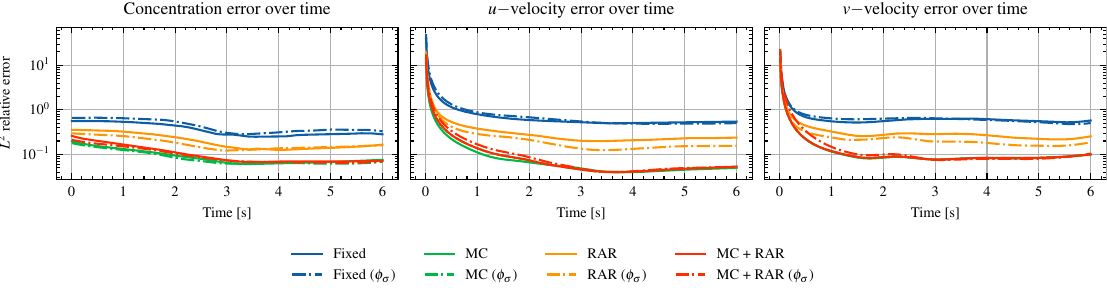}
	\caption{Concentration (\textbf{\textit{left}}), $u-$velocity (\textbf{\textit{middle}}), and $v-$velocity (\textbf{\textit{right}}) errors over time.}
	\label{fig:err_over_time_ddgf}
\end{figure}

\begin{figure}[!ht]
	\begin{subfigure}[b]{\textwidth}
		\centering
		\includegraphics[width=\textwidth]{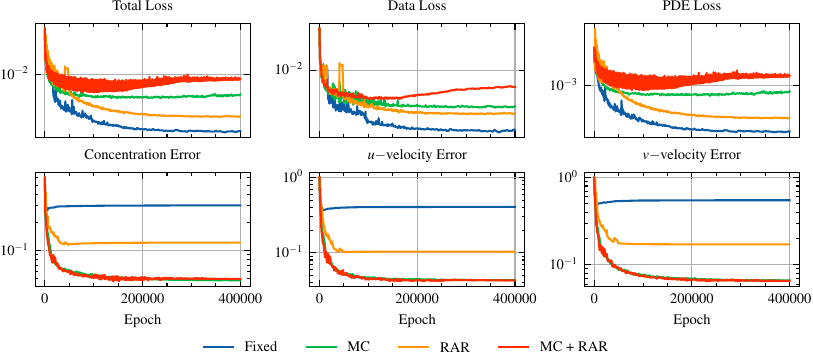}
		\caption{The loss components and errors are shown for all the sampling techniques for a case where the concentration output is mapped with a sigmoid function, and while using the dynamic loss weights. }
		\label{subfig:diffs_losses_err}
	\end{subfigure}
	\hfill
	\begin{subfigure}[b]{\textwidth}
		\centering
		\includegraphics[width=\textwidth]{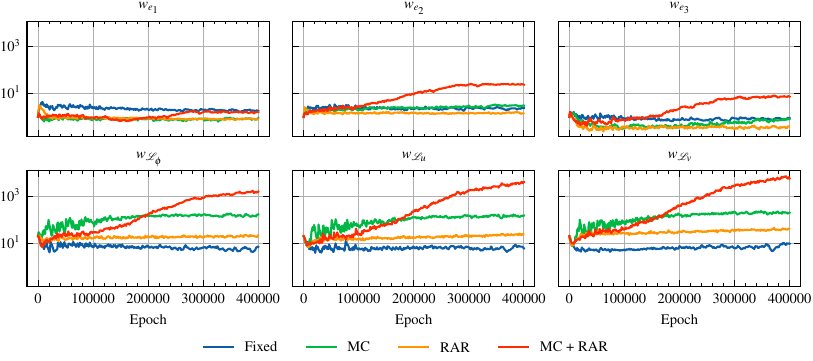}
		\caption{For all the sampling techniques, we show the evolution of the dynamic loss weights. }
		\label{subfig:diffs_dyn_weights}
	\end{subfigure}
	\caption{Evolution of the losses, errors, and dynamic loss weights for different collocation points sampling schemes. }
	\label{fig:loss_err_dynweights}
\end{figure}

To provide a better understanding of the performance of all sampling techniques and the use of the sigmoid function to normalize the concentration, Fig. \ref{fig:err_over_time_ddgf} shows the error over time for the concentration and both velocity components. The subscript $()_\sigma$ indicates that the results were obtained by applying the sigmoid function to the concentration. Since the initial conditions are not incorporated during training, the error is significantly large at $t=0s$, especially for the velocity components, but it quickly decays as the front starts to move. The MC sampling and MC sampling + RAR sampling schemes provide the best results overall.

In addition to the study of sampling techniques and output normalization for concentration, the effect of using dynamic loss weights is also examined. The results are shown in Fig. \ref{fig:loss_err_dynweights}. The weight for the continuity equation is fixed to $1$, while all the other weights are adapted during training.
The weights nomenclature follows the same pattern as in Figure \ref{fig:pinn_scheme_ddgf}, where $e_1$, $e_2$, and $e_3$ are the Advection-Diffusion, $x-$momentum, and $y-$momentum equations respectively, while $\mathcal{L}_{c}$, $\mathcal{L}_{u}$, and $\mathcal{L}_{v}$ are the concentration, $u-$velocity, and $v-$velocity data loss components. In all cases, the dynamic weights are updated only every 200 epochs.
In the absence of a ground truth solution, one might easily be misled by using only the loss function as a criterion for choosing the best sampling technique or even stopping the training. Even though the Fixed collocation points scheme shows a quick convergence when it comes to the loss function, this method is the one with the worst results when we compare its predictions with the ground truth solution along the training process, which happens to be the opposite of the behavior of both the MC and MC+RAR schemes. For the RAR scheme, the error decays up to a certain epoch as we keep increasing the number of collocation points, whereas the loss function exhibits spikes whenever we add new points. Meanwhile, the MC+RAR tends to show the same spikes, but more frequently throughout the training process, while exhibiting a damping behavior as we approach a good approximation.

As pointed out in \cite{JIN2021109951}, not only does the choice of loss weights depend on the problem, but even when using dynamic weights, their evolution is obviously problem-dependent. In our case, the evolution of the dynamic loss weights depends on the sampling scheme for the collocation points. When it comes to the weights for the differential equations ($w_{e_1}$, $w_{e_2}$, and $w_{e_3}$), all methods exhibit similar behavior, except for the MC+RAR scheme for the momentum equations, where it starts to show a growing for the weights until it becomes constant after some point. Regarding the data-related dynamic loss weights, the MC and MC+RAR schemes show higher weights as training progresses, but the MC+RAR scheme continues to increase until the end of training.

Taken together, Figs. \ref{fig:evo_points_rar}–\ref{fig:loss_err_dynweights} illustrate a cautionary result: the fixed-point scheme achieves a lower training loss than the adaptive schemes throughout training, yet produces the worst field reconstruction when compared against the ground truth. This shows that the loss function alone is an unreliable convergence diagnostic for PINNs. Among adaptive strategies, MC and MC+RAR consistently achieve relative $\mathcal{L}^2$ errors roughly one order of magnitude below the fixed scheme in the velocity components, confirming that the spatial distribution of collocation points has a larger impact on reconstruction quality than the number of training iterations.



\begin{figure}[!ht]
	\centering
	\includegraphics[width=\textwidth]{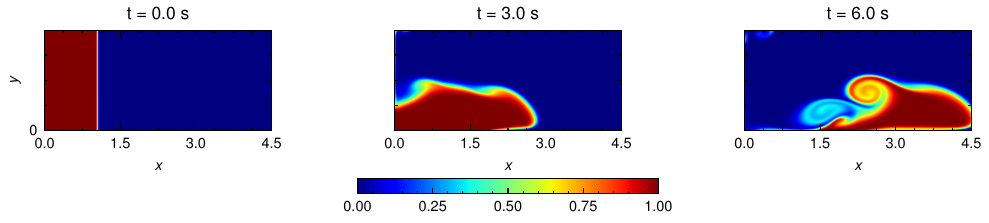}
	\caption{Snapshots showing the selected region for the training dataset used to estimate the Grashof number.}
	\label{fig:inv2_snapshots}
\end{figure}

Besides reconstructing the flow fields, one may also be interested in estimating some physical properties related to the problem. Estimating the Grashof number $Gr$ and the Schmidt number $Sc$, both shown in Eqs. \eqref{eq:momentum_pdgf}-\eqref{eq:transport_pdgf} form a reasonable set of parameters to be estimated for this PDE system. However, we restrict ourselves to only estimating the Grashof number. A high Grashof number indicates that the flow is turbulent. Therefore, by correctly identifying the Grashof number, we characterize the flow regime.
We now compute the Grashof number $Gr$, so there is no need to use the entire dataset from the FEM simulation. Therefore, we randomly selected $N_{data}$ points from the whole dataset for taking measurements for the concentration and velocity components, given that those points must be in a sub-region defined by $0 \leq x \leq 4.5$, $0 \leq y \leq 2$, and $0 \leq t \leq 6$. Figure \ref{fig:inv2_snapshots} shows 3 different snapshots for the constrained domain.


For this problem, several runs are performed to investigate how parameter initialization may affect the convergence of the Grashof number to its true value. It must be noted that $L_2$-regularization of the Grashof number was necessary to achieve reasonably accurate results. Thus, for each run, the $L_2$-regularization factor for the Grashof number will also be randomly picked. The hyperparameters for this problem are shown in Table \ref{tab:hyper_params_Gr}. The same number of points ($N_{data}$) is selected as the measurements for the velocity and concentration, as shown in Table  \ref{tab:hyper_params_Gr}

\begin{table}[!ht]
	\centering
	\caption{Hyper-parameters for Grashof Identification.}
	\begin{tabular}{@{}ll@{}}
		\toprule
		Hyper Parameters                                     & Value/Description               \\ \midrule
		\# of Hidden Layers                                  & 5                               \\
		\# of Neurons                                        & 50                              \\
		Optimizer                                            & Adam                            \\
		Learning Rate ($\eta$)                               & $0.0008$                        \\
		\# of Sampling points ($N_{\mathcal{F}}$)            & $5,000$                         \\
		\# of points in the test dataset                     & $2,500$                         \\
		\# of runs performed                                 & $50$                            \\
		\# of concentration data points ($N_{data}$)         & $20,000$                        \\
		\# of velocity data points  ($N_{data}$)             & $20,000$                        \\
		$L_2$-regularization factor for $Gr$ ($\alpha_{Gr}$) & $\mathcal{N}(2.75e-5, 8.25e-6)$ \\
		$Gr$ number initialization                           & $\mathcal{U}(0,10)\cdot 10^6$           \\
		Activation Function                                  & (adaptive) Swish                           \\ \bottomrule
	\end{tabular}
	\label{tab:hyper_params_Gr}
\end{table}

The $L_2$-regularization factor is included in the loss function for this problem (as shown in Figure \ref{fig:pinn_scheme_ddgf}) as

\begin{equation}
    \mathcal{L}(\mathbf{\theta}) = w_{data} \mathcal{L}_{data}(\mathbf{\theta}) + w_{PDE} \mathcal{L}_{PDE}(\mathbf{\theta}) + \frac{\alpha_{Gr}}{2} |Gr|^2
\end{equation}


\begin{figure}[!ht]
	\centering
	\includegraphics[width=\textwidth]{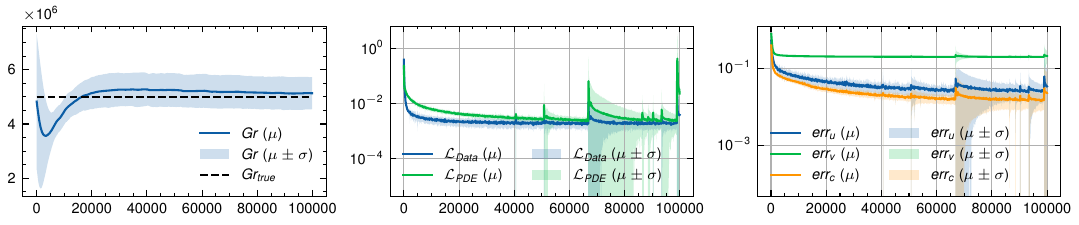}
	\caption{Convergence of the Grashof number \textbf{(left)}, loss function \textbf{(middle)} and relative error \textbf{(right)} for several runs.}
	\label{fig:Gr_conv}
\end{figure}

For each of the 50 runs performed for this case, not only are the parameters for the neural network randomly initialized as usual, but also the initial guess for $Gr$ and the $L_2$-regularization factor $\alpha_{Gr}$. From Figure \ref{fig:Gr_conv}, it is possible to see that within a few iterations, the mean value for the Grashof number across all the runs starts to float around its true value. Furthermore, notice that, at the beginning of the training, no matter the value into which the Grashof number is initialized, it always starts to decrease for a few iterations before converging to a value close to the real one. This behavior seems to be related to the fact that the components of a PINN's loss function have different scales, with the $L^2$-regularization term associated with the Grashof number dominating the loss at the very beginning of training.

\paragraph*{\textbf{Section Summary}} This section demonstrated the application of Physics-Informed Neural Networks to density-driven gravity flows, addressing both field reconstruction from sparse data and inverse parameter estimation. For the forward reconstruction problem, we showed that PINNs can recover velocity, pressure, and concentration fields from scattered concentration measurements alone, with $250{,}000$ data points providing sufficient information for accurate reconstruction. The choice of collocation-point sampling strategy proved critical: Monte Carlo and hybrid MC+RAR schemes consistently outperformed fixed-point distributions, particularly in capturing sharp concentration fronts and Kelvin--Helmholtz instabilities. Dynamic loss weighting further improved convergence by balancing the competing PDE, boundary, and data-misfit terms throughout training. For the inverse problem, PINNs successfully identified the Grashof number from partial observations, although $L_2$-regularization of the estimated parameter was necessary for stable convergence. These results highlight PINNs as a versatile tool for coupled flow problems, but also underscore the importance of careful hyperparameter selection and the difficulty of relying solely on the loss function as a convergence diagnostic.

\subsubsection{$\beta$-Variational Autoencoders}
\label{sec:betavaes}

\par
As discussed in Section \ref{sec:svd}, methods like POD and DMD assume that the system dynamics can be represented as a linear combination of orthogonal basis functions. However, many complex systems—particularly those involving nonlinear interactions, bifurcations, or multi-scale behavior—cannot be accurately captured within such a linear subspace. This limitation has motivated the development of nonlinear reduced-order models (ROMs) that learn low-dimensional manifolds to more effectively represent nonlinear dependencies in the data.

Nonlinear ROMs extend traditional linear dimensionality reduction by introducing nonlinear mappings between the high-dimensional physical space and a compact latent space. Among the techniques that have emerged in this context, Autoencoders (AEs) \cite{Velho2025} and their probabilistic variant, the Variational Autoencoder (VAE) \cite{diederik2019introduction, Cinelli2021}, have received increasing attention. Unlike POD, which projects data onto an orthogonal basis, autoencoders employ neural networks to encode input data into a latent representation and reconstruct it through a decoder, thereby learning a nonlinear manifold that approximates the underlying data distribution. The Variational Autoencoder introduces a probabilistic formulation in which the encoder learns to approximate the posterior distribution of the latent variables conditioned on the input, while the decoder reconstructs the input by sampling from this distribution. That is, the latent coordinates $\mathbf{z}$ comprise a combination of the mean $\mu$ and standard deviation $\sigma$ learned from the data. An additional noise $\epsilon$ is added to the scaled standard deviation to sample the latent vector $\mathbf{z}$, a method known as the reparameterization trick \cite{Kingma2014, NIPS2015_bc731692} that allows for backpropagation through the network. Figure \ref{fig:betavae_arch} illustrates the architecture. 

\begin{figure}[ht!]
    \centering
    \includegraphics[width=\linewidth]{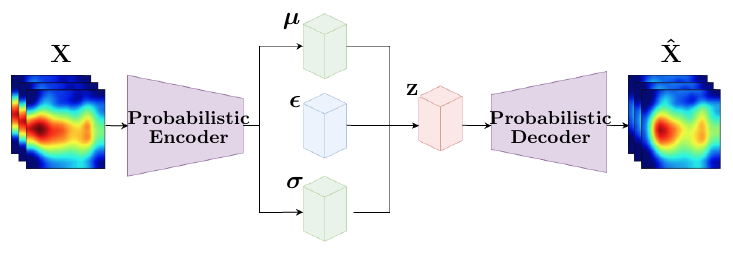}
    \caption{Variational Autoencoder (VAE) architecture. The encoder maps the high-dimensional input field into a latent representation parameterized by the mean $\boldsymbol{\mu}$ and standard deviation $\boldsymbol{\sigma}$.}
    \label{fig:betavae_arch}
\end{figure}

The training process of a VAE is based on the minimization of the Evidence Lower Bound (ELBO), which balances reconstruction accuracy and latent-space regularization through a Kullback–Leibler (KL) divergence term. The $\beta$-Variational Autoencoder ($\beta$-VAE) extends this formulation by introducing a weighting factor $\beta$ that modulates the relative importance of the KL divergence term, thus controlling the trade-off between reconstruction fidelity and latent-space disentanglement \cite{Burgess2018}. Higher values of $\beta$ promote a more structured and interpretable latent space at the expense of reconstruction precision, whereas lower values prioritize accuracy but may lead to entangled latent representations.

In the context of reduced-order modeling, $\beta$-VAEs are particularly well-suited to representing complex, nonlinear dynamics arising in high-dimensional systems such as turbulent flows \cite{Solera-Rico2024-VariationalFlows, EIVAZI2022117038}. By learning a nonlinear manifold that compactly captures the essential flow features, $\beta$-VAEs can serve as data-driven surrogates that generalize beyond the training distribution. Once trained, the decoder acts as a nonlinear reconstruction operator, enabling rapid recovery of the high-dimensional state from its low-dimensional latent representation. The evolution of the latent variables can then be modeled using separate dynamical systems, such as regression-based mappings, recurrent neural networks, or operator-learning frameworks, yielding an efficient surrogate capable of predicting temporal dynamics without directly integrating the full-order model.

Compared to linear techniques such as POD, $\beta$-VAEs can represent nonlinear correlations and multiscale structures that cannot be captured by orthogonal projections. The probabilistic nature of the latent space facilitates smooth interpolation between physical states and enhances robustness to noise, while the regularization term helps to prevent overfitting—an essential property for data-scarce scenarios common in scientific simulations. Despite these advantages, challenges remain regarding the interpretability and stability of the latent dynamics, as well as the need for careful hyperparameter tuning. Also, some \textit{a priori} knowledge regarding the distribution of the generated outputs must be known. The (parametric) distributions defining the probabilistic model must be chosen upfront in prescribed models \cite{gan_rozza}. Nevertheless, the quality of the model can be affected if too simplistic distributions are used \cite{rezende2018tamingvaes}.

With that, we recover the Rayleigh-Bénard dataset used in Section \ref{sec:svd} to evaluate the capabilities of the $\beta$-VAE framework. We use the networks proposed in \cite{EIVAZI2022117038, Solera-Rico2024-VariationalFlows} as our model. The model consists of a  convolutional encoder that progressively reduces the spatial dimensions of the input fields using a sequence of strided convolutional layers paired with Exponential Linear Unit (ELU) activations. Once the spatial features are extracted and downsampled, a fully connected layer processes the flattened output. This representation is then mapped into two parallel dense layers that define the mean $\mu$ and standard deviation $\sigma$ of a $d$-dimensional latent space. A sampling layer subsequently draws the latent vector using the reparameterization trick \cite{Kingma2014, NIPS2015_bc731692}. To reconstruct the physical data, the decoder mirrors the encoder structure. It utilizes fully connected layers followed by a sequence of strided transpose convolutions and ELU activations to progressively upsample the latent representation back to the original input resolution. The model is trained for $12,000$ epochs using the Adam Optimizer with a learning rate of 0.00025.

To effectively train the model and avoid the common computational issue of posterior collapse, we employ a $\beta$-annealing technique \cite{bowman2016generating}. This training strategy initializes the weight of the Kullback-Leibler divergence term near zero and gradually increases it over the training epochs until the target $\beta$ parameter is reached. This approach allows the autoencoder to first learn a meaningful deterministic representation of the complex fluid dynamics before progressively enforcing the probabilistic regularization on the latent space. We investigate the impact of this regularization parameter by training independent models with target values of $\beta$ equal to $10^{-5}$, $5\times10^{-5}$, $10^{-4}$, $5 \times 10^{-4}$, and $10^{-3}$. Given that our goal is to have disentangled representations of the complex Rayleigh-Bénard convection described in Fig. \ref{fig:rb-setup}, we need to assess if the modes are close to orthogonal. In this case, we compute $\eta_O$ such that:

\begin{equation}
    \eta_O = \dfrac{\Vert \mathbf{G} - \mathrm{diag}(\mathbf{G}) \Vert_F}{\Vert \mathbf{G} \Vert_F}
\end{equation}

\noindent where $\mathbf{G} = \mathbf{z}^T\mathbf{z}$, being $\mathbf{z}$ the layer that combines the contributions from $\boldsymbol{\sigma}$ and $\boldsymbol{\mu}$.

\par Table \ref{tab:1} highlights the fundamental trade-off inherent to $\beta$-VAEs regarding the balance between the reconstruction accuracy of the physical fields and the disentanglement of the underlying latent modes. As the hyperparameter $\beta$ decreases from $0.00100$ to $0.00001$, the Frobenius relative error ($\eta_F$) consistently drops for both Rayleigh numbers. Lowering $\beta$ relaxes the constraint on the Kullback-Leibler divergence, allowing the network to dedicate more capacity to minimizing the reconstruction loss. For the $Ra = 10^6$ dataset, $\eta_F$ improves by an order of magnitude, and a similar monotonic improvement is observed for $Ra = 10^{10}$, where the error falls from $0.260$ to $0.107$. Conversely, higher values of $\beta$ impose a stronger penalty on the latent distributions, driving them toward an isotropic Gaussian prior and encouraging the physical modes to remain independent and orthogonal. For $Ra = 10^6$, the best disentanglement is achieved at the highest $\beta$ of 0.00100, yielding an $\eta_O$ of $0.061$. For the highly turbulent $Ra = 10^{10}$ regime, the network finds an optimal balance at $\beta = 0.00050$, achieving an $\eta_O$ of $0.019$ before the orthogonality rapidly degrades at lower parameter values.

\begin{table}[!t]
\centering
\caption{Frobenius relative error ($\eta_F$) and orthogonality metric ($\eta_O$) for different values of $\beta$.}
\label{tab:1}       
\renewcommand{\arraystretch}{1.2} 
\setlength{\tabcolsep}{8pt} 
\resizebox{0.55\textwidth}{!}{
\begin{tabular}{c c c c c}
\hline\noalign{\smallskip}
        & \multicolumn{2}{c}{$Ra = 10^6$}               & \multicolumn{2}{c}{$Ra = 10^{10}$}            \\
\noalign{\smallskip}\cline{2-3} \cline{4-5}\noalign{\smallskip} 
$\beta$ & $\eta_F$ & $\eta_O$ & $\eta_F$ & $\eta_O$ \\
\noalign{\smallskip}\svhline\noalign{\smallskip}
0.00100   & 0.121 & 0.061 & 0.260  & 0.030  \\
0.00050  & 0.047 & 0.103 & 0.218 & 0.019 \\
0.00010  & 0.021 & 0.137 & 0.142 & 0.048 \\
0.00005 & 0.034 & 0.284 & 0.134 & 0.052 \\
0.00001 & 0.012 & 0.342 & 0.107 & 0.119 \\
\noalign{\smallskip}\hline\noalign{\smallskip}
\end{tabular}
}
\end{table}

This is supported by the illustration of the reconstruction of the last snapshot in Fig. \ref{fig:beta_comparison_grid}, where the same value of $\beta$ did not provide good representations for both cases. Of course, comparing the two regimes also reveals the profound impact of flow complexity. The $Ra = 10^{10}$ case represents a much more chaotic system, which naturally makes the model struggle to capture fine-scale structures. Consequently, its overall reconstruction errors are substantially higher across all evaluated $\beta$ values compared to the $Ra = 10^6$ case, making the selection of the penalty term a much more delicate compromise for highly turbulent flows. However, given that it is useful to have SciML models that generalize across physical regimes (see, for instance, the discussion of foundational models in SciML \cite{choi2025definingfoundationmodelscomputational}), it is ideal to have models that generalize during training regardless of the physical regime in the data. That said, we focus on automating the strategy for selecting the optimal value of $\beta$ based on the training dataset. Several strategies exist for different applications \cite{Burgess2018, bowman2016generating}. Here, we propose using a similar approach as described in the adaptive weights for losses exposed in Section \ref{sec:pinns}.

\begin{figure}[ht!]
    \sidecaption
    \centering
    \begin{minipage}[t]{0.99\textwidth} 
        \vspace{-0.5cm} 
        \begin{subfigure}{0.65\textwidth}
            \centering
            \includegraphics[width=\linewidth]{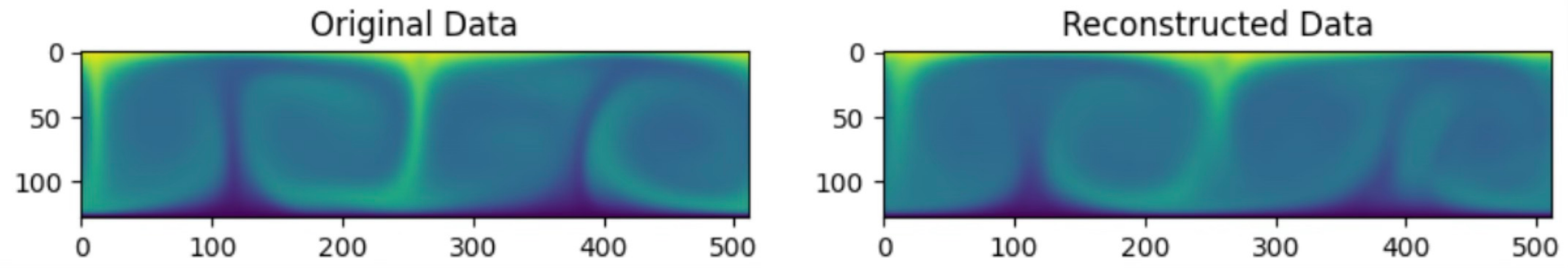}
            \label{fig:ra106_0001}
        \end{subfigure}\hfill
        \begin{subfigure}{0.31\textwidth}
            \centering
            \includegraphics[trim={1.6cm 0 0 0}, clip, width=\linewidth]{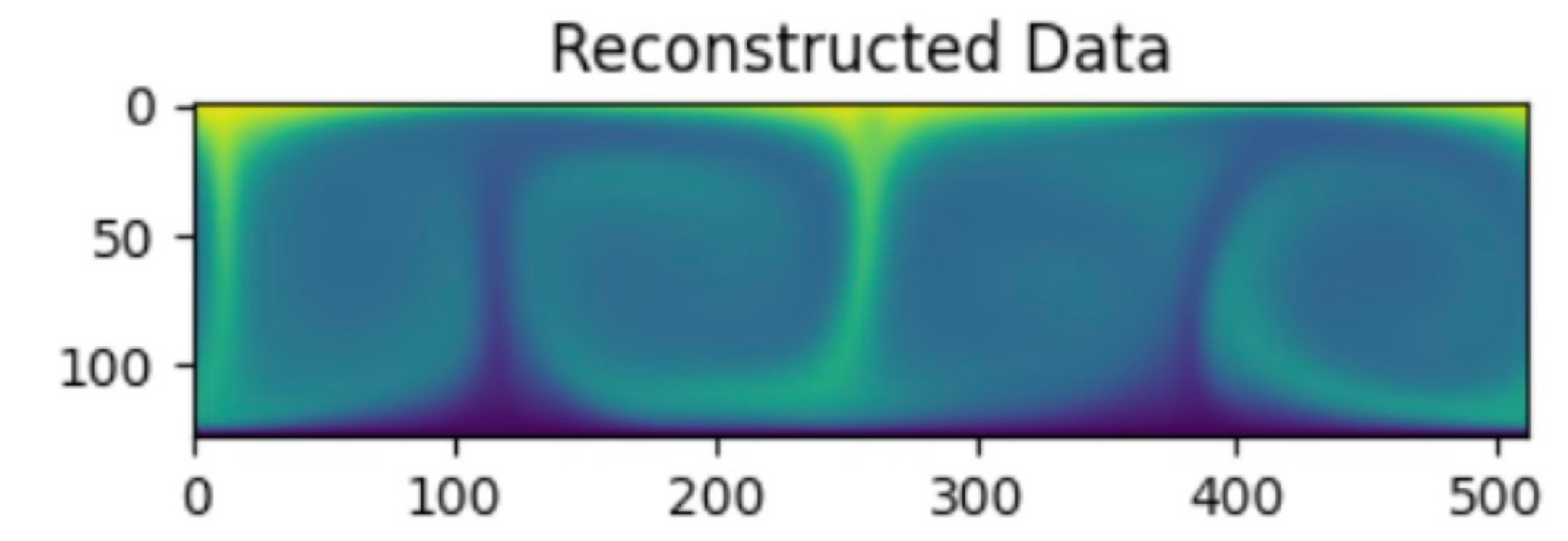}
            \label{fig:ra106_00001}
        \end{subfigure}
        \vspace{0.4cm} 
        \begin{subfigure}{0.65\textwidth}
            \centering
            \includegraphics[width=\linewidth]{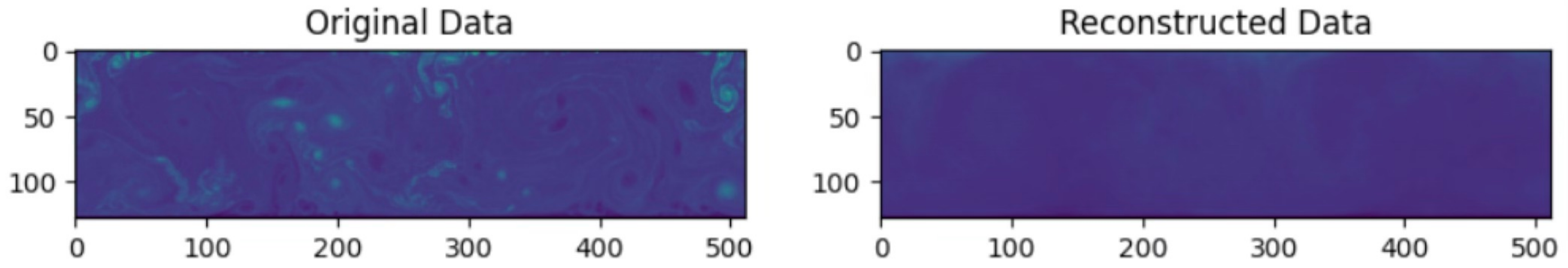}
            \label{fig:ra1010_0001}
        \end{subfigure}\hfill
        \begin{subfigure}{0.31\textwidth}
            \centering
            \includegraphics[trim={38.5cm 0 0 0}, clip, width=\linewidth]{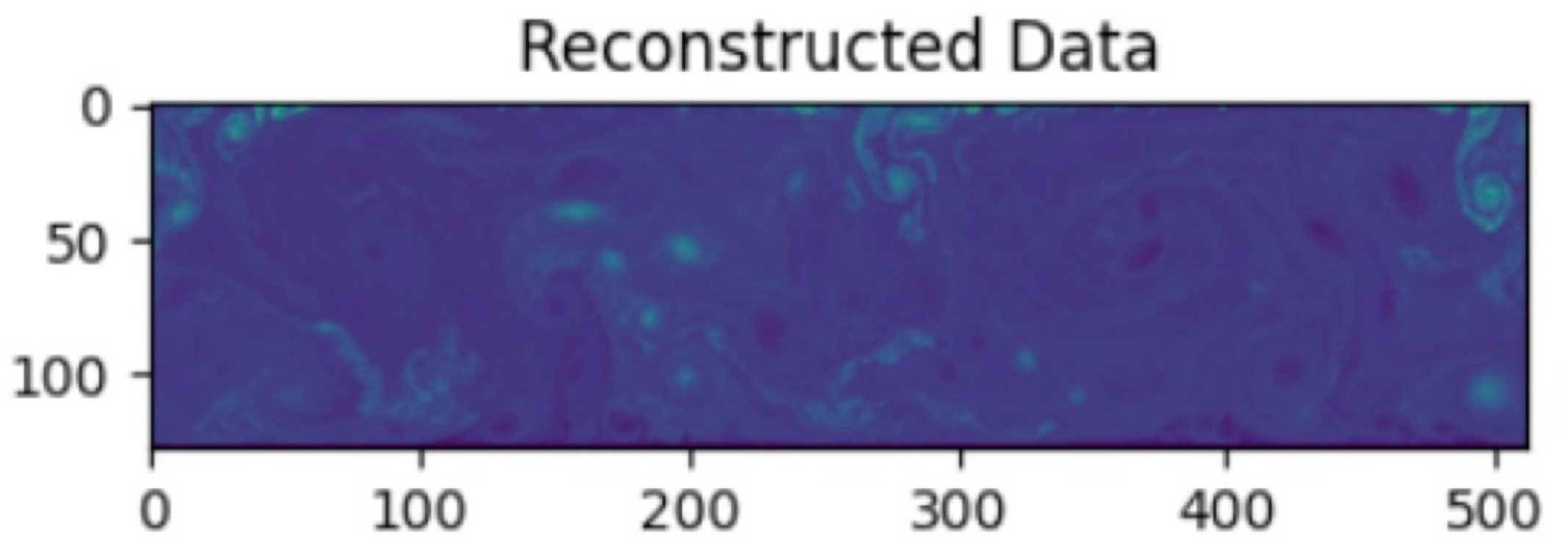}
            \label{fig:ra1010_000001}
        \end{subfigure}
    \end{minipage}
    \setcounter{figure}{29}
    \caption{Comparison of reconstructed fields and error metrics across different Rayleigh numbers and $\beta$ penalty constraints. Top figures are results for $Ra = 10^6$ and bottom results are for $Ra = 10^{10}$. Left, center and right plots are, respectively, ground truth, reconstruction using $\beta = 0.00100$ and $\beta = 0.00001$.}
    \label{fig:beta_comparison_grid}
\end{figure}

Considering that PINNs often have to balance the effects of the partial loss terms (such as boundary conditions, initial conditions and the PDE itself, as previously described in Section \ref{sec:pinns} in Eq. (\ref{eq:weights})), several works have been proposed to find an optimal set of weights to properly optimize total loss function \cite{relobralo, softadapt}. It seems natural to translate this strategy to $\beta-$VAEs. The total loss term here contains two terms that are antagonistic: as we improve the reconstruction, we get more entangled modes, and vice versa. Based on the strategy of adaptive weights, we test ReLoBRaLo \cite{relobralo} and SoftAdapt \cite{softadapt} to see if the $\beta-$VAE models can properly learn representations with a sufficient level of disentanglement without tuning the hyperparameter $\beta$.

\begin{figure}[ht!]
    \sidecaption
    \centering
    \begin{minipage}[t]{0.65\textwidth} 
        \vspace{-3.5cm} 
        \begin{subfigure}{\textwidth}
            \centering
            \includegraphics[width=\linewidth]{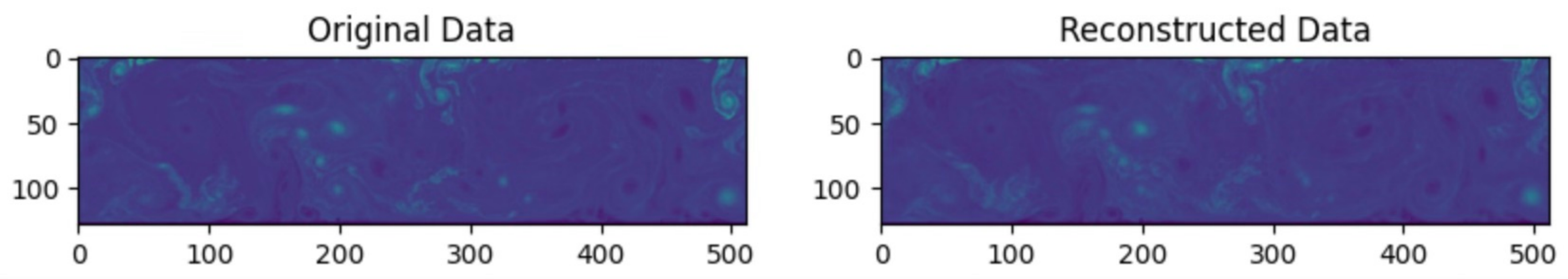}
            \label{fig:beta1}
        \end{subfigure}\\
        \vspace{0.4cm} 
        \begin{subfigure}{\textwidth}
            \centering
            \includegraphics[width=\linewidth]{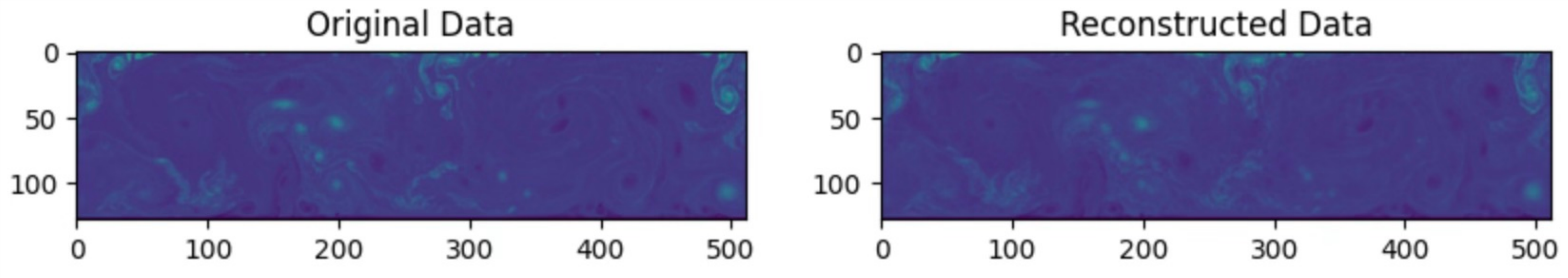}
            \label{fig:beta2}
        \end{subfigure}\\
        \begin{subfigure}{0.6\textwidth}
            \hspace{2cm}
            \includegraphics[width=\linewidth]{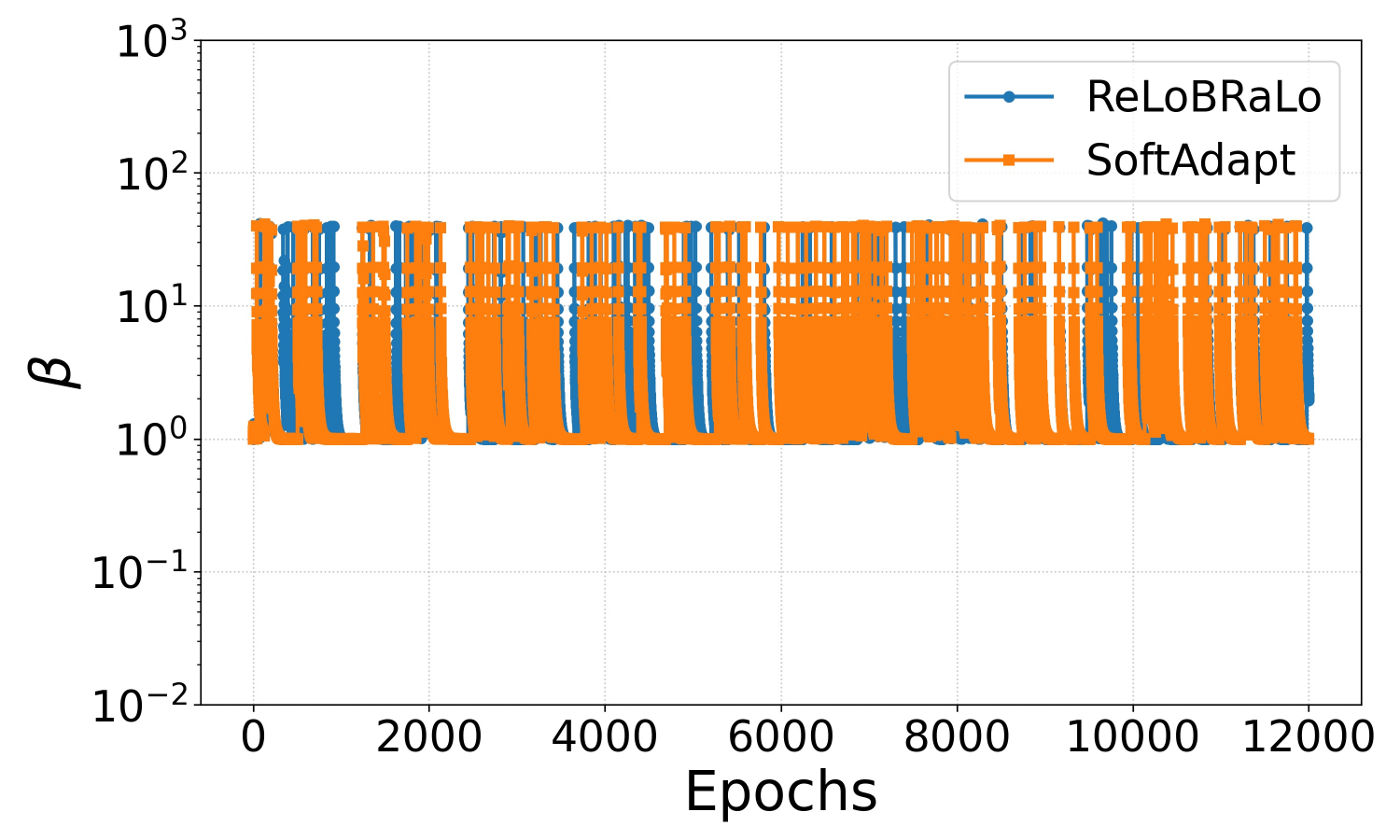}
            \label{fig:beta3}
        \end{subfigure}
    \end{minipage}
    \setcounter{figure}{30}
    \caption{Field approximations for ReLoBRaLo (top), SoftAdapt (middle) and $\beta$ evolution over training time. The ratio between the weights can oscillate up to $2$ orders of magnitude.}
    \label{fig:adaptive_beta}
\end{figure}

\par Figure \ref{fig:adaptive_beta} shows the temperature field for the $\beta-$VAE reconstruction after using both adaptivity strategies. First, we notice the similarities between the reconstructions and the best solutions shown in Fig. \ref{fig:beta_comparison_grid}. We also notice how $\beta$ varies two orders of magnitude during training, meaning that, as the reconstructions improve, $\beta$ needs to be adapted to optimize the Kullback-Leibler divergence term. Table \ref{tab:betas} shows the metrics for different values of $\beta$. They yield intermediate values compared with those in Table \ref{tab:1}, indicating that both strategies can be used to converge to solutions with adequate metrics without tuning $\beta$.

\begin{table}[!t]
\centering
\caption{Frobenius relative error ($\eta_F$) and orthogonality metric ($\eta_O$) for different values of $\beta$.}
\label{tab:betas}       
\renewcommand{\arraystretch}{1.2} 
\setlength{\tabcolsep}{8pt} 
\resizebox{0.55\textwidth}{!}{
\begin{tabular}{c c c c c}
\hline\noalign{\smallskip}
        & \multicolumn{2}{c}{$Ra = 10^6$}               & \multicolumn{2}{c}{$Ra = 10^{10}$}            \\
\noalign{\smallskip}\cline{2-3} \cline{4-5}\noalign{\smallskip} 
$\beta$ & $\eta_F$ & $\eta_O$ & $\eta_F$ & $\eta_O$ \\
\noalign{\smallskip}\svhline\noalign{\smallskip}
ReLoBRaLo   & 0.033 & 0.011 & 0.125  & 0.021  \\
SoftAdapt  & 0.027 & 0.011 & 0.139 & 0.041 \\
\noalign{\smallskip}\hline\noalign{\smallskip}
\end{tabular}
}
\end{table}

\paragraph*{\textbf{Section Summary}} This section introduced $\beta$-Variational Autoencoders as a nonlinear reduced-order modeling framework for complex turbulent flows where linear methods such as POD and DMD are insufficient. Applied to Rayleigh--B\'{e}nard convection data at $Ra=10^6$ and $Ra=10^{10}$, the $\beta$-VAE demonstrated the fundamental trade-off between reconstruction accuracy and latent-space disentanglement: lower values of $\beta$ improve field reconstruction at the cost of entangled latent representations, while higher values promote orthogonal, interpretable modes but degrade accuracy. This trade-off is particularly delicate for highly turbulent regimes, where fine-scale structures are inherently difficult to capture. To address the challenge of manually tuning $\beta$, we proposed and validated the use of adaptive loss-weighting strategies (ReLoBRaLo and SoftAdapt), originally developed for PINNs, to automatically balance the reconstruction and KL-divergence terms during training. Both strategies converged to solutions with competitive reconstruction errors and near-orthogonal latent modes without requiring manual hyperparameter selection, demonstrating their potential for building generalizable SciML models across physical regimes.

\section{Conclusions and Future Directions}

This chapter examined the application of Scientific Machine Learning to coupled fluid flow and transport problems, surveying methods across two broad categories. SVD-based approaches, particularly Dynamic Mode Decomposition, demonstrated their effectiveness as computationally efficient surrogates for density-driven gravity currents, remaining robust under adaptive mesh refinement and lossy data compression. Neural network-based methods extended these capabilities to more challenging regimes: PINNs enabled simultaneous field reconstruction and parameter identification from sparse measurements, while $\beta-$Variational Autoencoders provided a nonlinear dimensionality reduction framework capable of capturing the complex dynamics of turbulent Rayleigh-Bénard convection. Across both families of methods, practical strategies such as adaptive collocation, dynamic loss weighting, and automated hyperparameter selection proved essential for achieving reliable results. 

Looking ahead, several open challenges remain. Linear methods like DMD are inherently limited in highly turbulent regimes where the singular value decay is slow, motivating the continued development of nonlinear surrogates. For $\beta-$VAEs, automating the selection of the regularization parameter and improving latent space interpretability are key priorities. More broadly, building SciML models capable of generalizing across physical regimes — rather than being tuned to a single set of governing parameters — represents a central goal for the field.

\section*{Acknowledgements}
This study was partially financed by the Coordenação de Aperfeiçoamento de Pessoal de Nível Superior-Brasil (CAPES)—Finance Code 001. It is also partially supported by CNPq, Brazilian Petroleum Agency, and Petrobras. Figure \ref{fig:lock2d_modes} awarded the first author with a special mention at the Arts \& Science Contest at the World Congress on Computational Mechanics (WCCM) 2020.

\section*{Credit Authorship Contribution Statement}
Gabriel F. Barros: Conceptualization, Methodology, Software, Validation, Formal Analysis, Visualization, Writing – original draft, Writing – review \& editing. Romulo M. Silva: Conceptualization, Methodology, Formal Analysis, Visualization, Writing – original draft, Writing – review \& editing. Alvaro L. G. A. Coutinho: Conceptualization, Methodology, Validation, Formal Analysis, Visualization, Writing – original draft, Writing – review \& editing, Resources, Supervision, Funding Acquisition.

\section*{Note}
This manuscript has been accepted for publication as a chapter in the book Scientific Machine Learning for Predictive Modeling: Bridging Data-Driven and Physics-Based Approaches in Computational Science and Engineering*, edited by A. Cunha Jr, F. P. Santos, F. A. Rochinha, A. L. G. A. Coutinho, to be published by Springer Nature. The final authenticated version will be available through Springer Nature.

\section*{Conflict of Interest Statement}
The authors have no conflicts of interest to declare that are relevant to the content of this Chapter.

\section*{Declaration of generative AI and AI-assisted technologies in the manuscript preparation process}
During the preparation of this work the author(s) used Grammarly in order to improve text readability. After using this tool/service, the author(s) reviewed and edited the content as needed and take(s) full responsibility for the content of the published Chapter.

\bibliographystyle{spmpsci}
\bibliography{bibliography}

@article{Castaing_Gunaratne_Heslot_Kadanoff_Libchaber_Thomae_Wu_Zaleski_Zanetti_1989, title={Scaling of hard thermal turbulence in Rayleigh-Bénard convection}, volume={204}, DOI={10.1017/S0022112089001643}, journal={Journal of Fluid Mechanics}, author={Castaing, Bernard and Gunaratne, Gemunu and Heslot, François and Kadanoff, Leo and Libchaber, Albert and Thomae, Stefan and Wu, Xiao-Zhong and Zaleski, Stéphane and Zanetti, Gianluigi}, year={1989}, pages={1--30}}

@article{padula2024,
  author  = {Padula, Guglielmo and Girfoglio, Michele and Rozza, Gianlugi},
  title   = {A brief review of reduced order models using intrusive and non-intrusive techniques},
  journal = {Proceedings in Applied Mathematics and Mechanics},
  volume  = {24},
  number  = {4},
  pages   = {e202400210},
  doi     = {10.1002/pamm.202400210},
  year    = {2024}
}

@article{Barros2022,
  author  = {Barros, Gabriel F. and Grave, Mal{\'{u}} and Viguerie, Alex and Reali, Alessandro and Coutinho, Alvaro L.G.A.},
  title   = {{Dynamic mode decomposition in adaptive mesh refinement and coarsening simulations}},
  journal = {Engineering with Computers},
  volume  = {38},
  number  = {5},
  pages   = {4241--4268},
  doi     = {10.1007/s00366-021-01485-6},
  year    = {2022}
}

@article{Barros2023,
  author  = {Barros, Gabriel F. and Grave, Mal{\'{u}} and Camata, Jos{\'{e}} J. and Coutinho, Alvaro L.G.A.},
  title   = {{Enhancing dynamic mode decomposition workflow with in situ visualization and data compression}},
  journal = {Engineering with Computers},
  volume  = {40},
  number  = {1},
  pages   = {455--476},
  doi     = {10.1007/s00366-023-01805-y},
  year    = {2024}
}

@article{Chen31122024,
  author    = {Xinhai Chen and Zhichao Wang and Liang Deng and Junjun Yan and Chunye Gong and Bo Yang and Qinglin Wang and Qingyang Zhang and Lihua Yang and Yufei Pang and Jie Liu},
  title     = {Towards a new paradigm in intelligence-driven computational fluid dynamics simulations},
  journal   = {Engineering Applications of Computational Fluid Mechanics},
  volume    = {18},
  number    = {1},
  pages     = {2407005},
  year      = {2024},
  publisher = {Taylor \& Francis},
  doi       = {10.1080/19942060.2024.2407005}
}

@article{chen2026aihw2035shapingdecade,
  title   = {AI+HW 2035: Shaping the Next Decade}, 
  author  = {Deming Chen and Jason Cong and Azalia Mirhoseini and Christos Kozyrakis and Subhasish Mitra and Jinjun Xiong and Cliff Young and Anima Anandkumar and Michael Littman and Aron Kirschen and Sophia Shao and Serge Leef and Naresh Shanbhag and Dejan Milojicic and Michael Schulte and Gert Cauwenberghs and Jerry M. Chow and Tri Dao and Kailash Gopalakrishnan and Richard Ho and Hoshik Kim and Kunle Olukotun and David Z. Pan and Mark Ren and Dan Roth and Aarti Singh and Yizhou Sun and Yusu Wang and Yann LeCun and Ruchir Puri},
  journal = {arXiv preprint arXiv:2603.05225},
  year    = {2026},
  url     = {https://arxiv.org/abs/2603.05225}
}

@article{patil2014contaminant,
  title     = {Contaminant transport through porous media: An overview of experimental and numerical studies},
  author    = {Patil, SB and Chore, HS},
  journal   = {Advances in environmental research},
  volume    = {3},
  number    = {1},
  pages     = {45--69},
  year      = {2014},
  publisher = {Techno-Press}
}

@article{leonard1988direct,
  title     = {Direct numerical simulation of turbulent flows with chemical reaction},
  author    = {Leonard, Andy D and Hill, James C},
  journal   = {Journal of scientific computing},
  volume    = {3},
  number    = {1},
  pages     = {25--43},
  year      = {1988},
  publisher = {Springer}
}

@article{altman2018curse,
  title   = {The curse (s) of dimensionality},
  author  = {Altman, Naomi and Krzywinski, Martin},
  journal = {Nat Methods},
  volume  = {15},
  number  = {6},
  pages   = {399--400},
  year    = {2018}
}

@article{sreekumar2026snapshots,
  title     = {From Snapshots to Manifolds: A Practical Roadmap for Interpolatory Reduced-Order-Modeling},
  author    = {Sreekumar, Abhilash and Komis, Giannikos and Barman, Swarup and Astolfi, Alessandro and Chronopoulos, Dimitrios},
  journal   = {Proceedings Of The Royal Society A-Mathematical Physical And Engineering Sciences},
  year      = {2026},
  publisher = {Royal Society, The}
}

@book{quarteroni2015reduced,
  title     = {Reduced basis methods for partial differential equations: an introduction},
  author    = {Quarteroni, Alfio and Manzoni, Andrea and Negri, Federico},
  year      = {2015},
  publisher = {Springer}
}

@incollection{chipman2020proofs,
  title     = {“Proofs” and Proofs of the Eckart--Young Theorem},
  author    = {Chipman, John S},
  booktitle = {Stochastic processes and functional analysis},
  pages     = {71--83},
  year      = {2020},
  publisher = {CRC Press}
}

@phdthesis{barros2023phd,
  author  = {Gabriel Freguglia Barros},
  title   = {Contributions to the Application of Snapshot-Based Data-Driven Methods in Computational Science and Engineering},
  school  = {Federal University of Rio de Janeiro (COPPE/UFRJ)},
  year    = {2023},
  address = {Rio de Janeiro, Brazil},
  type    = {Dissertation},
  note    = {Awarded the 2024 ABMEC Best Doctoral Thesis Prize}
}

@phdthesis{romulo2023phd,
  author  = {Rômulo Montalvão Silva},
  title   = {Advances in Physics-Informed Neural Networks for Parameter Estimation and Reconstruction of Multiphase Fluid Flows},
  school  = {Federal University of Rio de Janeiro (COPPE/UFRJ)},
  year    = {2023},
  address = {Rio de Janeiro, Brazil},
  type    = {Dissertation}
}

@inproceedings{zandieh2026turboquant,
  title     = {TurboQuant: Online Vector Quantization with Near-optimal Distortion Rate},
  author    = {Amir Zandieh and Majid Daliri and Majid Hadian and Vahab Mirrokni},
  booktitle = {The Fourteenth International Conference on Learning Representations},
  year      = {2026}
}

@incollection{axtmann2017scalability,
  title     = {Scalability of OpenFOAM with large eddy simulations and DNS on high-performance systems},
  author    = {Axtmann, Gabriel and Rist, Ulrich},
  booktitle = {High Performance Computing in Science and Engineering{\'{}} 16: Transactions of the High Performance Computing Center, Stuttgart (HLRS) 2016},
  pages     = {413--424},
  year      = {2017},
  publisher = {Springer}
}

@article{zhao2025applications,
  title     = {Applications of machine learning in real-time control systems: a review},
  author    = {Zhao, Xiaoning and Sun, Yougang and Li, Yanmin and Jia, Ning and Xu, Junqi},
  journal   = {Measurement Science and Technology},
  volume    = {36},
  number    = {1},
  pages     = {012003},
  year      = {2025},
  publisher = {IOP Publishing}
}

@article{wang2022recent,
  title     = {Recent advances in surrogate modeling methods for uncertainty quantification and propagation},
  author    = {Wang, Chong and Qiang, Xin and Xu, Menghui and Wu, Tao},
  journal   = {Symmetry},
  volume    = {14},
  number    = {6},
  pages     = {1219},
  year      = {2022},
  publisher = {MDPI}
}

@article{dbouk2017review,
  title     = {A review about the engineering design of optimal heat transfer systems using topology optimization},
  author    = {Dbouk, Talib},
  journal   = {Applied Thermal Engineering},
  volume    = {112},
  pages     = {841--854},
  year      = {2017},
  publisher = {Elsevier}
}

@article{alizadeh2020managing,
  title     = {Managing computational complexity using surrogate models: a critical review},
  author    = {Alizadeh, Reza and Allen, Janet K. and Mistree, Farrokh},
  journal   = {Research in Engineering Design},
  volume    = {31},
  number    = {3},
  pages     = {275--298},
  year      = {2020},
  publisher = {Springer Science and Business Media LLC},
  doi       = {10.1007/s00163-020-00336-7}
}

@article{ruthotto2020deep,
  title     = {Deep neural networks motivated by partial differential equations},
  author    = {Ruthotto, Lars and Haber, Eldad},
  journal   = {Journal of Mathematical Imaging and Vision},
  volume    = {62},
  number    = {3},
  pages     = {352--364},
  year      = {2020},
  publisher = {Springer}
}

@incollection{COLBROOK2024127,
  title     = {Chapter 4 - The multiverse of dynamic mode decomposition algorithms},
  author    = {Matthew J. Colbrook},
  editor    = {Siddhartha Mishra and Alex Townsend},
  booktitle = {Numerical Analysis Meets Machine Learning},
  series    = {Handbook of Numerical Analysis},
  publisher = {Elsevier},
  volume    = {25},
  pages     = {127--230},
  year      = {2024},
  doi       = {10.1016/bs.hna.2024.05.004}
}

@inproceedings{Li2021,
  title     = {{Fourier Neural Operator for Parametric Partial Differential Equations}},
  author    = {Li, Zongyi and Kovachki, Nikola and Azizzadenesheli, Kamyar and Liu, Burigede and Bhattacharya, Kaushik and Stuart, Andrew and Anandkumar, Anima},
  booktitle = {ICLR 2021 - 9th International Conference on Learning Representations},
  year      = {2021}
}

@article{Lu2021,
  title   = {{Learning nonlinear operators via DeepONet based on the universal approximation theorem of operators}},
  author  = {Lu, Lu and Jin, Pengzhan and Pang, Guofei and Zhang, Zhongqiang and Karniadakis, George Em},
  journal = {Nature Machine Intelligence},
  year    = {2021},
  doi     = {10.1038/s42256-021-00302-5}
}

@article{PEHERSTORFER2016196,
  title   = {Data-driven operator inference for nonintrusive projection-based model reduction},
  author  = {Benjamin Peherstorfer and Karen Willcox},
  journal = {Computer Methods in Applied Mechanics and Engineering},
  volume  = {306},
  pages   = {196--215},
  year    = {2016},
  doi     = {10.1016/j.cma.2016.03.025}
}

@article{karniadakis2021physics,
  title     = {Physics-informed machine learning},
  author    = {Karniadakis, George Em and Kevrekidis, Ioannis G and Lu, Lu and Perdikaris, Paris and Wang, Sifan and Yang, Liu},
  journal   = {Nature Reviews Physics},
  volume    = {3},
  number    = {6},
  pages     = {422--440},
  year      = {2021},
  publisher = {Nature Publishing Group UK London}
}

@inproceedings{rudi2015extreme,
  title     = {An extreme-scale implicit solver for complex PDEs: Highly heterogeneous flow in earth's mantle},
  author    = {Rudi, Johann and Malossi, A Cristiano I and Isaac, Tobin and Stadler, Georg and Gurnis, Michael and Staar, Peter WJ and Ineichen, Yves and Bekas, Costas and Curioni, Alessandro and Ghattas, Omar},
  booktitle = {Proceedings of the international conference for high performance computing, networking, storage and analysis},
  pages     = {1--12},
  year      = {2015}
}

@article{XING201353,
  title   = {A three-dimensional hydrodynamic and salinity transport model of estuarine circulation with an application to a macrotidal estuary},
  author  = {Yan Xing and Congfang Ai and Sheng Jin},
  journal = {Applied Ocean Research},
  volume  = {39},
  pages   = {53--71},
  year    = {2013},
  doi     = {10.1016/j.apor.2012.10.003}
}

@article{bae2005direct,
  title     = {Direct numerical simulation of turbulent supercritical flows with heat transfer},
  author    = {Bae, Joong Hun and Yoo, Jung Yul and Choi, Haecheon},
  journal   = {Physics of fluids},
  volume    = {17},
  number    = {10},
  year      = {2005},
  publisher = {AIP Publishing}
}

@article{deen2012direct,
  title     = {Direct numerical simulation of flow and heat transfer in dense fluid--particle systems},
  author    = {Deen, Niels G and Kriebitzsch, Sebastian HL and van der Hoef, Martin A and Kuipers, JAM},
  journal   = {Chemical engineering science},
  volume    = {81},
  pages     = {329--344},
  year      = {2012},
  publisher = {Elsevier}
}

@article{Meiburg,
  title     = {Turbidity Currents and Their Deposits},
  author    = {Meiburg, Eckart and Kneller, Ben},
  journal   = {Annual Review of Fluid Mechanics},
  volume    = {42},
  number    = {Volume 42, 2010},
  pages     = {135--156},
  year      = {2010},
  publisher = {Annual Reviews},
  doi       = {10.1146/annurev-fluid-121108-145618}
}

@article{NECKER_HARTEL_KLEISER_MEIBURG_2005,
  title   = {Mixing and dissipation in particle-driven gravity currents},
  author  = {Necker, F. and Hartel, C. and Kleiser, L. and Meiburg, E.},
  journal = {Journal of Fluid Mechanics},
  volume  = {545},
  pages   = {339--372},
  year    = {2005},
  doi     = {10.1017/S0022112005006932}
}

@article{ohana2024well,
  title   = {The well: a large-scale collection of diverse physics simulations for machine learning},
  author  = {Ohana, Ruben and McCabe, Michael and Meyer, Lucas and Morel, Rudy and Agocs, Fruzsina and Beneitez, Miguel and Berger, Marsha and Burkhart, Blakesly and Dalziel, Stuart and Fielding, Drummond and others},
  journal = {Advances in Neural Information Processing Systems},
  volume  = {37},
  pages   = {44989--45037},
  year    = {2024}
}

@book{heath2018scientific,
  title     = {Scientific computing: an introductory survey, revised second edition},
  author    = {Heath, Michael T},
  year      = {2018},
  publisher = {SIAM}
}

@article{NECKER2002279,
  title   = {High-resolution simulations of particle-driven gravity currents},
  author  = {F. Necker and C. Härtel and L. Kleiser and E. Meiburg},
  journal = {International Journal of Multiphase Flow},
  volume  = {28},
  number  = {2},
  pages   = {279--300},
  year    = {2002},
  doi     = {10.1016/S0301-9322(01)00065-9}
}

@article{valli_1,
  title   = {On decoupled time step/subcycling and iteration strategies for multiphysics problems},
  author  = {Valli, A. M. P. and Carey, G. F. and Coutinho, A. L. G. A.},
  journal = {Communications in Numerical Methods in Engineering},
  volume  = {24},
  number  = {12},
  pages   = {1941--1952},
  year    = {2008},
  doi     = {10.1002/cnm.1085}
}

@incollection{Cinelli2021,
  title     = {{Variational methods for machine learning with applications to deep networks}},
  author    = {Cinelli, Lucas Pinheiro and Marins, Matheus Ara{\'{u}}jo and da Silva, Eduardo Ant{\^{o}}nio Barros and Netto, S{\'{e}}rgio Lima},
  booktitle = {Variational Methods for Machine Learning with Applications to Deep Networks},
  publisher = {Springer},
  year      = {2021},
  doi       = {10.1007/978-3-030-70679-1}
}

@inproceedings{Kingma2014,
archivePrefix = {arXiv},
arxivId = {1312.6114},
author = {Kingma, Diederik P. and Welling, Max},
booktitle = {2nd International Conference on Learning Representations, ICLR 2014 - Conference Track Proceedings},
doi = {10.61603/ceas.v2i1.33},
eprint = {1312.6114},
title = {{Auto-encoding variational bayes}},
year = {2014}
}

@article{diederik2019introduction,
  title     = {An introduction to variational autoencoders},
  author    = {Kingma, Diederik P and Welling, Max},
  journal   = {Foundations and Trends{\textregistered} in Machine Learning},
  volume    = {12},
  number    = {4},
  pages     = {307--392},
  year      = {2019},
  publisher = {Emerald Publishing Limited}
}

@article{Burgess2018,
  title   = {{Understanding disentangling in $\beta$-VAE}},
  author  = {Burgess, Christopher P. and Higgins, Irina and Pal, Arka and Matthey, Loic and Watters, Nick and Desjardins, Guillaume and Lerchner, Alexander},
  journal = {arXiv preprint arXiv:1804.03599},
  year    = {2018},
  url     = {https://arxiv.org/abs/1804.03599}
}

@article{EIVAZI2022117038,
  title   = {Towards extraction of orthogonal and parsimonious non-linear modes from turbulent flows},
  author  = {Hamidreza Eivazi and Soledad {Le Clainche} and Sergio Hoyas and Ricardo Vinuesa},
  journal = {Expert Systems with Applications},
  volume  = {202},
  pages   = {117038},
  year    = {2022},
  doi     = {10.1016/j.eswa.2022.117038}
}

@article{gan_rozza,
  title   = {Generative adversarial reduced order modelling},
  author  = {Coscia, Dario and Demo, Nicola and Rozza, Gianluigi},
  journal = {Scientific Reports},
  volume  = {14},
  number  = {1},
  pages   = {3826},
  year    = {2024},
  doi     = {10.1038/s41598-024-54067-z}
}

@article{choi2025definingfoundationmodelscomputational,
  title   = {Defining Foundation Models for Computational Science: A Call for Clarity and Rigor},
  author  = {Youngsoo Choi and Siu Wun Cheung and Youngkyu Kim and Ping-Hsuan Tsai and Alejandro N. Diaz and Ivan Zanardi and Seung Whan Chung and Dylan Matthew Copeland and Coleman Kendrick and William Anderson and Traian Iliescu and Matthias Heinkenschloss},
  journal = {arXiv preprint arXiv:2505.22904},
  year    = {2025},
  url     = {https://arxiv.org/abs/2505.22904}
}

@article{Krake2021,
  title   = {Visualization and selection of Dynamic Mode Decomposition components for unsteady flow},
  author  = {T. Krake and S. Reinhardt and M. Hlawatsch and B. Eberhardt and D. Weiskopf},
  journal = {Visual Informatics},
  volume  = {5},
  number  = {3},
  pages   = {15--27},
  year    = {2021},
  doi     = {10.1016/j.visinf.2021.06.003}
}

@inproceedings{NIPS2015_bc731692,
  title     = {Variational Dropout and the Local Reparameterization Trick},
  author    = {Kingma, Durk P and Salimans, Tim and Welling, Max},
  booktitle = {Advances in Neural Information Processing Systems},
  editor    = {C. Cortes and N. Lawrence and D. Lee and M. Sugiyama and R. Garnett},
  volume    = {28},
  year      = {2015},
  publisher = {Curran Associates, Inc.}
}

@article{Barros2020,
  title   = {{Dynamic mode decomposition for density-driven gravity current simulations}},
  author  = {Barros, G F. and C{\^{o}}rtes, A. M. A. and Coutinho, A. L.G.A.},
  journal = {CILAMCE 2020 - Proceedings of the XLI Ibero-Latin-American Congress on Computational Methods in Engineering},
  year    = {2020}
}

@article{lindstrom2014fixed,
  title     = {Fixed-rate compressed floating-point arrays},
  author    = {Lindstrom, Peter},
  journal   = {IEEE transactions on visualization and computer graphics},
  volume    = {20},
  number    = {12},
  pages     = {2674--2683},
  year      = {2014},
  publisher = {IEEE}
}

@article{cappello2019use,
  title     = {Use cases of lossy compression for floating-point data in scientific data sets},
  author    = {Cappello, Franck and Di, Sheng and Li, Sihuan and Liang, Xin and Gok, Ali Murat and Tao, Dingwen and Yoon, Chun Hong and Wu, Xin-Chuan and Alexeev, Yuri and Chong, Frederic T},
  journal   = {The International Journal of High Performance Computing Applications},
  volume    = {33},
  number    = {6},
  pages     = {1201--1220},
  year      = {2019},
  publisher = {Sage Publications Sage UK: London, England}
}

@article{di2025survey,
  title     = {A survey on error-bounded lossy compression for scientific datasets},
  author    = {Di, Sheng and Liu, Jinyang and Zhao, Kai and Liang, Xin and Underwood, Robert and Zhang, Zhaorui and Shah, Milan and Huang, Yafan and Huang, Jiajun and Yu, Xiaodong and others},
  journal   = {ACM computing surveys},
  volume    = {57},
  number    = {11},
  pages     = {1--38},
  year      = {2025},
  publisher = {ACM New York, NY}
}

@inproceedings{bowman2016generating,
  title     = {Generating sentences from a continuous space},
  author    = {Bowman, Samuel and Vilnis, Luke and Vinyals, Oriol and Dai, Andrew and Jozefowicz, Rafal and Bengio, Samy},
  booktitle = {Proceedings of the 20th SIGNLL conference on computational natural language learning},
  pages     = {10--21},
  year      = {2016}
}

@article{rezende2018tamingvaes,
  title   = {Taming VAEs},
  author  = {Danilo Jimenez Rezende and Fabio Viola},
  journal = {arXiv preprint arXiv:1810.00597},
  year    = {2018},
  url     = {https://arxiv.org/abs/1810.00597}
}

@incollection{Peherstorfer2018SurveyOptimization,
  title     = {{Survey of multifidelity methods in uncertainty propagation, inference, and optimization}},
  author    = {Peherstorfer, Benjamin and Willcox, Karen and Gunzburger, Max},
  booktitle = {SIAM Review},
  year      = {2018},
  doi       = {10.1137/16M1082469}
}

@article{Velho2025,
  title     = {{Advances in data-driven reduced order models using two-stage dimension reduction for coupled viscous flow and transport}},
  author    = {Velho, Roberto M. and C{\^{o}}rtes, Adriano M.A. and Barros, Gabriel F. and Rochinha, Fernando A. and Coutinho, Alvaro L.G.A.},
  journal   = {Finite Elements in Analysis and Design},
  volume    = {248},
  year      = {2025},
  publisher = {Elsevier B.V.},
  doi       = {10.1016/j.finel.2025.104355}
}

@article{Solera-Rico2024-VariationalFlows,
  title     = {{{$\beta$}-Variational autoencoders and transformers for reduced-order modelling of fluid flows}},
  author    = {Solera-Rico, Alberto and Sanmiguel Vila, Carlos and G{\'{o}}mez-L{\'{o}}pez, Miguel and Wang, Yuning and Almashjary, Abdulrahman and Dawson, Scott T.M. and Vinuesa, Ricardo},
  journal   = {Nature Communications},
  volume    = {15},
  number    = {1},
  year      = {2024},
  publisher = {Nature Research},
  doi       = {10.1038/s41467-024-45578-4}
}

@article{valli2005control,
  title     = {Control strategies for timestep selection in finite element simulation of incompressible flows and coupled reaction--convection--diffusion processes},
  author    = {Valli, AMP and Carey, GF and Coutinho, ALGA},
  journal   = {International Journal for numerical methods in fluids},
  volume    = {47},
  number    = {3},
  pages     = {201--231},
  year      = {2005},
  publisher = {Wiley Online Library}
}

@book{carey-grids,
  title     = {{Computational Grids: Generations, Adaptation \& Solution Strategies}},
  author    = {Carey, Graham F.},
  publisher = {CRC Press},
  year      = {1997}
}

@article{logg2012automated,
  title     = {Automated solution of differential equations by the finite element method: the FEniCS book},
  author    = {Logg, Anders and Mardal, Kent-Andre and Wells, Garth},
  journal   = {Lecture notes in computational science and engineering},
  year      = {2012},
  publisher = {Berlin: Springer}
}

@article{lins2010residual,
  title     = {Residual-based variational multiscale simulation of free surface flows},
  author    = {Lins, Erb F and Elias, Renato N and Rochinha, Fernando A and Coutinho, Alvaro LGA},
  journal   = {Computational Mechanics},
  volume    = {46},
  number    = {4},
  pages     = {545--557},
  year      = {2010},
  publisher = {Springer}
}

@article{guerra2013numerical,
  title     = {Numerical simulation of particle-laden flows by the residual-based variational multiscale method},
  author    = {Guerra, Gabriel M and Zio, Souleymane and Camata, Jose J and Rochinha, Fernando A and Elias, Renato N and Paraizo, Paulo LB and Coutinho, Alvaro LGA},
  journal   = {International Journal for Numerical Methods in Fluids},
  volume    = {73},
  number    = {8},
  pages     = {729--749},
  year      = {2013},
  publisher = {Wiley Online Library}
}

@article{raissi2019physics,
  title     = {Physics-informed neural networks: A deep learning framework for solving forward and inverse problems involving nonlinear partial differential equations},
  author    = {Raissi, Maziar and Perdikaris, Paris and Karniadakis, George E},
  journal   = {Journal of Computational Physics},
  volume    = {378},
  pages     = {686--707},
  year      = {2019},
  publisher = {Elsevier}
}

@article{Lu2021_hc,
  title     = {Physics-Informed Neural Networks with Hard Constraints for Inverse Design},
  author    = {Lu Lu and Raphaël Pestourie and Wenjie Yao and Zhicheng Wang and Francesc Verdugo and Steven G. Johnson},
  journal   = {{SIAM} Journal on Scientific Computing},
  volume    = {43},
  number    = {6},
  pages     = {B1105--B1132},
  year      = {2021},
  publisher = {Society for Industrial {\&} Applied Mathematics ({SIAM})},
  doi       = {10.1137/21m1397908}
}

@article{Wu2023,
  title     = {A comprehensive study of non-adaptive and residual-based adaptive sampling for physics-informed neural networks},
  author    = {Wu, Chenxi and Zhu, Min and Tan, Qinyang and Kartha, Yadhu and Lu, Lu},
  journal   = {Computer Methods in Applied Mechanics and Engineering},
  volume    = {403},
  pages     = {115671},
  year      = {2023},
  publisher = {Elsevier BV},
  doi       = {10.1016/j.cma.2022.115671}
}

@article{githubGitHubErichsonoptht,
  title   = {The Optimal Hard Threshold for Singular Values is $4/\sqrt{3}$},
  author  = {Gavish, Matan and Donoho, David L.},
  journal = {IEEE Transactions on Information Theory},
  volume  = {60},
  number  = {8},
  pages   = {5040--5053},
  year    = {2014},
  doi     = {10.1109/TIT.2014.2323359}
}

@book{Jiji2009,
  title     = {Heat Convection},
  author    = {Jiji, Latif M.},
  year      = {2009},
  publisher = {Springer Berlin Heidelberg},
  doi       = {10.1007/978-3-642-02971-4}
}

@article{deteix2014,
  title   = {A Coupled Prediction Scheme for Solving the Navier--Stokes and Convection-Diffusion Equations},
  author  = {Deteix, J. and Jendoubi, A. and Yakoubi, D.},
  journal = {SIAM Journal on Numerical Analysis},
  volume  = {52},
  number  = {5},
  pages   = {2415--2439},
  year    = {2014},
  doi     = {10.1137/130942516}
}

@article{allen1982numerical,
  title     = {Numerical simulation of contaminant dispersion in estuary flows},
  author    = {Allen, Catherine M},
  journal   = {Proceedings of the Royal Society of London. A. Mathematical and Physical Sciences},
  volume    = {381},
  number    = {1780},
  pages     = {179--194},
  year      = {1982},
  publisher = {The Royal Society London}
}

@article{oberainotes,
  title   = {Deep Learning and Computational Physics (Lecture Notes)},
  author  = {Ray, Deep and Pinti, Orazio and Oberai, Assad A.},
  journal = {arXiv preprint arXiv:2301.00942},
  year    = {2023},
  url     = {https://arxiv.org/abs/2301.00942}
}

@article{Cuomo2022,
  title     = {Scientific Machine Learning Through Physics{\textendash}Informed Neural Networks: Where we are and What's Next},
  author    = {Salvatore Cuomo and Vincenzo Schiano Di Cola and Fabio Giampaolo and Gianluigi Rozza and Maziar Raissi and Francesco Piccialli},
  journal   = {Journal of Scientific Computing},
  volume    = {92},
  number    = {3},
  year      = {2022},
  publisher = {Springer Science and Business Media {LLC}},
  doi       = {10.1007/s10915-022-01939-z}
}

@article{PSAROS2023111902,
  title   = {Uncertainty quantification in scientific machine learning: Methods, metrics, and comparisons},
  author  = {Apostolos F. Psaros and Xuhui Meng and Zongren Zou and Ling Guo and George Em Karniadakis},
  journal = {Journal of Computational Physics},
  volume  = {477},
  pages   = {111902},
  year    = {2023},
  doi     = {10.1016/j.jcp.2022.111902}
}

@article{relobralo,
  title     = {Multi-Objective Loss Balancing for Physics-Informed Deep Learning},
  author    = {Bischof, Rafael and Kraus, Michael A.},
  journal   = {Computer Methods in Applied Mechanics and Engineering},
  volume    = {439},
  pages     = {117914},
  year      = {2025},
  publisher = {Elsevier BV},
  doi       = {10.1016/j.cma.2025.117914}
}

@article{gradnorm,
  title   = {GradNorm: Gradient Normalization for Adaptive Loss Balancing in Deep Multitask Networks},
  author  = {Chen, Zhao and Badrinarayanan, Vijay and Lee, Chen-Yu and Rabinovich, Andrew},
  journal = {arXiv preprint arXiv:1711.02257},
  year    = {2017},
  url     = {https://arxiv.org/abs/1711.02257}
}

@article{softadapt,
  title   = {SoftAdapt: Techniques for Adaptive Loss Weighting of Neural Networks with Multi-Part Loss Functions},
  author  = {Heydari, A. Ali and Thompson, Craig A. and Mehmood, Asif},
  journal = {arXiv preprint arXiv:1912.12355},
  year    = {2019},
  url     = {https://arxiv.org/abs/1912.12355}
}

@article{McClenny2022,
  title     = {Self-Adaptive Physics-Informed Neural Networks},
  author    = {Levi McClenny and Ulisses Braga-Neto},
  journal   = {{SSRN} Electronic Journal},
  year      = {2022},
  publisher = {Elsevier {BV}},
  doi       = {10.2139/ssrn.4086448}
}

@article{JIN2021109951,
  title   = {NSFnets (Navier-Stokes flow nets): Physics-informed neural networks for the incompressible Navier-Stokes equations},
  author  = {Xiaowei Jin and Shengze Cai and Hui Li and George Em Karniadakis},
  journal = {Journal of Computational Physics},
  volume  = {426},
  pages   = {109951},
  year    = {2021},
  doi     = {10.1016/j.jcp.2020.109951}
}

@incollection{Lumley1967,
  title     = {{The structure of inhomogeneous turbulence}},
  author    = {Lumley, J L},
  booktitle = {Atmospheric Turbulence and Radio Wave Propagation, edited by A. M. Yaglom and V. I. Tatarski (Nauka, Moscow)},
  pages     = {166--178},
  year      = {1967}
}

@incollection{Schmid2021DynamicVariants,
  title     = {{Dynamic Mode Decomposition and Its Variants}},
  author    = {Schmid, Peter J.},
  booktitle = {Annual Review of Fluid Mechanics},
  volume    = {54},
  pages     = {225--254},
  year      = {2021},
  doi       = {10.1146/annurev-fluid-030121-015835}
}

@article{Schmid2010DynamicData,
  title   = {{Dynamic mode decomposition of numerical and experimental data}},
  author  = {Schmid, Peter J.},
  journal = {Journal of Fluid Mechanics},
  volume  = {656},
  pages   = {5--28},
  year    = {2010},
  doi     = {10.1017/S0022112010001217}
}

@article{Kovachki2023,
  title   = {{Neural Operator: Learning Maps Between Function Spaces With Applications to PDEs}},
  author  = {Kovachki, Nikola and Zongyi Li, Nvidia and Burigede Liu, Caltech and Azizzadenesheli, Kamyar and Kaushik Bhattacharya, Nvidia and Stuart, Andrew and Anandkumar, Anima and Editor, Caltech and Rosasco, Lorenzo and Li, Zongyi and Liu, Burigede and Bhattacharya, Kaushik},
  journal = {Journal of Machine Learning Research},
  volume  = {24},
  pages   = {1--97},
  year    = {2023}
}

@article{Taira2017,
  title   = {{Modal analysis of fluid flows: An overview}},
  author  = {Taira, Kunihiko and Brunton, Steven L. and Dawson, Scott T.M. and Rowley, Clarence W. and Colonius, Tim and McKeon, Beverley J. and Schmidt, Oliver T. and Gordeyev, Stanislav and Theofilis, Vassilios and Ukeiley, Lawrence S.},
  journal = {AIAA Journal},
  volume  = {55},
  number  = {12},
  pages   = {4013--4041},
  year    = {2017},
  doi     = {10.2514/1.J056060}
}
\end{document}